%% file: iclr2023_conference.tex
\documentclass{article} 

\usepackage{tabularx}
\usepackage{iclr2023_conference,times}

\input{math_commands.tex}

\usepackage{hyperref}
\usepackage{url}

\usepackage{tikz}
\usetikzlibrary{arrows, decorations.text, shapes.geometric, positioning, decorations.pathreplacing, calligraphy}
\usepackage{pgfplots}
\pgfplotsset{compat=1.18}
\usepackage{booktabs}
\usepackage{subcaption}
\usepackage{wrapfig,lipsum,booktabs}
\usepackage{wrapfig}
\usepackage{multirow}

\title{PolyLM: An Open Source Polyglot Large Language Model}

\iclrfinalcopy
\makeatletter

\author{\centerline{Xiangpeng Wei$^*$, Haoran Wei$^*$, Huan Lin$^*$, Tianhao Li$^*$, Pei Zhang$^*$, Xingzhang Ren$^*$, Mei Li$^*$}\\ \centerline{\textbf{Yu Wan}\thanks{Major contributors.},\ \   \textbf{Zhiwei Cao}$^\dagger$, \textbf{Binbin Xie}$^\dagger$, \textbf{Tianxiang Hu}$^\dagger$, \textbf{Shangjie Li}\thanks{Contribution during internship at Alibaba DAMO Academy.},\ \  \textbf{Binyuan Hui}, \textbf{Bowen Yu}}\\
\centerline{\textbf{Dayiheng Liu}$^\ddagger$, \textbf{Baosong Yang}\thanks{Corresponding authors: \{liudayiheng.ldyh, yangbaosong.ybs\}@alibaba-inc.com},\ \  \textbf{Fei Huang}, \textbf{Jun Xie}}\\ 
\\
\centerline{DAMO Academy, Alibaba Group}
}

\newcommand{\mySFTDatasetName}{\textcolor{black}{\textsc{MultiAlpaca}}} 
\newcommand{\mySFTDatasetNameS}{\textcolor{black}{\textsc{MultiAlpaca}}~} 

\begin{document}

\maketitle

\input{tex/0_abstract}


\input{tex/1_introduction}

\input{tex/6_related_work}

\input{tex/2_approach}

\input{tex/3_self_instruction}

\input{tex/4_benchmark}

\input{tex/5_experiments}

\input{tex/7_conclusion}



\bibliography{iclr2023_conference}
\bibliographystyle{iclr2023_conference}

\input{tex/appendix}

\end{document}

%% file: math_commands.tex
\usepackage{amsmath,amsfonts,bm}

\def\eqref#1{equation~\ref{#1}}

\def\1{\bm{1}}

\DeclareMathAlphabet{\mathsfit}{\encodingdefault}{\sfdefault}{m}{sl}
\SetMathAlphabet{\mathsfit}{bold}{\encodingdefault}{\sfdefault}{bx}{n}



%% file: tex/0_abstract.tex
\begin{abstract}

Large language models (LLMs) demonstrate remarkable ability to comprehend, reason, and generate following nature language instructions. However, the development of LLMs has been primarily focused on high-resource languages, such as English, thereby limiting their applicability and research in other languages. Consequently, we present \textsc{Poly}LM, a multilingual LLM trained on 640 billion (B) tokens, avaliable in two model sizes: 1.7B and 13B. To enhance its multilingual capabilities, we 1) integrate bilingual data into training data; and 2) adopt a curriculum learning strategy that increases the proportion of non-English data from 30\% in the first stage to 60\% in the final stage during pre-training. Further, we propose a multilingual self-instruct method which automatically generates 132.7K  diverse multilingual instructions for model fine-tuning. To assess the model's performance, we collect several existing multilingual tasks, including multilingual understanding, question answering, generation, and translation. Extensive experiments show that \textsc{Poly}LM surpasses other open-source models such as LLaMA and BLOOM on multilingual tasks while maintaining comparable performance in English. 
Our models, alone with the instruction data and multilingual benchmark, are available at: \url{https://modelscope.cn/models/damo/nlp_polylm_13b_text_generation}.
\end{abstract}

%% file: tex/1_introduction.tex
\section{Introduction}
Large language models (LLMs) are trained on vast amounts of data in a self-supervised fashion, which has shown promising performance in a variety of zero-shot and few-shot tasks \citep{gpt3,Chowdhery2022PaLMSL}. Fine-tuning these models on a diverse set of tasks allows them to handle unseen tasks following natural language instructions~\citep{ouyang2022training,longpre2023flan,taori2023alpaca,gpt4all}. These properties have attracted significant attention from the Artificial Intelligence community and offering a potential path towards artificial general intelligence. Unfortunately, most LLMs are developed for English, such as LLaMA~\citep{touvron2023llama}, BLOOM~\citep{scao2022bloom}, Chinchilla~\citep{hoffmann2022chinchilla}, OPT~\citep{zhang2022opt}. A main reason stems from recent findings that model performance is closely related to the scale of the training dataset~\citep{kaplan2020scaling,rae2021scaling,biderman2023pythia,touvron2023llama}, leading to predominant focus on resource-rich languages, particularly English.

The relatively high concentration of studies on English limits the research and usage of LLMs in other languages. 
For instance, Thai and Indonesian have over 300 million (M) speakers, yet the size of these two languages in common crawl-based dataset such as mC4~\citep{xue2020mt5} is only 80 billion (B) tokens, comprising a mere 3\% of the English data. Due to the insufficient high-quality internet data, LLM capabilities on low-resource languages fail to be easily improved through expanding their data size like English~\citep{kaplan2020scaling,rae2021scaling,biderman2023pythia}. As a result, existing open-source LLMs such as XGLM~\citep{lin-etal-2022-xglm}, BLOOM~\citep{scao2022bloom}, and LLaMA~\citep{touvron2023llama} perform relatively poor on these languages, some of which are entirely overlooked. It is crucial to explore multilingual LLMs to bridge this gap and achieve academic and social significance. 

Our goal is to enhance the exploration and utilization of LLMs for non-native English speakers. In this work, we fill three significant gaps in this field: 1) the absence of an open-source multilingual LLM; 2) the inadequate availability of multilingual instruction data; and 3) the lack of a unified evaluation benchmark for multilingual settings. 

Concretely, we first develop an open-source multilingual LLM from scratch, called \textbf{Poly}glot Large \textbf{L}anguage \textbf{M}odel (\textsc{Poly}LM, Section \ref{sec2}). Contrary to existing open-source multilingual LLMs that lack 13B model, we release \textsc{Poly}LM-13B and  \textsc{Poly}LM-1.7B to facilitate its usage. To construct \textsc{Poly}LM, we leverage a massive dataset of 640B tokens, culled from publicly available sources such as Wikipedia, mC4~\citep{xue2020mt5}, CC-100~\citep{Conneau2019UnsupervisedCR}. This dataset contains over 30\% of non-English languages, specifically covering 18 of the most commonly spoken languages.\footnote{According to \url{https://www.ethnologue.com/insights/most-spoken-language/}. 
Some languages with interchangeable and more widely used official languages are not given priority, such as Hindi, Wu Chinese, and Cantonese.} 
To alleviate the problem of insufficient data for low-resource languages, we propose a curriculum learning strategy. The training schedule increases the amount of data available for training in English during the initial phases, then ramping up the ratio of high-quality, low-resource languages as training progresses. We expect the method to enable the transfer of general knowledge from English to other languages, leading to significant improvements in overall performance.

In light of the supervised fine-tuning (SFT) stage, we construct a multilingual instruction dataset termed \mySFTDatasetNameS with 132,701 samples (Section \ref{sec3}). 
At present, there is a dearth of high-quality open-source multilingual SFT datasets. 
On the one hand, extant multilingual SFT datasets, e.g. xP3-MT~\citep{muennighoff2022crosslingual}, are acquired via machine translation, which potentially yields a style of translationese, a lack of cultural nuances, as well as translation errors. On the other hands, manually annotating instructions is a laborious and costly process that does not lend itself well to the incorporation of creative flourishes.  
Drawing inspiration from recent advances in self-instruct~\citep{wang2022self,taori2023alpaca}, we devise a multilingual self-instruct method to automatically generate instruction data. Utilizing 175 English seeds as a starting point, our method leverage multilingual seed translation, instruction generation, and filtering mechanisms to deliver high quality multilingual instruction data.

In order to assess the multilingual capabilities of LLM, we curate a benchmark derived from existing multilingual tasks (Section \ref{sec4}), including QA~\citep{tydiqa}, understanding~\citep{Conneau2019UnsupervisedCR,Yang2019PAWSXAC,tikhonov2021heads,ponti2020xcopa}, generation~\citep{Chen2021MTGAB}, and cross-lingual machine translation~\citep{barrault-etal-2020-findings}. The benchmark is constructed with meticulously prompting and finally covers 10 tasks across 15 languages. Extensive experiments (Section \ref{sec5}) demonstrate that our pretrained model outperforms open-source models of comparable model size (e.g. BLOOM, LLaMA, etc.) in non-English languages. Through in-depth analyses, we identify finding that the proposed curriculum training strategy boosts the multilingual performance while maintain the English proficiency. 
In addition, the use of multilingual instruction data markedly enhances the ability of \textsc{Poly}LM to tackle multilingual zero-shot tasks.

%% file: tex/6_related_work.tex
\section{Preliminary}

In this section, we begin with a review of the background on language modeling. We then examine previous research on knowledge transferring, and instruction learning of pre-trained LLMs, with a focus on their relevance to \textsc{PolyLM}. Finally, we outline our rationale for training \textsc{PolyLM}.

\textbf{Language Modeling} refers to the process of estimating the probability of a sequence of tokens, i.e. $p(\mathbf{x}) = p(x_1, x_2, ..., x_T) = \prod_{t=1}^{T} p(x_t|\mathbf{x}_{<t})$. This is also commonly referred to as autoregressive sequence modeling, as it involves predicting the future token at each time-step based on the preceding context. The initial language models were predominantly $n$-gram models that evaluate the likelihood of a sequence of tokens based on the frequency of its occurrence in a training corpus. Over the last two decades, neural networks have proven to be effective in the task of language modeling, including feed-forward models~\citep{conf/interspeech/MikolovKBCK10} and recurrent neural networks~\citep{bengio2000neural}. More recently, Transformer~\citep{Vaswani2017Attention}, a self-attention based neural network, has shown unparalleled language model performance~\citep{devlin-etal-2019-bert,radford2018improving}, and become the \textit{de facto} backbone of LLMs emerged in the past three years, such as GPT3~\citep{gpt3}, Gopher~\citep{rae2021scaling}, PaLM~\citep{anil2023palm}, BLOOM~\citep{scao2022bloom}, Chinchilla~\citep{hoffmann2022chinchilla}, GLM~\citep{zeng2022glm} and LLaMA~\citep{touvron2023llama}.

\textbf{Transfer Learning} is a rapidly evolving field of research that has garnered significant interest in recent years. In this scenario, models are initially trained on extensive unlabeled data, and then their acquired knowledge is applied to various downstream tasks through fine-tuning. Some of the most prominent works in this area include the ELMo~\citep{peters-etal-2018-deep}, BERT~\citep{devlin-etal-2019-bert} and GPT~\citep{radford2018improving} have demonstrated remarkable success. These developments subsequently prompt work~\citep{raffel2020exploring,radford2019language,xue2020mt5} on better results by adopting larger scale data and parameters to further improve model performance. Although pretraing-then-finetuning is still effective in achieving high performance with limited labeled data, recent advancements has shown that language models with extremely large scale parameters can perform tasks without further optimization. The most exemplary model is GPT3~\citep{gpt3}, which utilizes a contextualized approach by incorporating multiple input-output demonstrations and presenting them alongside the query. This effectively stimulates the model to generate accurate predictions, showcasing encouraging outcomes in zero/few-shot situations.

\textbf{Instruction Learning} aims to bring together various natural language processing tasks by framing them as question-answering exercises that operate over a given context. This approach enhances the value of LLMs by leveraging their existing knowledge. With the success of language models, there has been a growing interest in exploring their potential to comprehend and execute instructions. Several advanced researches~\citep{ouyang2022training,wei2022finetuned,peng2023instruction,ye2023context,zhou2023lima} have demonstrated a remarkable ability to generalize to new zero-shot tasks. However, they rely heavily on human-generated instruction data, which is frequently constrained in terms of quantity, diversity, and creativity, which is very time-consuming and labor-intensive. \cite{wang2022self} make an effort to construct a self-Instruct framework for improving the instruction-following capabilities of LLMs. Similarly, \cite{xu2023wizardlm} propose an evol-instruct framework to automatically rewrite simple human-written instructions step by step into more complex ones, to further improve instruction-followed LLMs.

In this paper, we propose \textsc{PolyLM} to address the following blanks and limitations in current LLM research, offering a comprehensive and innovative solution to advance this field.
\begin{itemize}
    \item We provide a 13B scale model that is proficient in the major non-English languages spoken worldwide, such as Spanish, Russian, Arabic, Japanese, Korean, Thai, Indonesian, and Chinese etc. It is a perfect complement to the existing open-source models, including: (1) LLaMA, English is predominant among the whole dataset. (2) BLOOM, lack of 13B version and fail to address languages spoken by significant populations, such as Japanese, Korean and Thai. (3) XGLM~\citep{lin-etal-2022-xglm}, the maximum version is 7B. (4) mGPT~\citep{shliazhko2022mgpt}, only 1.3B version is available.
    \item We suggest an advanced curriculum learning approach that facilitates the transfer of commonsense knowledge, acquired mainly in English, to diverse non-English languages and specific NLP downstream tasks such as machine translation.
    \item We propose \mySFTDatasetNameS to complement \textsc{Alpaca}~\citep{taori2023alpaca} and \textsc{Chinese-Alpaca}~\citep{cui2023efficient}, making LLMs better follow multilingual instructions, particularly those coming from non-native English speakers.
\end{itemize}

%% file: tex/2_approach.tex
\section{\textsc{Poly}LM: a polyglot large language model}
\label{sec2}

In this section, we present the design of \textsc{PolyLM}, which includes a detailed description of its training dataset (Section 3.1), architecture (Section 3.2), and training process (Section 3.3).

\subsection{Dataset}

The composition of the pre-training dataset used for \textsc{Poly}LM is shown in Table~\ref{tab:dataset}. Our pre-training dataset contains 640B tokens in total, of which English data accounts for 68\%. To develop \textsc{Poly}LM with multilingual capabilities, the pre-training dataset has about 32\% non-English multilingual data, which is a higher percentage of non-English data than most previous open-sourced large language models~\citep{biderman2023pythia,zhang2022opt,touvron2023llama,penedo2023refinedweb}. To be concrete, the English data contains documents with 425B tokens from multiple sources, such as The Pile~\citep{gao2020pile}, mC4~\citep{xue2020mt5}, and Wikipedia. While the 204B multilingual data tokens come from CC-100~\citep{Conneau2019UnsupervisedCR}, mC4~\citep{xue2020mt5}, Wikipedia. 
The multilingual data mainly covers the following languages: \texttt{zh}, \texttt{ar}, \texttt{es}, \texttt{fr}, \texttt{de}, \texttt{it}, \texttt{nl}, \texttt{ru}, \texttt{id}, \texttt{pl}, \texttt{pt}, \texttt{ja}, \texttt{th}, \texttt{tr}, \texttt{he}, \texttt{ko}, \texttt{vi}, with the distribution given in Table~\ref{table:rate_of_langs_in_train_data}. To enable the model ability of code understanding and generation, we also incorporate code data of 7.5B tokens from GitHub with permissioned licenses into our pre-training dataset. In order to further improve the cross-lingual and multilingual ability of the \textsc{Poly}LM, similar to PaLM2~\citep{anil2023palm}, we employ parallel multilingual data of 1B tokens into our pre-training dataset.

\begin{table}[t]
  \center
   \setlength{\tabcolsep}{3pt}
  \begin{tabular}{@{}l@{}cccr@{}}
\toprule
  Source &  Fraction &  Tokens & Type \\  %
 \midrule
  mC4   &  49.95\%  & 321.7B & Web-text (Multilingual) \\
  CC-100  & 32.31\%  & 208.1B  & Web-text (Multilingual) \\
  The Pile         &16.41\% & 105.7B & Web-text \& books (English) \\
  GitHub  & 1.17\% & 7.5B & Code \\
  OPUS  & 0.16\% & 1.0B &  Parallel Multilingual Data \\ 
  \midrule
  Sum   &-  & 638B \\
 \bottomrule
  \end{tabular}
  \caption{The composition of the \textsc{Poly}LM pre-training dataset.
  \label{tab:dataset}
  }
\end{table}

\begin{table}[]
    \centering
    \begin{tabular}{crr|crr}
        \toprule
            \textbf{Language} & \textbf{Tokens (B)} & \textbf{Percentage (\%)} & \textbf{Language} & \textbf{Tokens (B)} & \textbf{Percentage (\%)}\\
        \midrule
            En & 424.96 & 67.56 & Vi & 4.13 & 0.66 \\
            Zh & 139.29 & 22.14 & Id & 3.91 & 0.62 \\
            Ru & ~~7.61 & ~1.21 & Pl & 3.84 & 0.61 \\
            Es & ~~5.62 & ~0.89 & Nl & 3.52 & 0.56 \\
            De & ~~5.56 & ~0.88 & Ar & 3.48 & 0.55 \\
            Fr & ~~5.10 & ~0.81 & Tr & 3.42 & 0.54 \\
            It & ~~4.31 & ~0.69 & Th & 2.89 & 0.46 \\
            Pt & ~~4.27 & ~0.68 & He & 2.10 & 0.33 \\
            Ja & ~~4.19 & ~0.67 & Ko & 0.84 & 0.13 \\
        \bottomrule
    \end{tabular}
    \caption{Language distribution of the training data (excluding code and multilingual parallel data).}
    \label{table:rate_of_langs_in_train_data}
\end{table}

To build the pre-training dataset, we also develop a comprehensive data pre-processing pipeline that implements multiple techniques for data cleaning and filtering. The pipeline consists of the following stages: 

1) \textbf{Language identification}. We classify documents according to their primary languages and remove those with low confidence in classification, leveraging inexpensive n-gram models (e.g., fastText~\citep{joulin2016fasttext}).

2) \textbf{Rule-based filtering}. Following~\citet{rae2021scaling,scao2022bloom}, we eliminate irrelevant or low-quality content using various rules and heuristics, including \textit{repetition removal} (the document with the excessive line, paragraph, or n-gram repetitions is removed), \textit{document-wise filtering} (removing outlier documents by overall length, symbol-to-word ratio, the ratio of ellipsis, invisible characters, numbers, and dates, etc.), and \textit{line-wise corrections} (such as URL filtering, long words removal, and whitespace standardization).

3) \textbf{ML-based quality filtering}. We further filter low-quality multilingual documents using several small n-gram-based language models (e.g., KenLM~\citep{heafield2011kenlm}) for different languages trained on their gold-standard corpora. In addition, similar to~\citet{raffel2020exploring,smith2022using}, we also train a 2-gram fastText~\citep{joulin2016fasttext} classifier to filter the low-quality English documents. This classifier uses Wikipedia, and Books from The Pile~\citep{gao2020pile} as the positive samples and CommonCrawl web documents as the negative samples. To sum up, about 28.3\% data are filtered with Rule-based filtering and ML-based quality filtering.

4) \textbf{Deduplication}. In line with~\citet{raffel2020exploring}, we remove similar documents to reduce data redundancy with MinHashLSH-based fuzzy deduplication technology, where 23.1\% English documents and 18.6\% non-English documents are removed.

\begin{figure}
    \centering
    \includegraphics[width=\linewidth]{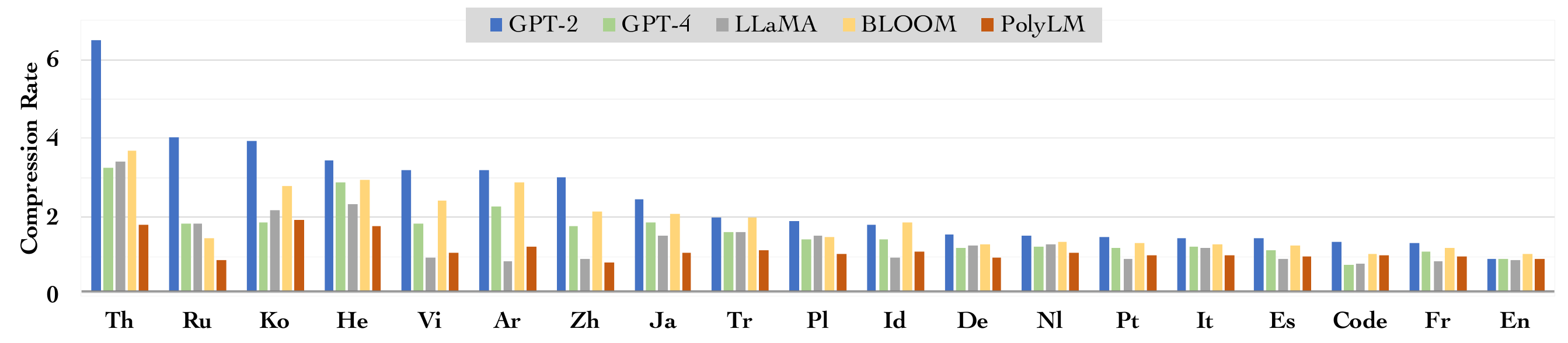}
    \caption{The compression rate of different tokenizers. We take XLM-R~\citep{Conneau2019UnsupervisedCR} tokenizer as the baseline, and set the compression rate of XLM-R tokenizer to 1.}
    \label{fig:tokenizer}
\end{figure}

\begin{table}[t]
\begin{center}
    \begin{tabular}{l|cc}
    Hyperparameter ($\downarrow$) & \textsc{Poly}LM-1.7B & \textsc{Poly}LM-13B \\ 
    \midrule
    \multicolumn{3}{c}{\emph{Architecture hyperparameters}} \\
    \midrule
    Number of parameters & 1,722M & 13,003M \\ 
    Precision & \multicolumn{2}{c}{\texttt{bfloat16}} \\ 
    Number of layers& 24 & 40 \\ 
    Hidden dimension & 2048 & 5120 \\ 
    Attention heads& 16 & 40 \\
    Vocab size & \multicolumn{2}{c}{256,000} \\ 
    Sequence length & \multicolumn{2}{c}{2048} \\
    Activation & \multicolumn{2}{c}{{GELU}} \\
    Position embedding & \multicolumn{2}{c}{{Absolute}} \\ 
    \midrule
    \multicolumn{3}{c}{\emph{Pretraining hyperparameters}} \\
    \midrule
    Global Batch Size & 512 & 2048 \\ 
    Learning rate peak & $1 \times 10^{-4}$ &  $6 \times 10^{-5}$  \\ 
    Total training tokens & \multicolumn{2}{c}{638B} \\
    Gradient clipping & \multicolumn{2}{c}{1.0} \\
    Weight decay & \multicolumn{2}{c}{0.1} \\
    \midrule
    \multicolumn{3}{c}{\emph{Multilingul Self-instruction finetuning hyperparameters}} \\
    \midrule
    Global Batch Size & 32 & 64 \\ 
    Sequence strategy & \multicolumn{2}{c}{The length is 2048 with packing} \\ 
    Learning rate & \multicolumn{2}{c}{1e-5} \\
    Total training tokens & \multicolumn{2}{c}{16M tokens} \\
    \end{tabular}
\caption{\textsc{Poly}LM Architecture and Training Hyperparameters.}
\label{tbl:architecture}
\end{center}
\end{table}

Based on the \textsc{Poly}LM multilingual pre-training dataset, we derived a vocabulary with 256K token entries using Byte-Pair Encoding (BPE)~\citep{sennrich2015neural} with the implementation from SentencePiece~\citep{kudo2018sentencepiece}. To enhance the mathematical capabilities of our model, we follow~\citet{touvron2023llama} to split all numbers into individual digits. The unknown characters are fallback to byte encoding of UTF-8 to guarantee the coverage of rare words (e.g., emoji, and special symbols). For tokenizer training, we sample multilingual documents with a similar distribution as~\citet{Conneau2019UnsupervisedCR} used to increase the number of vocabulary tokens associated with low-resource languages and alleviate the bias towards high-resource languages. We compare the compression rate on different language corpora of different tokenizers. We use XLM-R~\citep{Conneau2019UnsupervisedCR} tokenizer, which supports 100 languages, as the baseline (the compression rate of XLM-R tokenizer is set to 1). As shown in Figure~\ref{fig:tokenizer}, \textsc{Poly}LM has achieved significantly better compression rates in most covered languages, while maintaining the compression rate in English as BLOOM~\citep{scao2022bloom}, LLaMA~\citep{touvron2023llama}, GPT-2~\citep{radford2019language}, and GPT-4~\citep{openai2023gpt4}. Note that some open source models that are not friendly to language extensions, for example, LLaMA~\citep{touvron2023llama} only contain a 32K size vocabulary mostly composed of English tokens, which is not friendly to non-Latin languages. When improving a certain non-Latin language ability, the vocabulary needs to be expanded like Chinese-LLaMA~\citep{cui2023efficient}. On the contrary, \textsc{Poly}LM allows researchers to improve the model's ability in a covered language by simply continuing monolingual pre-training without expanding the vocabulary.

\subsection{Architecture}

It has become apparent that the computational cost of exploring different architectural designs for LLMs is prohibitive. Therefore, we present the distinctive design options of \textsc{Poly}LM\footnote{Recent research indicates that Rotary Position Encoding (RoPE)~\citep{su2021roformer} yields superior performance. Accordingly, we will switch to the latest Megatron-LM branch and promptly release 13B and 1.7B versions featuring RoPE.} in this section.

Following some endeavours on large language models, we develop a decoder-only autoregressive Transformer architecture detailed in~\cite{radford2019language}. To stabilize the training, we adopt Pre-LN~\citep{xiong2020layer}, i.e. $y = x + {\rm LayerNorm}(f(x))$ (where $f({\cdot})$ indicates the layer function) for layer normalization, and apply the Xavier normal initialization~\citep{Glorot2010UnderstandingTD} with bias terms are initialized to zero. To improve FFNs in Transformer, we replace ReLU with GeLU activation~\citep{Hendrycks2016GaussianEL}.

In this paper we present two Transformer language models with 1.7 billion and 13 billion parameters, respectively. The architectural details are displayed in Table~\ref{tbl:architecture}.

\subsection{Training}

\begin{figure*}
\footnotesize
\begin{subfigure}[b]{0.33\textwidth}
    \centering
    \resizebox{\linewidth}{!}{
        \includegraphics{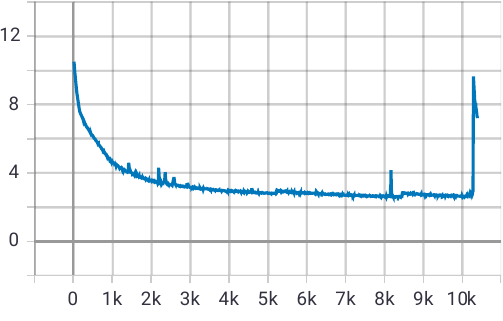}
    }
    \caption{Loss value}
    \label{fig:sec-2-3-lr-a}
\end{subfigure}
\begin{subfigure}[b]{0.33\textwidth}
    \centering
    \resizebox{\linewidth}{!}{
        \includegraphics{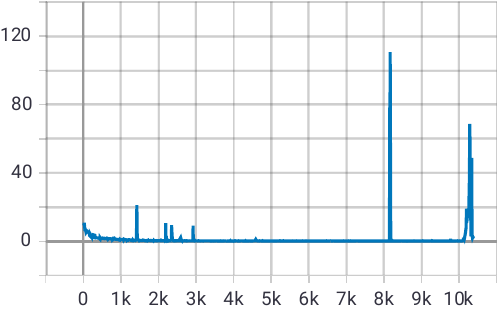}
    }
    \caption{Gradient norm}
    \label{fig:sec-2-3-lr-b}
\end{subfigure}
\begin{subfigure}[b]{0.3\textwidth}
    \centering
    \resizebox{\linewidth}{!}{
        \includegraphics{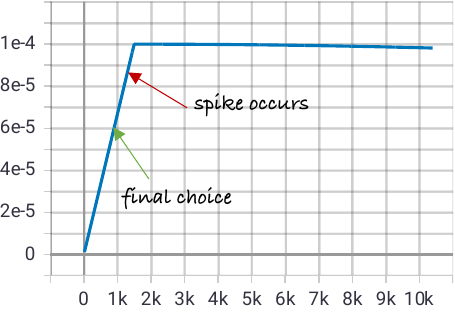}
    }
    \caption{Learning rate}
    \label{fig:sec-2-3-lr-c}
\end{subfigure}
\caption{Training curves over iterations for the 13B model with learning rate as $1\times 10^{-4}$.}
\label{fig:2-3-lr}
\end{figure*}

\begin{figure*}[t]
\footnotesize
\begin{subfigure}[b]{0.33\textwidth}
    \centering
    \resizebox{\linewidth}{!}{
        \includegraphics{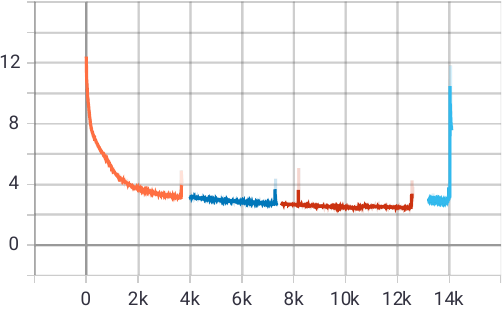}
    }
    \caption{Loss with \texttt{bfloat16} precision.}
    \label{fig:sec-2-3-mp-a}
\end{subfigure}
\begin{subfigure}[b]{0.33\textwidth}
    \centering
    \resizebox{\linewidth}{!}{
        \includegraphics{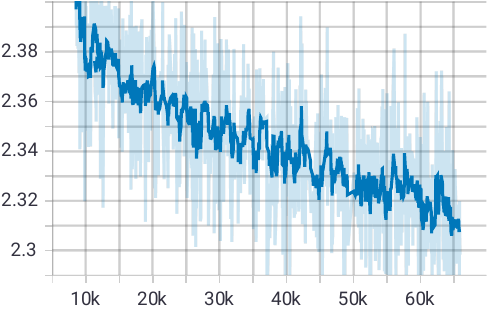}
    }
    \caption{Loss with mixed-precision.}
    \label{fig:sec-2-3-mp-b}
\end{subfigure}
\begin{subfigure}[b]{0.3\textwidth}
    \centering
    \resizebox{\linewidth}{!}{
        \includegraphics{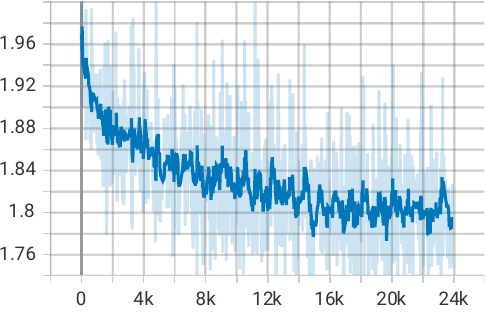}
    }
    \caption{Loss on curriculum learning.}
    \label{fig:sec-2-3-mp-c}
\end{subfigure}
\caption{Training curves over iterations for the 13B model with learning rate as $6\times 10^{-5}$.}
\label{fig:2-3-mp}
\end{figure*}

We train all models with a 2048 token context window, using the Adam ($\beta_1=0.9$, $\beta_2=0.95$) optimizer. We warm-up the learning rate from $1e^{-7}$ to the maximum learning rate over the first 2000 steps, and then decay it to 10\% of the maximal learning rate using a cosine schedule. We use a weight decay of 0.1 and gradient clipping of 1.0.

\textsc{Poly}LM was trained using Megatron-LM~\footnote{\url{https://github.com/NVIDIA/Megatron-LM}} on a cluster of 32 A100 GPU (8$\times$80G) servers. We apply tensor model parallelism within a single node, setting \texttt{tensor-model-parallel-size} as 8. When training a 13B-parameter model, our code processes around 1170 tokens/sec/GPU, thus training over our dataset containing 640B tokens takes approximately 29 days. However, we faced numerous unforeseen spikes and deviations in losses, which prolonged the entire training process to a duration of two months. There are several possible conditions that result in training collapses, and our unique choices to enhance training stability.

\textbf{Lower Maximal Learning Rate.} Learning rate is an important hyperparameter in neural network models that controls the magnitude of parameter updates. In our first few attempts, we drew inspiration from previous research which indicated that smaller models tend to benefit from higher learning rates. As such, we opted to set the learning rate to $1 \times 10^{-4}$. Without exception, all attempts to train \textsc{Poly}LM-13B have resulted in loss spikes with this choice in early stage, which tend to occur more frequently as the training progresses, as illustrated in Figure~\ref{fig:sec-2-3-lr-a}. We have noticed that the gradient norm shows significant fluctuations during the warm-up phase, when the learning rate is increasing linearly (see Figure~\ref{fig:sec-2-3-lr-b}).

The fundamental issue with instability during training is that a large learning rate can cause the gradient to grow too large, surpassing the model's capacity and resulting in a gradient explosion that prevents parameter updates. The problem is handled via reducing learning rate to $6 \times 10^{-5}$, i.e. a proper learning rate located before the step where the initial spike in loss occurs (Cf. Figure~\ref{fig:sec-2-3-lr-c}). 

\begin{figure}
    \centering
    \includegraphics[width=\linewidth]{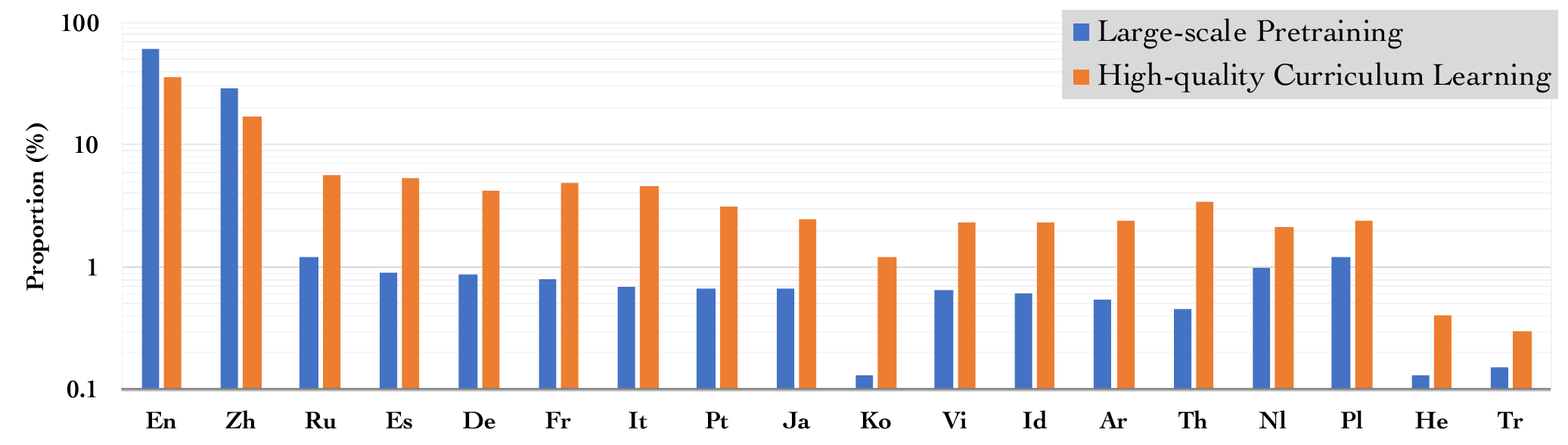}
    \caption{The proportion of multilingual data in curriculum learning significantly exceed that in the pretraining phrase.}
    \label{fig:lang_propertion}
\end{figure}

\begin{figure}
    \centering
    \includegraphics[width=\linewidth]{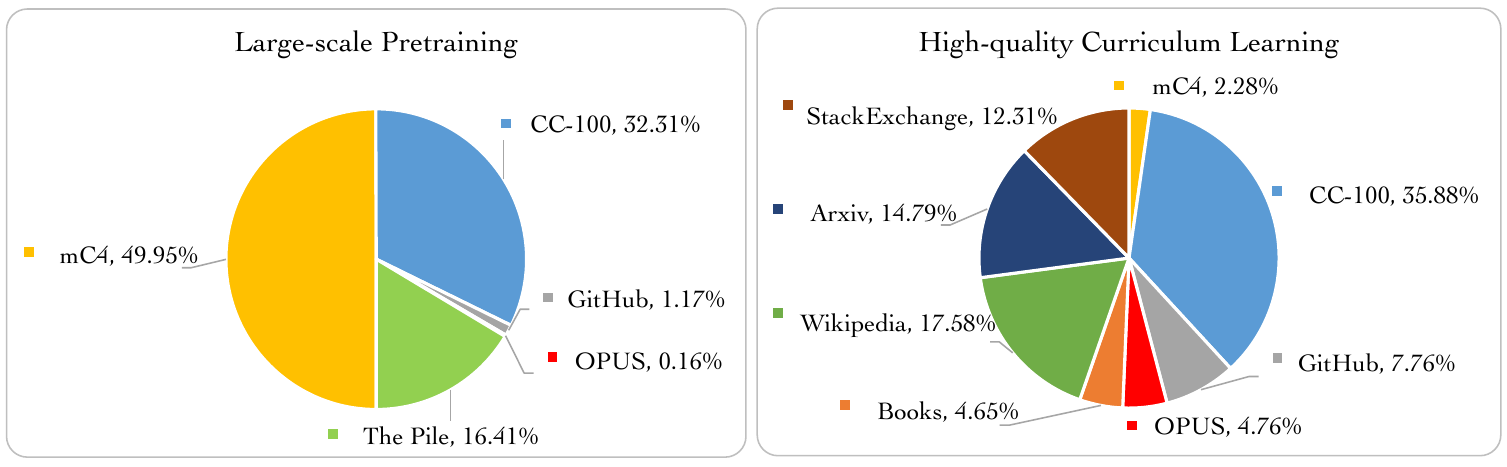}
    \caption{The proportion of high-quality and multilingual source is raised in the curriculum learning dataset.}
    \label{fig:source_porpertion}
\end{figure}

\textbf{Mixed-Precision.} Despite the potential instabilities associated with training models using half precision (\texttt{float16}) activations and model parameters that arise from the limited numerical range, it has been proposed that the numbers represented by \texttt{bfloat16} allow for training of models and can avoid performance degradation compared to full \texttt{float32} training. Thus, we incorporate the \texttt{bfloat16} numerical format to reduce memory and increase training efficiency. However, similar to OPT-175B~\citep{zhang2022opt}, BLOOM-176B~\citep{scao2022bloom} and GLM-130B~\citep{zeng2022glm}, the training of \textsc{Poly}LM-13B still faces frequent loss spikes while lowering learning rate. We attempted to address such challenge via manually skipping data and restart the straining, it unfortunately tends to become increasingly severe as the training does on (Cf. Figure~\ref{fig:sec-2-3-mp-a}).

After conducting two weeks of investigation, we have come to the realization that the instabilities we are encountering may not be due to the training data under the mutlilingual scenario (with the vocabulary up to 256,000), but rather due to the model itself. Specifically, we suspect that there could be a risk of overflow in the attention or residual connectivity layers. Taking this into account, we have configured the residual connection and attention layers to have a numerical precision of float32 to ensure optimal performance, resulting in a highly stable training process (Cf. Figure~\ref{fig:sec-2-3-mp-b}).

\textbf{Curriculum Learning.} Optimizing LLMs to learn knowledge encoded in multiple languages simultaneously is a significant challenge. 
We concretely formulate this problem as transferring general knowledge to low-resource languages while maintaining the advantage of high-resource language in the model. To address this issue, we adopt a curriculum learning strategy~\citep{conf/icml/BengioLCW09,conf/nips/KumarPK10,conf/icml/JaegleGBVZC21} that ramps up the ratio of high-quality and low-resource languages during training. 
Specifically, the training process is divided into two stages. 
In the first stage, we use the whole pre-training dataset to train a base model yields commonsense generalization ability. In the second stage, we transition to a subset of the pre-training dataset that boasts superior quality and a greater proportion of multilingual content, to further strengthen the model's multilingual capabilities. Figure \ref{fig:lang_propertion} compares the language distribution of training data in two stages, indicating that the proportion of most low-resource languages has been increased in the sub-dataset. 

To build the sub-dataset for curriculum learning, we first manually evaluate the quality of publicly available data source in the pre-training dataset, and sample about 97B tokens from the high-quality sources while increasing the proportion of languages other than Chinese and English. We also enhance the proportion of parallel data (OPUS) to facilitate the modeling of cross-lingual representation. The detail of the sub-dataset are illustrated in Figure \ref{fig:source_porpertion}. According to our established setup, the curriculum training process is highly stable (Cf. Figure~\ref{fig:sec-2-3-mp-c}).

%% file: tex/3_self_instruction.tex
\section{\mySFTDatasetName: A Multilingual Self-Instruction Dataset}
\label{sec3}
Fine-tuning LLMs with instruction-based tasks has been proven effective in practice~\citep{ouyang2022training,wei2022finetuned,peng2023instruction,ye2023context}.
By providing accurate task instructions during the SFT phase, LLMs can not only learn to understand the requirements of each task via the instruction part, but also show extensive abilities to cope with other types of tasks which are even unseen during training~\citep{wei2022finetuned}.
Nevertheless, tuning multilingual LLMs is still troubled by the scarcity of current SFT datasets.
On the one hand, most instruction-based datasets are mainly in resource-rich languages (\textit{e.g.}, English or Chinese).
To the best of our knowledge, there is currently no high-quality multilingual instruction-based SFT dataset for LLM training.
On the other hand, most instructions are manufactured by experienced language speakers~\citep[\textit{e.g.,}][]{wei2022finetuned}.
Although the quality of instructions is well preserved, the amount of tasks is rather scarce for fine-tuning LLMs.

\begin{figure}
    \centering
    \includegraphics[width=\textwidth]{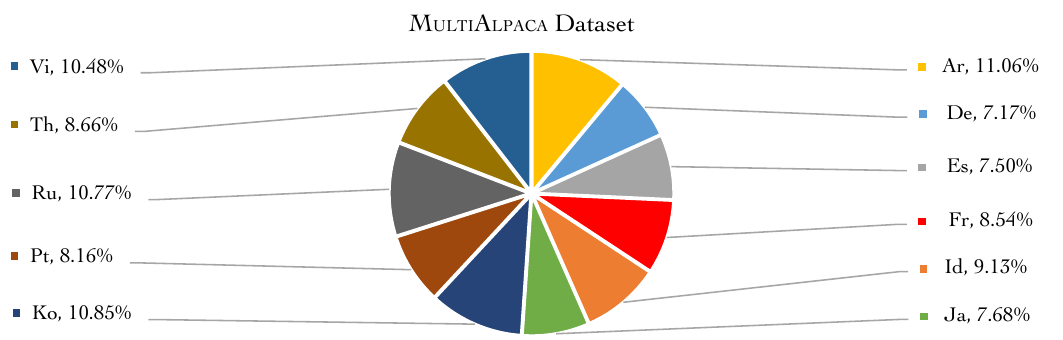}
    \caption{Statistics on the number of self-instruct tasks for each language in \mySFTDatasetName. For English and Chinese subsets, we directly use the released tasks by \textsc{Alpaca}~\cite{taori2023alpaca} and \textsc{Chinese-Alpaca}~\cite{cui2023efficient} for \textsc{PolyLM} training.}
    \label{figure.3.dataset_scale}
\end{figure}

To overcome these two drawbacks, we determine to extend the generality of our proposed \textsc{PolyLM} via creating a multilingual SFT dataset -- \mySFTDatasetNameS(Figure~\ref{figure.3.dataset_scale}).
Following the self-instruct paradigm proposed by recent studies~\citep{wang2022self,taori2023alpaca}, we query the available LLM for responses, iteratively collecting and filtering self-instruct examples to build our dataset.
\mySFTDatasetNameS delivers comprehensive support on multilingualism, covering up to 11 languages including Arabic (Ar), German (De), Spanish (Es), French (Fr), Indonesian (Id), Japanese (Ja), Korean (Ko), Portuguese (Pt), Russian (Ru), Thai (Th), and Vietnamese (Vi).
For each language, the number of tasks in \mySFTDatasetNameS varies from 9,515 to 14,671, yielding 132,701 tasks in total.

\subsection{Task Format}
We first form the format of our tasks by referring to~\cite{taori2023alpaca}, where each task contains three parts:
1) ``\texttt{instruction}'' describes the requirements of the corresponding task;
2) ``\texttt{input}'' can complement the ``\texttt{instruction}'' to a complete question;
and 3) ``\texttt{output}'' is a correct answer of the question. 
We notice that,~\cite{taori2023alpaca} constructed their dataset where each ``\texttt{instruction}'' can be equipped with multiple ``\texttt{input}-\texttt{output}'' instances.
For simplicity, we only assign each ``\texttt{instruction}'' with one ``\texttt{input}-\texttt{output}'' instance.

\subsection{\mySFTDatasetNameS Construction}
As shown in Figure~\ref{figure.3.flowchart}, we construct the \mySFTDatasetNameS dataset based on the following steps:\footnote{See Appendix~\ref{Appendix.MultiAlpacaDataset} for more details.}
\paragraph{Collecting Multilingual Seed Tasks}
\begin{figure}
    \centering
    \includegraphics[width=\columnwidth]{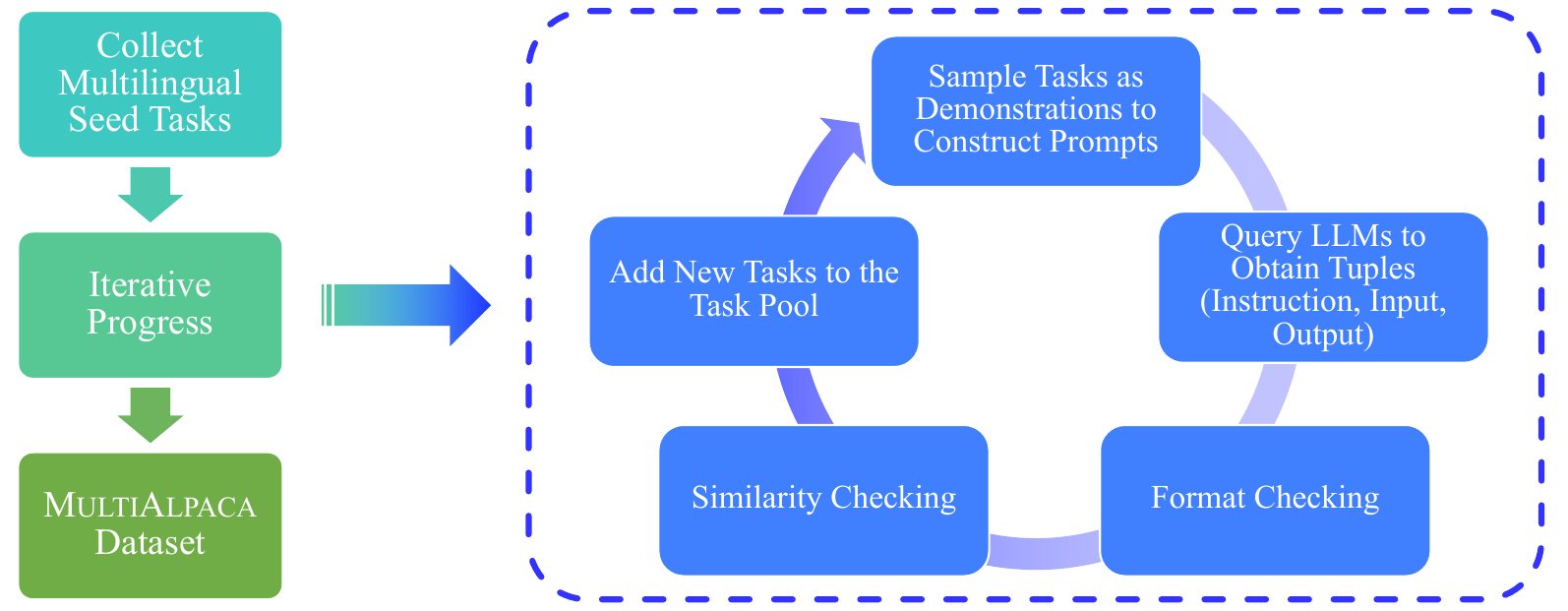}
    \caption{Illustration on the construction of \mySFTDatasetNameS. We first translate English seed tasks from \textsc{Alpaca}~\citep{taori2023alpaca} into multilingual ones. For each language, we arrange iterative progress to collect the \mySFTDatasetNameS dataset, including constructing prompts, collecting tasks via querying LLMs, format checking, filtering on diversity, and adding the new tasks into the task pool.}
    \label{figure.3.flowchart}
\end{figure}

We first obtain 175 seed tasks from~\cite{taori2023alpaca} to construct the multilingual ones for \mySFTDatasetName.
After manually checking them, we remove the cases where answering the questions requires cultural backgrounds (\textit{e.g.}, idiom explanation, character-level riddle, and lyrics generation).
Then, we marked the cases whose original ``\texttt{input}'' or ``\texttt{output}'' should be reserved (\textit{e.g.}, single-choice question, translation, bias identification, and code generation), where those tasks will directly use the original ``\texttt{input}'' or ``\texttt{output}'' across different languages for \mySFTDatasetName.
Finally, we filter out 13 inappropriate seed tasks, and modified 23 ones marked due to the reuse of ``\texttt{input}'' or ``\texttt{output}'' parts.
We translate the remaining 162 tasks into the other 11 languages, yielding multilingual seed tasks for each language.

\paragraph{Iterative Progress}
We manage the \mySFTDatasetNameS dataset construction progress as an iterative one with multiple rounds.
For each round, we manage the following five substeps in order:

\begin{itemize}
    \item \textbf{Prompt Construction} We follow \cite{taori2023alpaca} to construct the prompts for \mySFTDatasetNameS when querying LLM for completion.
    When handling each involved language, for each prompt, we sample two seed tasks and one \mySFTDatasetNameS task as the demonstrations, and guide the LLM to complete the other 17 tasks in the response.
    For each round, we construct 100 prompts for querying the completion by LLM.\footnote{Except for the first round where the task pool is empty, we arrange 10 prompts for completion due to the small number of available tasks for demonstrations.}
    \item \textbf{Response Collection} We collect the responses from \textsc{ChatGPT} via the OpenAI API service. The model we use is ``gpt-3.5-turbo-0301'', which supports the processing of tokens up to 4,096.
    \item \textbf{Format Checking} When checking the format, we first remove the last task if the response is stopped due to the exceeding of max sequence length. Then, we use the pre-defined task format to help split the response string, so as to make sure each of the tasks contains ``\texttt{instruction}'', ``\texttt{input}'', and ``\texttt{output}'' parts.
    \item \textbf{Similarity Checking} After that, to preserve the diversity of \mySFTDatasetNameS dataset, we further check the similarity between the tasks that are newly collected and those from the task pool. Following~\cite{taori2023alpaca}, we compute the Rouge-L F-scores between the instruction of each newly collected task and those of all collected ones.
    For each newly collected task, it would be added to the task pool only if all the scores are lower than 0.7.
    \item \textbf{Task Pool Updating} In the end, we update the task pool by adding the newly collected tasks, and arrange the next round for collecting \mySFTDatasetNameS self-instruct tasks. 
\end{itemize}

\paragraph{\mySFTDatasetNameS Dataset Export}
Totally, we arrange 10 rounds in the iterative progress when constructing the \mySFTDatasetNameS dataset.
We export all tasks from the task pool as the \mySFTDatasetNameS dataset for SFT learning.

%% file: tex/4_benchmark.tex
\section{Multilingual Benchmark}

We aim to assess the capabilities of \textsc{Poly}LM from various perspectives: 1) the ability of large language models (LLMs) to understand and generate natural languages, as well as the ability to grasp world knowledge; 2) the performance of LLMs across different languages; and 3) their capacity to handle cross-lingual tasks. Following the experiment design of previous work~\citep{scao2022bloom, Ahuja2023MEGAME}, we gather a subset of datasets from previous NLP tasks to construct a multilingual benchmark. The brief statistics of all datasets in the benchmark can be found in Table~\ref{tbl:overview-of-evaluation-dataset}. The details of how we frame all the tasks with prompting are listed in Appendix~\ref{appendix_task_formatting}.

\begin{table*}[h]
\centering
\footnotesize
\resizebox{\textwidth}{!}{
\begin{tabular}{llrrrl}\\ 
\toprule
\textbf{Task category} & \textbf{Task} & \textbf{Test} & \textbf{Lang.} & \textbf{Metric} & \textbf{Prompt} \\
\toprule
\multirow{4}{*}{NLU} & XNLI & 5,010 & 15 & Acc.  & \texttt{[Premise]}, right? \{Yes/Also/No\}, \texttt{[Hypothesis]} \\
~& XCOPA & 500 & 11 & Acc. & \texttt{[Prefix]} \{because/therefore\} \{choice1/choice2\} \texttt{[Suffix]} \\
~& PAWS-X & 2,000 & 7       & Acc.   & \texttt{[Sentence1]}, right? \{Yes/No\}, \texttt{[Sentence2]}    \\
~& XWINOGRAD & 83-2,325       & 6 & Acc. & \texttt{[Prefix]} \{choice1/choice2\} \texttt{[Suffix]} \\
\midrule
Knowledge & TydiQA & 1,625-14,805        & 9      & F1 & \texttt{[Context]}\texttt{[Question]}\texttt{[Answer]} \\
\midrule
NLG & MTG & 200        & 5     & Rouge & \texttt{[Prompt]}\texttt{[Input]}\texttt{[Output]} \\
\midrule
MT & WMT20 & 991-3,002        & 8     & BLEU & \texttt{[INPUT]} Translate this sentence from \texttt{[SRC]} to \texttt{[TGT]}. \\
\bottomrule
\end{tabular}
}
\caption{Multilingual Benchmark}
\label{tbl:overview-of-evaluation-dataset}
\end{table*}

\subsection{Tasks in Benchmark}

All the datasets in the above multilingual benchmark can be divided into four groups: Natural Language Understanding, Knowledge, Natural Language Generation and Machine Translation. The details of each dataset that we use for benchmarking are given below.

\label{sec4}

To assess the comprehension capability of large models across various languages, we collect the multilingual versions of datasets from seberal wide-used NLP benchmarks~\citep{Wang2018GLUEAM,Wang2019SuperGLUEAS}. 

\textbf{XNLI} \citep{Conneau2019UnsupervisedCR} serves as a benchmark to evaluate a model's proficiency in predicting textual entailment. The task entails the evaluation of whether two given sentences, A and B, convey the same meaning, are contradictory, or are unrelated. The dataset has been professionally translated into 14 languages from the original English XNLI dataset.

\textbf{PAWS-X} \citep{Yang2019PAWSXAC} is a benchmark to evaluate the model's ability to judge whether one sentence is the paraphrase of another. It is professionally translated from the PAWS~\citep{paws2019naacl} dataset into 6 diverse languages.

\textbf{XWinograd} \citep{tikhonov2021heads} serves as a benchmark to measure a model's common sense reasoning ability. Specifically, the task entails presenting the model with a brief contextual passage and requiring it to select the accurate term from a set of two options for a pronoun in the passage.

\textbf{XCOPA} \citep{ponti2020xcopa} is another benchmark intended to assess the proficiency of models in commonsense reasoning across languages. The dataset comprises translations and re-annotations of the English COPA~\cite{Gordon2011SemEval2012T7}, spanning 11 languages around the globe. Based on the given premise and prompt, the task is to choose the more plausible response between two answer choices that can be inferred from the premise.

\textbf{TyDi QA}~\citep{tydiqa} is a question-answering dataset covering 11 typologically diverse languages with 200K question-answer pairs. We use this dataset to evaluate the ability to grasp knowledge from natural text. Unlike previous datasets such as MLQA~\citep{MLQA} and MKQA~\citep{Longpre2020MKQAAL},  this dataset is collected directly in each language without the use of translation. We select 5 languages out of 11 that are included in the pretraining corpora of \textsc{Poly}LM. Following the PaLM ~\citep{Chowdhery2022PaLMSL}, we evaluate models on the Gold passage task, which requires answering questions based on a passage that is guaranteed to contain the answer.

\textbf{MTG}~\citep{Chen2021MTGAB} is used to assess the efficacy of large language models in generating longer responses across diverse usage scenarios and multiple languages. MTG covers four different generation tasks: Story Ending Generation (SG), Title Generation (TG), Question Generation (QG), and Summarization (Summ). The datasets are originally written in English, subsequently extended into four other languages (German, French, Spanish, and Chinese) through the use of machine translation and human annotation. The effectiveness of LLM-generated responses is evaluated using the average of Rouge1, Rouge2, and RougeL.

\textbf{WMT20}~\citep{barrault-etal-2020-findings} is used to study the cross-lingual proficiency of large language models in accomplishing translation tasks, as the process of translation entails both comprehending the semantic of the input in one language and expressing it in another. We select translation tasks between English and each of the following languages as benchmark languages: German, Japanese, Russian, and Chinese. The results are evaluated using the SacreBLEU~\citep{post-2018-call} and the scores for BLEU~\citep{Papineni2002BleuAM} on the test set are reported.

\subsection{Evaluation Design}

For metric evaluation, the tasks included in our benchmark can be divided into two categories: classification-style tasks and generation-style tasks.

Classification-style tasks require selecting the correct option from several options, such as the XNLI dataset.
To evaluate these tasks, following the way in~\cite{eval-harness}, we design the problem in the form of a cloze test, where each option is filled in to construct a complete sentence. We then choose the correct answer by separately calculating the log-likelihood of each completed sentence and selecting the one with the highest value.

Generation-style tasks, such as machine translation, require generating answers with several natural sentences. For these tasks, we adopt greedy decoding for deterministic results. Considering the efficiency of decoding, we restrict the maximum number of generated tokens to 256. For foundation models, we choose the result before the first `\textbackslash n' as the answer, while for models that have undergone instruction tuning, we decode until the EOS token appears.

In evaluating foundation models, considering that models have not been able to understand instructions, we adopt in-context learning~\citep{gpt3} to evaluate the model for generation-style tasks. We generally choose no more than five examples due to the model's context window limitation. For tasks that have well-divided training/development sets, we randomly draw examples from them for each test sample. Otherwise, we draw examples randomly from the test sets except for the current sample.

%% file: tex/5_experiments.tex
\section{Experiments}
\label{sec5}

In this section, we provide separate comparison results for the pre-training and SFT models. Then, we analyze the effectiveness of our model in three aspects: curriculum learning, multilingual instruction finetuning, and the scaling for model size.

\begin{figure}[t]
    \centering
    \includegraphics[width=0.9\columnwidth]{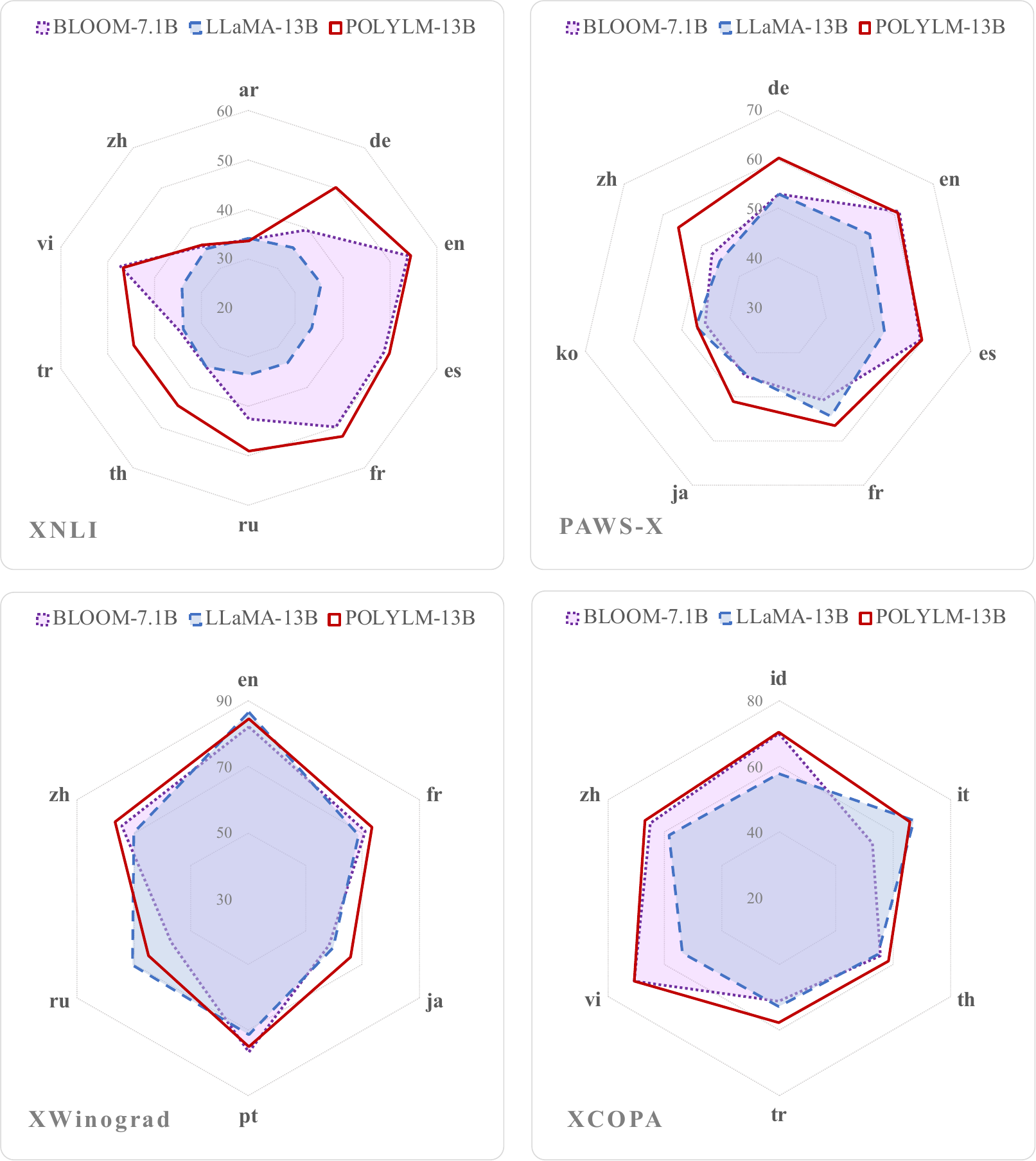}
    \caption{Accuracy of NLU tasks under the zero-shot setting. Best reviewed in colors. Results indicate that \textsc{PolyLM} performs comparably or better than LLaMA-13B in the English scenario, and exhibits significant enhancements in multilingual evaluation.
}
    \label{fig:results_nlu}
\end{figure}

\subsection{Comparisons between Pre-trained Foundational Models}
\label{sec5-2}
For the pre-trained models, we selected two mainstream open-source models as our baselines.
\begin{itemize}
    \item LLaMA~\citep{touvron2023llama} is a pre-trained language model released by MetaAI, which includes 7B, 13B, 30B, and 65B versions. The pre-training dataset is sourced from publicly available corpora. The 33B and 65B models are trained on 1.4 T tokens, while the 7B and 13B models are trained on 1 T tokens. To ensure an equal parameter count comparison with \textsc{PolyLM}, we mainly take the 13B version into consideration. 
    \item BLOOM~\citep{scao2022bloom} is a multilingual model that covers 46 natural languages and 13 programming languages with a maximum of 176B parameters. Since BLOOM has not released a 13B version, we opt for the BLOOM-7.1B model as our baseline.
\end{itemize}

We evaluate \textsc{PolyLM} across various multilingual tasks, covering natural language understanding (NLU), knowledge, natural language generation (NLG) and machine translation (MT). To make a clearer comparison of the multilingual capabilities of different models, we present the results using radar charts, with detailed results available in the \ref{Appendix.results}.

\begin{figure*}[t]
    \centering
    \begin{subfigure}[b]{0.45\textwidth}
        \centering
        \includegraphics[width=\textwidth]{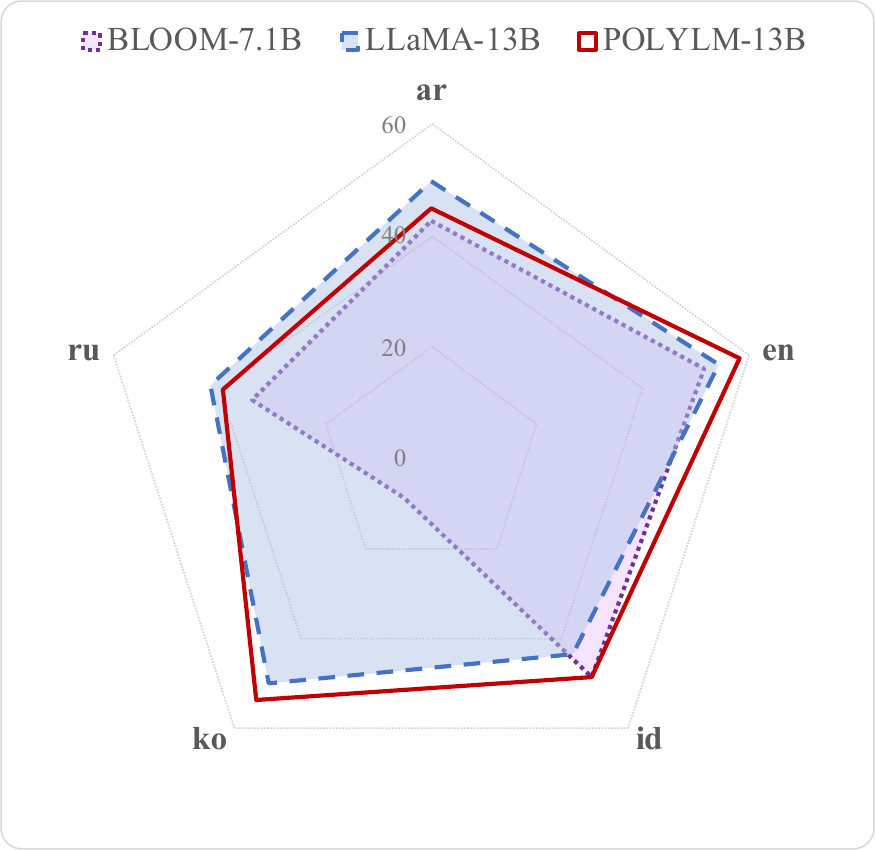}
        \caption{F1 Scores on TyDiQA. }
        \label{fig:results_pretrain_tydiqa}
    \end{subfigure}
    \begin{subfigure}[b]{0.45\textwidth}
        \centering
        \includegraphics[width=\textwidth]{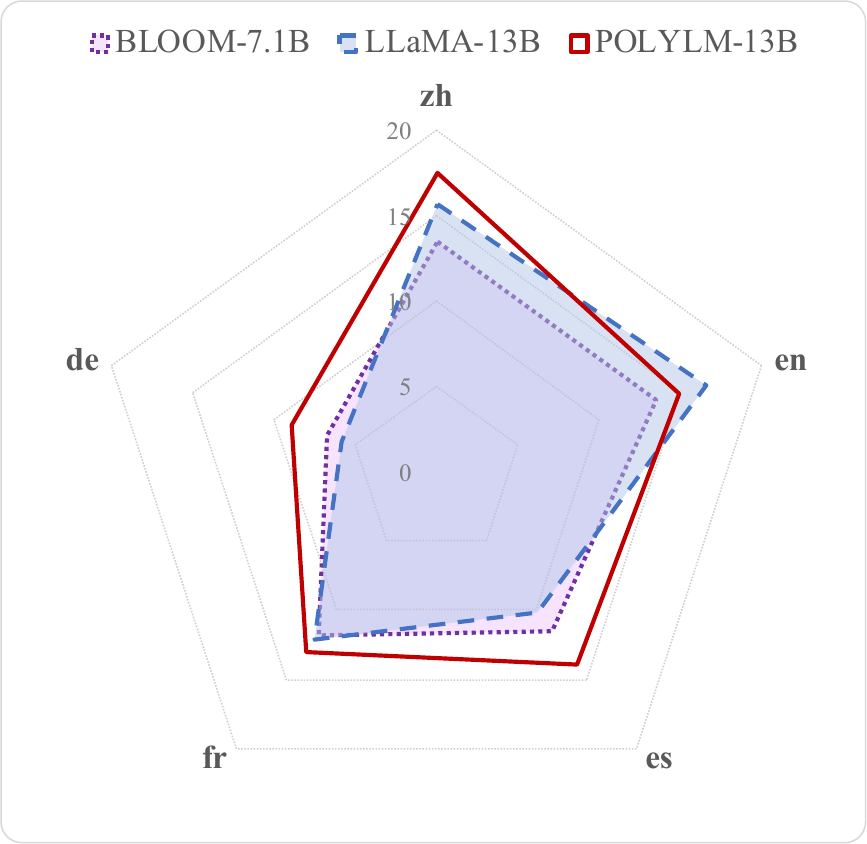}
        \caption{Average Rouge Scores on MTG. }
        \label{fig:results_pretrain_mtg}
    \end{subfigure}
    \begin{subfigure}[b]{0.45\textwidth}
        \centering
        \includegraphics[width=\textwidth]{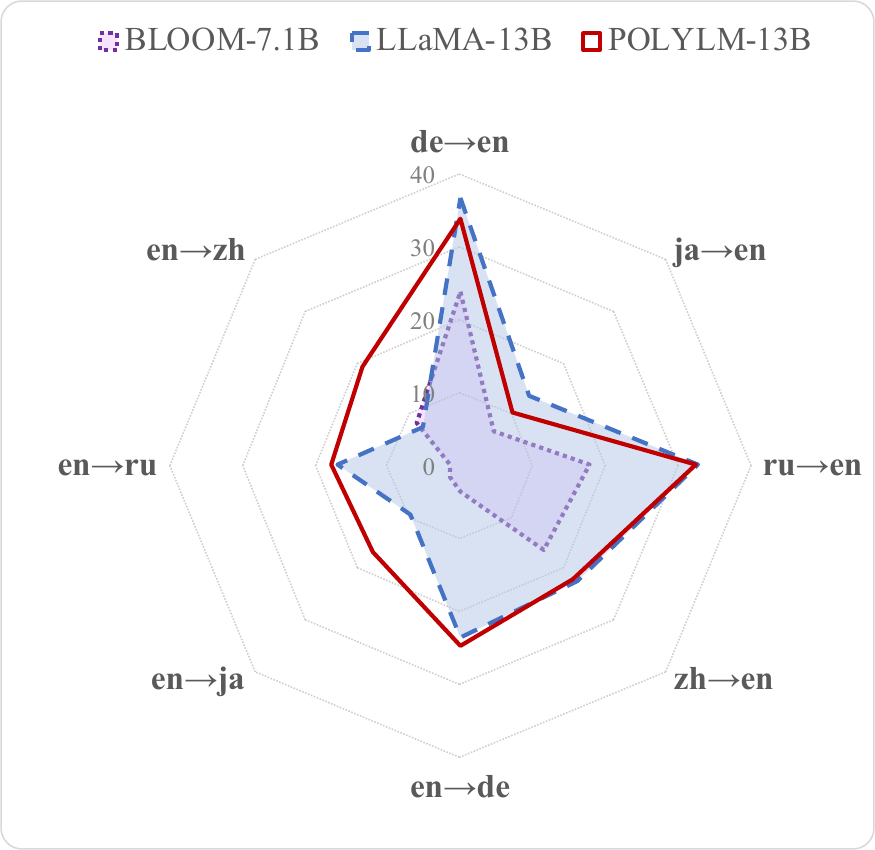}
        \caption{BLEU Scores on WMT20.}
        \label{fig:results_pretrain_mt}
    \end{subfigure}
    \caption{Performance on knowledge, neural language generation and machine translation tasks under the one-shot setting. Best reviewed in colors.}
\end{figure*}

\begin{figure}[h!]
    \centering
    \includegraphics[width=0.9\columnwidth]{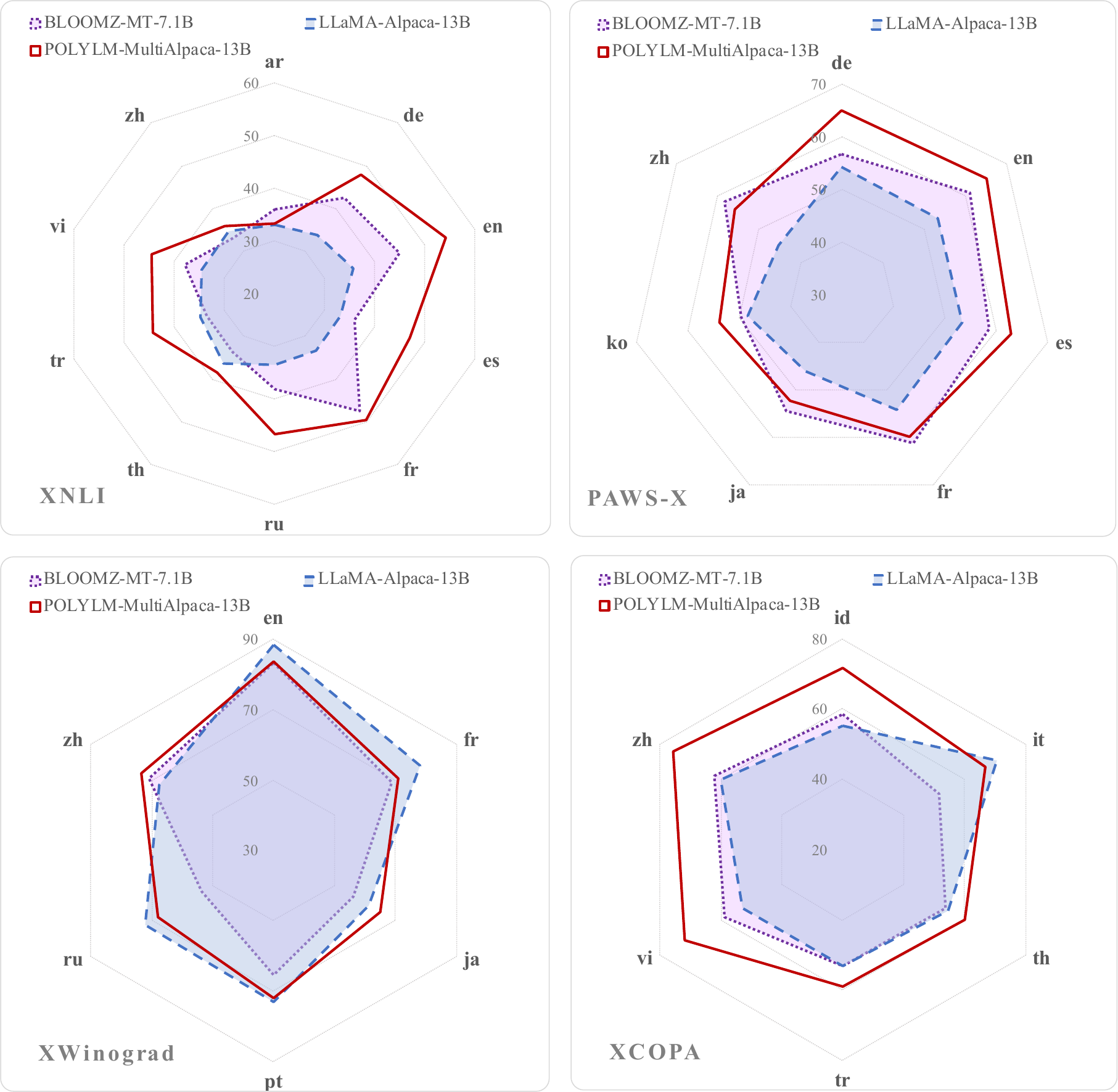}
    \caption{Performance of instruction-followed models on NLU tasks under the zero-shot setting. Best reviewed in colors.}
    \label{fig:results_sft_nlu}
\end{figure}

\begin{figure*}[ht!]
    \centering
    \begin{subfigure}[b]{0.45\textwidth}
        \centering
        \includegraphics[width=\textwidth]{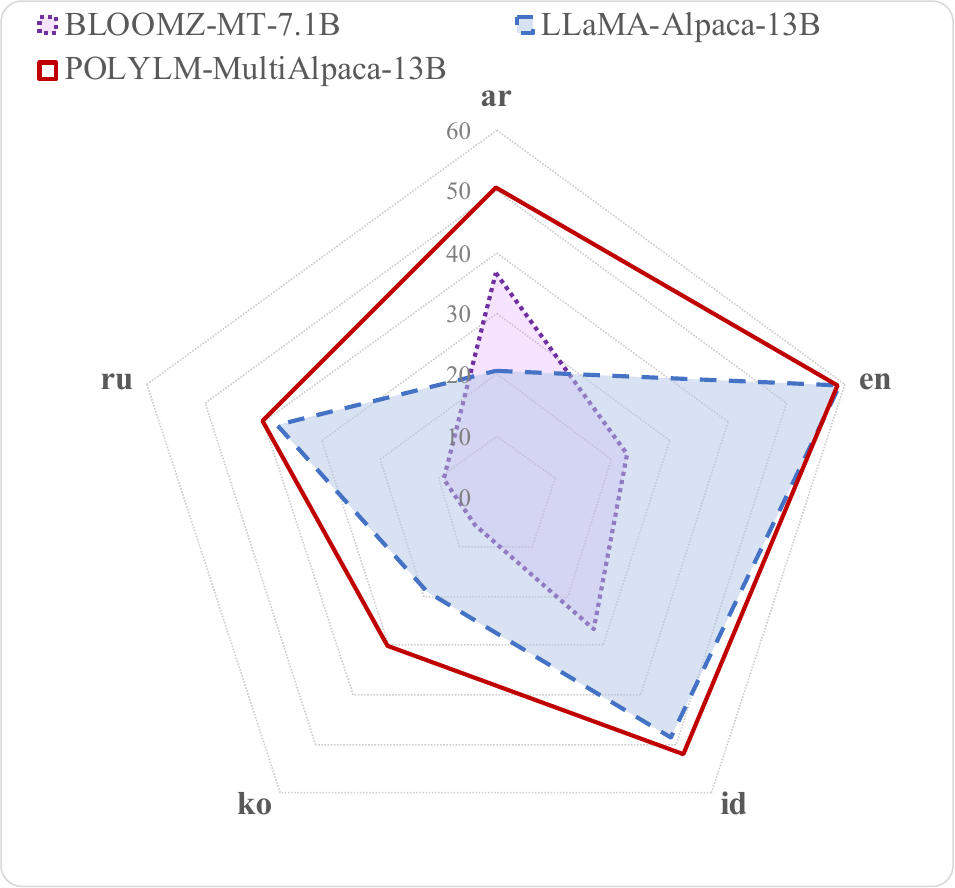}
        \caption{F1 Scores on TyDiQA. }
        \label{fig:results_sft_tydiqa}
    \end{subfigure}
    \begin{subfigure}[b]{0.45\textwidth}
        \centering
        \includegraphics[width=\textwidth]{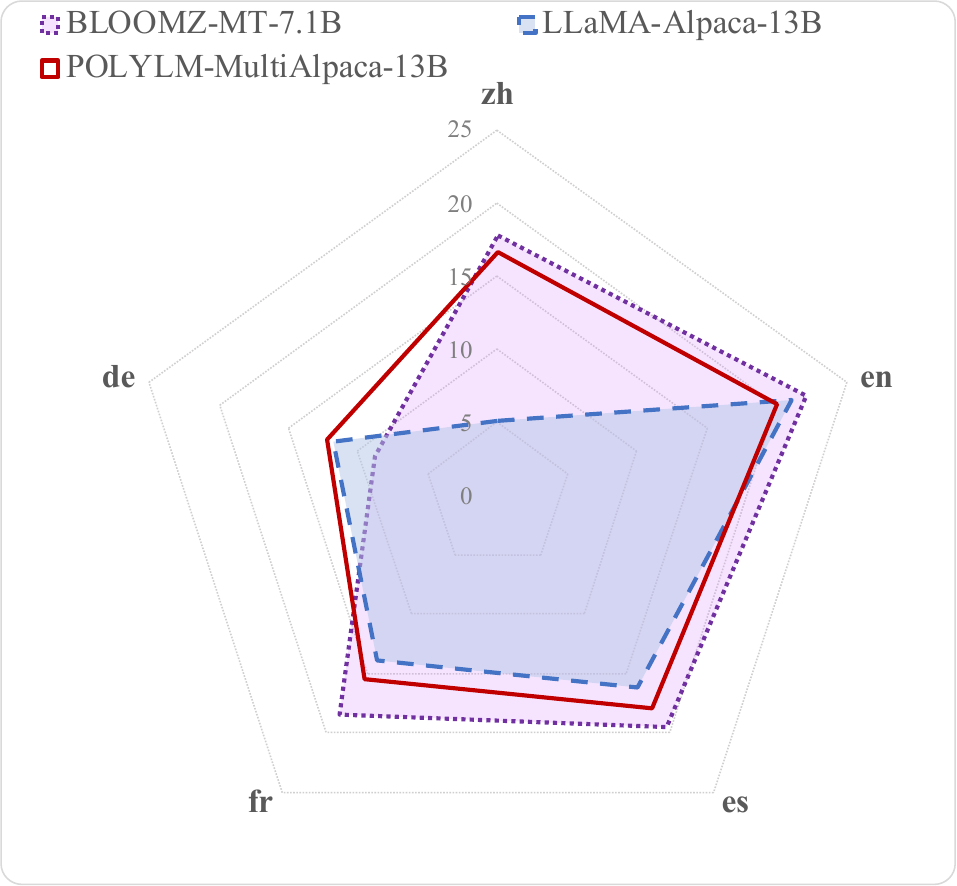}
        \caption{Average Rouge Scores on MTG. }
        \label{fig:results_sft_mtg}
    \end{subfigure}
    \begin{subfigure}[b]{0.45\textwidth}
        \centering
        \includegraphics[width=\textwidth]{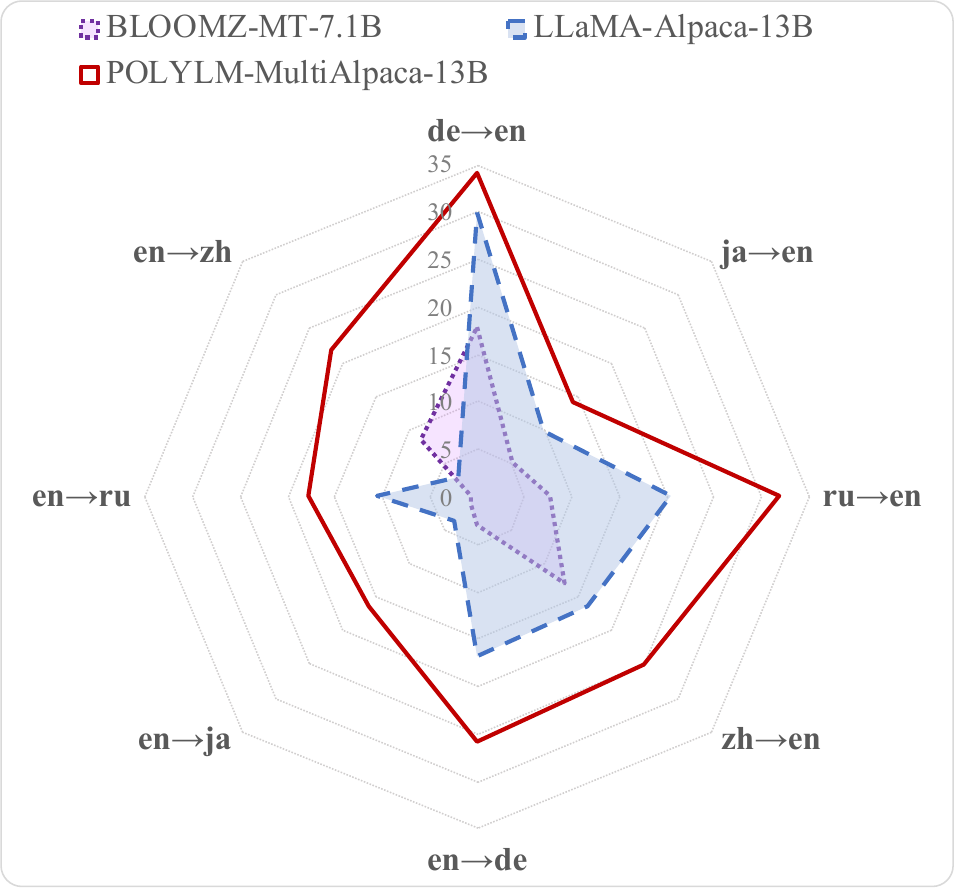}
        \caption{BLEU Scores on WMT20.}
        \label{fig:results_sft_mt}
    \end{subfigure}
    \caption{Performance of instruction-followed models on knowledge, neural language generation and machine translation tasks under the zero-shot setting. Best reviewed in colors.}
    \label{fig:results_sft}
\end{figure*}

\textbf{Natural Language Understanding.} Figure~\ref{fig:results_nlu} shows the results on four NLU tasks under the zero-shot setting. \textsc{Poly}LM-13B shows comparable performance to the English-centric LLaMA-13B model in the English scenario. Moreover, it yields substantial improvements of 7.2\% and 19.1\% on PAWS-X and XNLI respectively. For languages other than English (the multilingual column), \textsc{Poly}LM-13B outperforms LLaMA-13B with average improvement up to 7.6\%, 5.6\%, 3\%, and 11\% on XCOPA, PAWS-X, XWinagrad, and XNLI, respectively. When compared to the multilingual language model BLOOM-7.1B, \textsc{PolyLM}-13B outperforms with an average improvement of 4.2\%, 4.1\%, 3.4\%, and 4\% points on the respective tasks. This improvement can be attributed to the higher percent of multilingual text during pre-training and curriculum learning strategy.

\textbf{Knowledge.} We evaluate our model on grasping multilingual knowledge by using the TyDiQA benchmark in the one-shot setting. Upon careful analysis of Figure~\ref{fig:results_pretrain_tydiqa}, it is evident that BLOOM-7.1B experiences significant performance drops in the Korean (ko) and Russian (ru) language directions, whereas LLaMA-13B and \textsc{Poly}LM-13B exhibit better balance across all five languages.
Furthermore, \textsc{Poly}LM-13B has an additional advantage of an average 1.2-point lead over LLaMA-13B.

\textbf{Natural Language Generation.} Figure~\ref{fig:results_pretrain_mtg} displays the Rouge scores of four diverse NLG tasks in multilingual settings. From a multilingual perspective, \textsc{Poly}LM-13B outperforms all other models across four languages, namely Chinese (zh), Spanish (es), French (fr), and German (de). Moreover, in terms of task types, \textsc{Poly}LM-13B performs the best in question generation (QG) and summarization (Sum) tasks, while also showing comparable performance to the best model LLaMA-13B in the text generation (TG) task.
Across all MTG tasks and languages, \textsc{Poly}LM-13B has an average score advantage of 1.6 and 2.3 compared to LLaMA-13B and BLOOM-7.1B, respectively.

\textbf{Machine Translation}
We focus on evaluating the translation performance on four typologically diverse
languages from WMT20 datasets, including translation directions both from and to English. Results of Figure~\ref{fig:results_pretrain_mt} show that \textsc{Poly}LM-13B achieves similar performance to LLaMA-13B in the multilingual to English directions and surpasses LLaMA-13B and BLOOM-7.1B with average BLEU scores of 5.4 and 15.8 in the English to multilingual directions.

\subsection{Comparisons between Instruction-followed Models}
\label{sec5-3}

This section focuses on evaluating the effectiveness of instruction-followed models founded on the pre-trained language models discussed in Section~\ref{sec5-2}. We conduct a comparative analysis of \textsc{Poly}LM-\mySFTDatasetName-13B that is fine-tuned on \textsc{Poly}LM-13B using \mySFTDatasetName, against two other publicly available models:
\begin{itemize}
    \item BLOOMZ-MT-7B is initially pre-trained on BLOOM-7B, and later fine-tuned on the multilingual task mixture xP3-MT~\citep{muennighoff2022crosslingual}.
    \item LLaMA-Alpaca-13B is built based on the pre-trained model LLaMA-13B and fine-tuned on the English self-instruction dataset \textsc{Alpaca}~\citep{taori2023alpaca}.
\end{itemize}

Figure~\ref{fig:results_sft_nlu} and \ref{fig:results_sft} present the performance comparisons of instruction-followed models with the zero-shot setting, considering various tasks and language directions. The results indicate that \textsc{Poly}LM-\mySFTDatasetName-13B is comparable or superior to LLaMA-Alpaca-13B on all English tasks, although the latter is primarily trained on English-only instructions. On other non-English tasks, \textsc{Poly}LM-\mySFTDatasetName-13B significantly outperforms LLaMA-Alpaca-13B. This superiority can be attributed to the inclusion of more well-balanced multilingual datasets during the pre-training and instruction fine-tuning. In comparison to BLOOMZ-MT-7B, \textsc{Poly}LM-\mySFTDatasetName-13B has demonstrated consistent improvements across all tasks and languages. We have observed an outlier MTG, and we speculate that this may be due to the fact that MTG testsets are part of the xP3 dataset. We plan to refine our instruction tuning process for \textsc{Poly}LM by utilizing the xP3 dataset in order to delve deeper into this inconsistency.

Note that it is not feasible to fully assess the effectiveness of the model's performance through downstream NLP tasks after instruction fine-tuning. Therefore, we have presented selected examples for qualitative analysis, which are fully outlined in Appendix~\ref{appendix:demo}.

\begin{figure}
    \centering
    \includegraphics[width=0.9\linewidth]{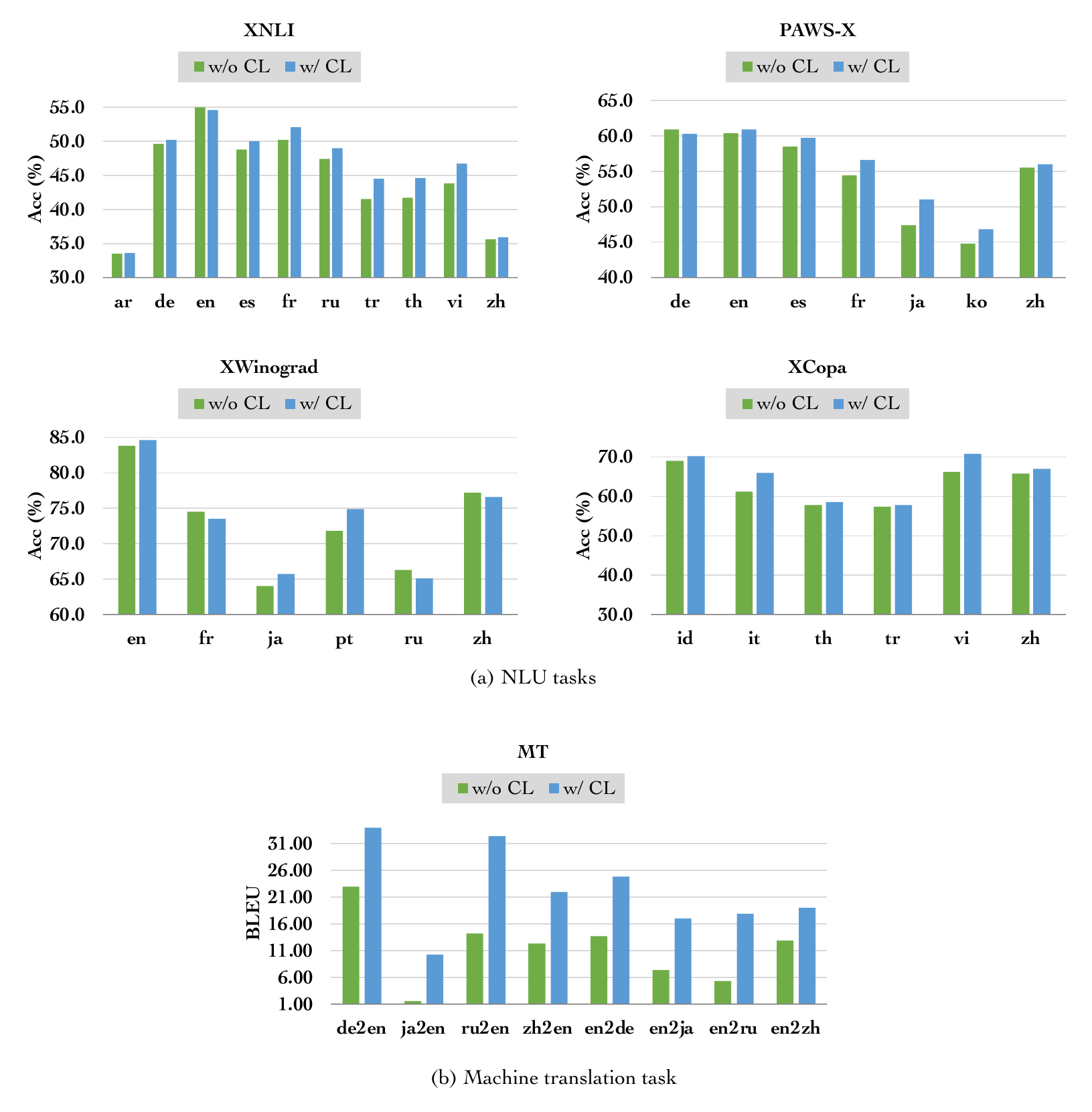}
    \caption{\textsc{Poly}LM-13B trained with curriculum learning reveals better performance in multiple languages in NLU and MT tasks.}
    \label{fig:ct}
\end{figure}

\subsection{Analysis}\label{sec:ablation}

\paragraph{Curriculum Learning.}
We validate the effectiveness of the curriculum learning strategy in NLU and MT tasks of multilingual benchmark (Section \ref{sec4}) by comparing the following variants:

\textbf{(1) w/o CL} \textsc{Poly}LM-13B trained without curriculum learning, which is only optimized in pretrained dataset.

\textbf{(2) w/ CL} \textsc{Poly}LM-13B trained with curriculum learning, using about 100B high-quality multilingual data selected from the pretrained dataset.

Please note that we only focus on the languages included during curriculum learning. Referring to Figure \ref{fig:ct}, the model with curriculum learning has achieved stable progress in mainly all languages in both NLU and MT tasks. 
First of all, the model performance is enhanced in most low-resource languages, indicating that the general knowledge can be effectively transferred to these languages through raising data proportion. 
Additionally, the model retains its superior performance in English, which illustrates that improving data quality for high-resource languages can achieve competitive results to training with larger amounts of data. 
Finally, it is worth noting that introducing more multilingual parallel data during the curriculum learning significantly boost the model performance on translation task.

\begin{figure}
    \centering
    \includegraphics[width=0.9\linewidth]{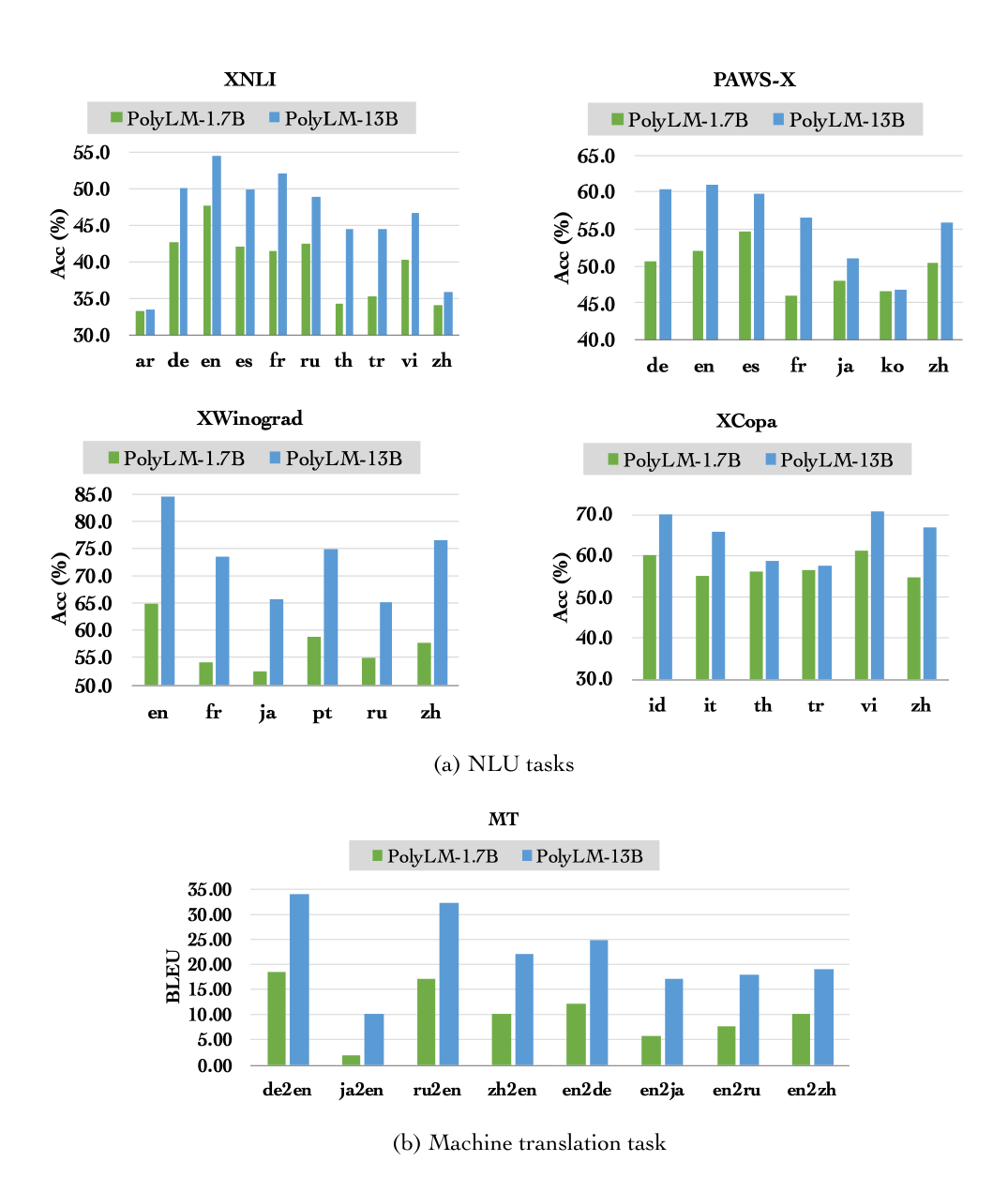}
    \caption{The performance of models with different scales on different tasks.}
    \label{fig:poly_as}
\end{figure}

\begin{table}[t]
\begin{center}
\resizebox{\textwidth}{!}{
\begin{tabular}{l|lllllllll|lllllll}
\toprule
\multirow{2}{*}{\textbf{Model}} & \multicolumn{9}{c|}{WMT20 Machine Translation} & \multicolumn{6}{c}{TyDiQA}\\
\cmidrule{2-16}
~ & en2zh   & en2de            & en2ru   & en2ja   & zh2en            & de2en   &ru2en    &ja2en   & Avg. & en   & ar            & id            & ko   & ru            & Avg.       \\ 
\midrule
\textsc{Poly}LM-Alpaca-13B & 10.0 & 17.1  & 5.0 & 6.3 & 12.2 & 23.8 & 24.7    & 10.0 & 13.6  & 53.7 & 45.5 & \textbf{55.3} & 29.9 & \textbf{40.4} & 45.0 \\
\textsc{Poly}LM-MultiAlpaca-13B & \textbf{21.9} & \textbf{25.9} & \textbf{17.9} & \textbf{16.2} & \textbf{25.0} & \textbf{34.2} & \textbf{31.8} & \textbf{14.3} & \textbf{23.4} & \textbf{58.7} & \textbf{50.7} & 52.1 & \textbf{30.1} & 40.3 & \textbf{46.4} \\
\bottomrule
\end{tabular}
}
\caption{BLEU scores on WMT20 machine translation tasks and F1 scores on Tydi-QA task.
}
\label{tab:ab_2nlg}
\end{center}
\end{table}

\paragraph{Multilingual Self-instruction.}
Here we highlight the advantages of \mySFTDatasetNameS over English-only \textsc{Alpaca}~\citep{taori2023alpaca}, particularly in cross-lingual tasks (i.e., machine translation). As illustrated in Table~\ref{tab:ab_2nlg}, compared to the model fine-tuned only using \textsc{Alpaca}, \textsc{Poly}LM-\mySFTDatasetName-13B exhibits substantial improvements in TyDiQA and multiple WMT20 translation tasks, with enhancements of +10 BLEU and +1.4\% F1. These results suggest that \mySFTDatasetNameS is capable of simulating the cross-lingual alignment ability of the foundational, as well as facilitating the comprehension of multilingual instructions.

\paragraph{Scaling for Model Size.}
In addition to the 13B model, we also release a smaller 1.7B model. Recent studies highlight the critical role of model size in the performance of large language models (LLMs), with much of this work focusing on English~\citep{kaplan2020scaling,rae2021scaling,biderman2023pythia,touvron2023llama}. In this section, we present results for \textsc{Poly}LM-13B and \textsc{Poly}LM-1.7B to investigate the impact of model size on multilingual abilities. Consistent with the aforementioned experimental setup for the validation of base model, we compare the two models using a one-shot setting. As illustrated in Figure \ref{fig:poly_as}, the 13B model significantly outperforms the 1.7B model across all compared multilingual tasks. We posit that multilingual problems are more complex than their monolingual counterparts and may depend more heavily on the model's throughput. Moving forward, we plan to release additional models of varying sizes, with the ultimate goal of refining the scaling law for multilingualism.

%% file: tex/7_conclusion.tex
\section{Conclusion}

Multilingualism poses an inevitable challenge for LLM due to the scarcity of resources. In this work, we release \textsc{Poly}LM -- a new multilingual LLM, alone with \mySFTDatasetNameS -- a multilingual instruction dataset, and a multilingual benchmark. Quantitative and qualitative analyses demonstrate the superiority of \textsc{Poly}LM over open-source models in non-English languages. We find that incorporating curriculum learning strategy can boost the performance of LLM on non-English languages, without impeding its English proficiency. In addition, fine-tuning LLM with multilingual instruction data can considerably improve zero-shot performance on these languages.

There is still ample opportunity for refinement in our work endeavors. For instance, while we briefly assess the model's capacity to comprehend multilingual instructions, there is potential for further optimization through the amalgamation of data sources~\citep{wang2023far,longpre2023flan}, evolutionary methods~\citep{xu2023wizardlm} and diversification strategies~\citep{zhou2023lima}. Moreover, in our current version, we adopt absolute position encoding, which adheres to the early default configuration in Megatron toolkit~\citep{shoeybi2020megatronlm}. Future iterations should incorporate techniques that facilitate the expansion of window size, such as rotary position encoding~\citep{su2021roformer,chen2023extending} or ALiBi~\citep{press2022alibi}.

Language serves as a conduit for culture, and the unique contributions of various languages enrich and diversify our global community. Nevertheless, the advancement of LLM may inadvertently amplify the influence of prominent languages and present a formidable obstacle for low-resource languages.  In light of these concerns, we aspire that our research will motivate further inquiry and innovation in the field of multilingual LLM.

\section*{Ethics Statement}
In this paper, we propose \textsc{PolyLM}, an LLM which offers a wider support on non-English languages.
Our contributions are fully methodological: adding the support of multilingualism to LLM during training and SFT phases.
However, when building our \textsc{PolyLM} model, it is unavoidable that our \textsc{PolyLM} might exhibit several common deficiencies of language models, \textit{e.g.}, hallucination and toxicity.
Specifically, as the collected \mySFTDatasetNameS dataset are generated by \textsc{ChatGPT}, the pseudo tasks might give inappropriate pseudo tasks which are hardly filtered out, \textit{e.g.}, hallucinated reasoning and anti-fact statements~\citep{gpt3,openai2023gpt4}.
Besides, \textsc{PolyLM} may deliver toxic texts, which might be gender- or race-biased like other existing LLMs~\citep{taori2023alpaca,cui2023efficient}.

Despite the ethical concerns above, we think that those problems are of vital importance to the AI community to study the deficiencies of LLMs.
We recommend that the users of \textsc{PolyLM} and \mySFTDatasetNameS deploy our released materials only for research proposals.
Besides, we suggest the users better identify the deficiencies of those contents, and welcome the following researchers to facilitate further research on the alignment between the LLM outputs and human values with \textsc{PolyLM} and \mySFTDatasetNameS materials.

%% file: tex/appendix.tex
\appendix

\section{Detailed Setting for \mySFTDatasetName  Dataset Construction}
\label{Appendix.MultiAlpacaDataset}

\subsection{Prompt for \mySFTDatasetNameS Dataset Construction}

We show the used prompt when constructing \mySFTDatasetNameS dataset in Table~\ref{table.poly_alpaca.sft.prompt}.
We mainly refer to \cite{taori2023alpaca}, and adopt our prompt to multilingual scenarios after minor revisions.
Briefly, in the prompt, we list several requirements of the self-instruct tasks in the prompt, \textit{i.e.}, the used language, the format, the diversity, and the lengths of tasks within each single response.
We also add three demonstrations to help the model generate the tasks which follow the pre-defined format.

\begin{table}[t]
    \centering
    \begin{tabular}{l}
        \toprule
            \textbf{Prompt template for \mySFTDatasetNameS dataset construction.}\\
        \midrule
            \begin{minipage}[t]{0.95\linewidth}
                You are asked to come up with a set of 20 diverse task instructions. These task instructions will be given to a GPT model and we will evaluate the GPT model for completing the instructions. \\
                \\
                Here are the requirements: \\
                1. Try not to repeat the verb for each instruction to maximize diversity.\\
                2. The language used for the instruction also should be diverse. For example, you should combine questions with imperative instructions.\\
                3. The type of instructions should be diverse. The list should include diverse types of tasks like open-ended generation, classification, editing, etc.\\
                4. A GPT language model should be able to complete the instruction. For example, do not ask the assistant to create any visual or audio output. For another example, do not ask the assistant to wake you up at 5pm or set a reminder because it cannot perform any action.\\
                5. The instructions should be in \texttt{[language]}.\\
                6. The instructions should be 1 to 2 sentences long. Either an imperative sentence or a question is permitted.\\
                7. You should generate an appropriate input to the instruction. The input field should contain a specific example provided for the instruction. It should involve realistic data and should not contain simple placeholders. The input should provide substantial content to make the instruction challenging but should ideally not exceed 100 words.\\
                8. Not all instructions require input. For example, when an instruction asks about some general information, ``what is the highest peak in the world'', it is not necessary to provide a specific context. In this case, we simply put ``<noinput>'' in the input field.\\
                9. The output should be an appropriate response to the instruction and the input. Make sure the output is less than 200 words.\\
                \\
                There are 3 examples.\\
                \\
                1. Instruction: \texttt{[task1\_instruction]}\\
                1. Input:\\
                \texttt{[task1\_input]}\\
                1. Output:\\
                \texttt{[task1\_output]}\\
                \\
                2. Instruction: \texttt{[task2\_instruction]}\\
                2. Input:\\
                \texttt{[task2\_input]}\\
                2. Output:\\
                \texttt{[task2\_output]}\\
                \\
                3. Instruction: \texttt{[task3\_instruction]}\\
                3. Input:\\
                \texttt{[task3\_input]}\\
                3. Output:\\
                \texttt{[task3\_output]}\\
                \\
                Please generate the following 17 tasks that are similar to the above examples.
            \end{minipage} \\

        \bottomrule
    \end{tabular}
    \caption{Prompt for constructing \mySFTDatasetNameS tasks. We specify the language of generated tasks with the parameter ``\texttt{language}'', and the used demonstrations with ``\texttt{task[123]\_\{instruction,input,output\}}''.}
    \label{table.poly_alpaca.sft.prompt}
\end{table}

\subsection{Format and Similarity Checking}

After collecting the pseudo tasks for each language, we first remove the cases which contain website links.
Then, we tokenize the ``\texttt{instruction}'', ``\texttt{input}'', and ``\texttt{output}'' with available tokenization toolkits.\footnote{\texttt{nltk.tokenize.wordpunct\_tokenize} for Ar, De, Es, Fr, Id, Pt, Ru, and Vi; \texttt{kytea} for Ja; \texttt{Okt} for Ko; \texttt{thai\_tokenizer} for Th.}

Besides, we found that some of the tasks may give redundant information within both ``\texttt{instruction}'' and ``\texttt{input}'' part.
We revise those examples to make them available as much as possible.
In detail, if the ``\texttt{input}'' part can be a sub-string of the ``\texttt{instruction}'', we mark the ``\texttt{input}'' part as an empty one (using the placeholder ``\texttt{<noinput>}'').
Otherwise, we compute the Rouge-L F-score between the ``\texttt{instruction}'' and the ``\texttt{input}'' part, filtering out the tasks whose result is over 0.5.
Specially, for the Ko, Vi, and Ar tasks, we determine the threshold of Rouge-L F-score as 0.3, 0.3, and 0.2 for higher diversity of tasks, respectively.

\begin{figure}
    \centering
    \begin{tikzpicture}
        \pgfplotsset{set layers}
        \pgfplotsset{every x tick label/.append style={font=\small}}
        \pgfplotsset{every y tick label/.append style={font=\small}}
        \begin{axis}[
                height=0.5 * \columnwidth,
                width=0.8 * \columnwidth,
                xmin=0, xmax=15000,
                ymin=0.4, ymax=1,
                xtick={0, 5000, 10000, 15000},
                xticklabels={$0$, $5000$, $10000$, $15000$, $20000$},
                ytick={0.4, 0.6, 0.8, 1},
                yticklabels={$40$, $60$, $80$, $100$},
                xlabel={Number of tasks before checking},
                ylabel={Percentage (\%)},
                ymajorgrids=true,
                grid style=dashed,
                legend cell align=left,
                legend style={
                    at={(1.03, 1.00)},
                    anchor=north west,
                    font=\tiny,
                    legend columns=1},
                scaled x ticks=false,
                scaled y ticks=false,
                every axis plot/.append style={thick},
            ]

            \addplot[
                color=orange,
                mark=o
                ]
                plot coordinates {
                    (137, 0.846715328)
                    (1399, 0.839144216)
                    (2677, 0.800469484)
                    (3931, 0.818181818)
                    (5185, 0.799043062)
                    (6460, 0.768627451)
                    (7729, 0.776989756)
                    (9016, 0.747474747)
                    (10340, 0.748489426)
                    (11628, 0.735248447)
                };
            \addlegendentry{Ar}

            \addplot[
                color=lime,
                mark=o
                ]
                plot coordinates {
                    (143, 0.937062937)
                    (1465, 0.736006051)
                    (2838, 0.614712309)
                    (4249, 0.541459957)
                    (5544, 0.504247104)
                    (6913, 0.520818115)
                    (8227, 0.477929985)
                    (9621, 0.454806313)
                    (10974, 0.464892831)
                    (12305, 0.456048084)
                };
            \addlegendentry{De}

            \addplot[
                color=green,
                mark=o
                ]
                plot coordinates {
                    (149, 0.825503356)
                    (1465, 0.720364742)
                    (2772, 0.649579189)
                    (4161, 0.609071274)
                    (5418, 0.571201273)
                    (6739, 0.542013626)
                    (8026, 0.523698524)
                    (9397, 0.528081692)
                    (10769, 0.502915452)
                    (12081, 0.493140244)
                };
            \addlegendentry{Es}

            \addplot[
                color=teal,
                mark=o
                ]
                plot coordinates {
                    (146, 0.952054795)
                    (1556, 0.812056738)
                    (2971, 0.65795053)
                    (4384, 0.619957537)
                    (5759, 0.584)
                    (7178, 0.580690627)
                    (8589, 0.589652729)
                    (10006, 0.539167255)
                    (11387, 0.519913106)
                    (12814, 0.521373511)
                };
            \addlegendentry{Fr}

            \addplot[
                color=violet,
                mark=o
                ]
                plot coordinates {
                    (148, 0.885135135)
                    (1535, 0.794520548)
                    (2963, 0.703781513)
                    (4362, 0.669049321)
                    (5789, 0.632796076)
                    (7147, 0.62371134)
                    (8493, 0.609212481)
                    (9851, 0.597938144)
                    (11247, 0.591690544)
                    (12635, 0.564841499)
                };
            \addlegendentry{Id}

            \addplot[
                color=brown,
                mark=o
                ]
                plot coordinates {
                    (136, 0.823529412)
                    (1555, 0.670894996)
                    (2977, 0.639943741)
                    (4353, 0.530523256)
                    (5758, 0.525266904)
                    (7160, 0.494293866)
                    (8546, 0.515873016)
                    (9921, 0.530909091)
                    (11250, 0.461249059)
                    (12609, 0.53200883)
                };
            \addlegendentry{Ja}

            \addplot[
                color=purple,
                mark=o
                ]
                plot coordinates {
                    (140, 0.864285714)
                    (1549, 0.806955287)
                    (2970, 0.767769177)
                    (4400, 0.744055944)
                    (5788, 0.713976945)
                    (7200, 0.685552408)
                    (8598, 0.714592275)
                    (10013, 0.714487633)
                    (11416, 0.667854597)
                    (12821, 0.691814947)
                };
            \addlegendentry{Ko}

            \addplot[
                color=magenta,
                mark=o
                ]
                plot coordinates {
                    (149, 0.791946309)
                    (1543, 0.728120516)
                    (2922, 0.670775925)
                    (4265, 0.619508563)
                    (5692, 0.604765242)
                    (7068, 0.561046512)
                    (8449, 0.585807386)
                    (9781, 0.503753754)
                    (11169, 0.538904899)
                    (12550, 0.500362056)
                };
            \addlegendentry{Pt}
            
            \addplot[
                color=pink,
                mark=o
                ]
                plot coordinates {
                    (121, 0.950413223)
                    (1463, 0.888971684)
                    (2844, 0.842867487)
                    (4197, 0.77383592)
                    (5558, 0.759735489)
                    (6894, 0.737275449)
                    (8258, 0.711143695)
                    (9582, 0.704682779)
                    (10885, 0.738296239)
                    (12220, 0.686142322)
                };
            \addlegendentry{Ru}
            
            \addplot[
                color=cyan,
                mark=o
                ]
                plot coordinates {
                    (106, 0.933962264)
                    (1155, 0.864632984)
                    (2195, 0.809615385)
                    (3227, 0.779069767)
                    (4273, 0.792543021)
                    (5356, 0.782086796)
                    (6432, 0.800185874)
                    (7489, 0.769157994)
                    (8549, 0.733962264)
                    (9618, 0.762394761)
                };
            \addlegendentry{Th}

            \addplot[
                color=blue,
                mark=o
                ]
                plot coordinates {
                    (158, 0.784810127)
                    (1451, 0.750193349)
                    (2800, 0.717568569)
                    (4129, 0.650112867)
                    (5457, 0.652108434)
                    (6788, 0.648384673)
                    (8140, 0.653106509)
                    (9516, 0.623546512)
                    (10858, 0.568554396)
                    (12119, 0.552735924)
                };
            \addlegendentry{Vi}
        \end{axis}
    \end{tikzpicture}
    \caption{The percentage (\%) of the \mySFTDatasetNameS examples which pass the format and similarity checking. For each language, we compute the ratio of collected tasks to that before format and similarity checking during each round.}
    \label{figure.poly_alpaca.filter_ratio}
\end{figure}
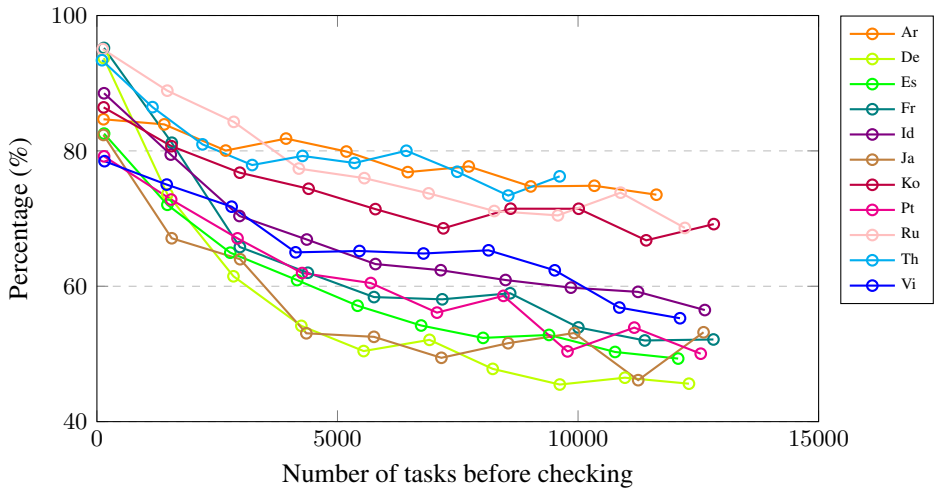

We show the percentage of remaining examples after similarity checking in Figure~\ref{figure.poly_alpaca.filter_ratio}. For each language, we show the number of self-instruct tasks in \mySFTDatasetNameS in Table~\ref{table.poly_alpaca.number_of_tasks}.

\begin{table}
    \centering
    \begin{tabular}{lcc}
        \toprule
            \textbf{Language} & \textbf{\# tasks} & \textbf{Ratio (\%)} \\
        \midrule
            Ar & ~~14,671 & ~~11.06 \\
            De & ~~~~9,515 & ~~~~7.17 \\
            Es & ~~~~9,958 & ~~~~7.50 \\
            Fr & ~~11,332 & ~~~~8.54 \\
            Id & ~~12,117 & ~~~~9.13 \\
            Ja & ~~10,191 & ~~~~7.58 \\
            Ko & ~~14,402 & ~~10.85 \\
            Pt & ~~10,825 & ~~~~8.16 \\
            Ru & ~~14,286 & ~~10.77 \\
            Th & ~~11,496 & ~~~~8.66 \\
            Vi & ~~13,908 & ~~10.48 \\
        \midrule
            Total & 132,701 & 100.00 \\
        \bottomrule
    \end{tabular}
    \caption{The number of tasks for each language in \mySFTDatasetNameS dataset.}
    \label{table.poly_alpaca.number_of_tasks}
\end{table}

\section{Details of Task Formatting}
\label{appendix_task_formatting}

We list the detailed format of all the tasks in our multilingual benchmark in the following tables.

\begin{table}[h!]
\centering
\label{tab:task_desc_xnli}
\begin{tabularx}{\linewidth}{r X}
\toprule
Context →          & Using these eight simple techniques, you can fabricate a news story in the comfort of your own home., right?                                                                                                                             \\ \midrule
Correct Answer →   & No, Only news reporters in a newsroom can write a news story, and it takes 20 steps to do it.                                                                                                                                            \\
Incorrect Answer → & Yes, Only news reporters in a newsroom can write a news story, and it takes 20 steps to do it.\newline Also, Only news reporters in a newsroom can write a news story, and it takes 20 steps to do it. \\ \bottomrule
\end{tabularx}%

\caption{The task format of XNLI. We normalize the log-likelihood by token numbers of all the correct and incorrect answers and choose the one with the largest score as prediction.}
\end{table}

\begin{table}[h!]
\centering

\label{tab:task_desc_xcopa}
\begin{tabularx}{\linewidth}{r X}
\toprule
Context →          & Il cursore sullo schermo del computer si è mosso perché \\ \midrule
Correct Answer →   & l'utente ha spostato il mouse.                          \\ 
Incorrect Answer → & l'utente ha cliccato il mouse.                          \\ \bottomrule
\end{tabularx}
\caption{The task format of XCOPA.}
\end{table}

\begin{table}[h!]
\centering

\label{tab:task_desc_xwinograd}
\begin{tabularx}{\linewidth}{r X}
\toprule
Context →          & He put snow on the smiley face because \\ \midrule
Correct Answer →   & snow was wet.                          \\ 
Incorrect Answer → & the smiley face was wet.                          \\ \bottomrule
\end{tabularx}
\caption{The task format of XWinograd.}
\end{table}

\begin{table}[h!]
\centering

\label{tab:task_desc_pawsx}
\begin{tabularx}{\linewidth}{r X}
\toprule
Context →          & The first category is monovalent verbs , where there is only one semantic argument and it consists of both unergative verbs and unaccusative verbs ., right?  \\ \midrule
Correct Answer →   & Yes, The first category is unergative verbs , where there is only one unaccountable argument and consists of both semantic verbs and monovalent verbs .                          \\ 
Incorrect Answer → & No, The first category is unergative verbs , where there is only one unaccountable argument and consists of both semantic verbs and monovalent verbs .                          \\ \bottomrule
\end{tabularx}
\caption{The task format of PAWS-X.}
\end{table}

\begin{table}[h!]
\centering

\label{tab:task_desc_tydiqa}
\begin{tabularx}{\linewidth}{r X}
\toprule
Context →          & Read the context and answer the question in one or a few words in English. \newline \newline Context (English): Football games last for a total of 60 minutes in professional and college play and are divided into two-halves of 30 minutes and four-quarters of 15 minutes.[74][75] High school football games are 48 minutes in length with two-halves of 24 minutes and four-quarters of 12 minutes.[76] (...). \newline Answer:  \\ \midrule
Target Completion →   & class B                         \\ 
\bottomrule
\end{tabularx}
\caption{The task format of Tydi-QA.}
\end{table}

\begin{table}[h!]
\centering

\label{tab:task_desc_mtg_tg}
\begin{tabularx}{\linewidth}{r X}
\toprule
Context →          & Please generate a title for the following document in English \newline document: justin timberlake's super bowl lii halftime show is approaching, and there are rumors circulating around the internet saying that nsync may have a reunion at the annual championship. well, it appears that fans of the boyband have to kiss that dream goodbye after joey fatone shuts down the rumors. tmz recently caught up with the 40-year-old hot dog purveyor and asked whether the reunion rumors were true. (...). \newline title: \\ \midrule
Target Completion →   & joey fatone shuts down nsync reunion rumors at the super bowl lii                         \\ 
\bottomrule
\end{tabularx}
\caption{The task format of Title Generation task in MTG.}
\end{table}

\begin{table}[h!]
\centering

\label{tab:task_desc_mtg_sg}
\begin{tabularx}{\linewidth}{r X}
\toprule
Context →          & Write a story end of the following story in just a few sentences in English. \newline story: john had a roommate he wanted to prank. john called and ordered ten pizzas to be delivered to the apartment. suddenly, john's roommate got a call and had to leave. \newline story ending: \\ \midrule
Target Completion →   & it was too late to cancel the pizza order! john wound up paying for all the pizzas!                   \\ 
\bottomrule
\end{tabularx}
\caption{The task format of Story Ending Generation task in MTG.}
\end{table}

\begin{table}[h!]
\centering

\label{tab:task_desc_mtg_qg}
\begin{tabularx}{\linewidth}{r X}
\toprule
Context →          & Given a passage and a concept that can be found in this passage, please generate a question in English, the answer of which is this concept and is answerable after reading this passage. \newline passage: a treaty is an official, express written agreement that states use to legally bind themselves. a treaty is the official document which expresses that agreement in words; and it is also the objective outcome of a ceremonial occasion which acknowledges the parties and their defined relationships. \newline answer: themselves \newline question: \\ \midrule
Target Completion →   & who is responsible for the legally-bound obligations of the parties to a treaty?                       \\ 
\bottomrule
\end{tabularx}
\caption{The task format of Question Generation task in MTG.}
\end{table}

\begin{table}[t]
\centering
\label{tab:task_desc_mtg_summary}
\begin{tabularx}{\linewidth}{r X}
\toprule
Context →          & Please generate a short summary of the given document in English \newline document: a man whose girlfriend ran off with his stepfather and gave birth to a baby nearly four years ago has spoken of his delight after dna tests proved the baby is actually his. love rat stan crowther, 47, and rachel delaney, 18 at the time, set up home together and rachel then gave birth to a baby daughter, whom stan believed was his. but recent bombshell dna tests showed that the baby, living in chorley, lancashire, actually belongs to her former partner, and stan's former stepson, ashley mercer, 27. (...). \newline summary: \\ \midrule
Target Completion →   & ashley mercer was 22 when he discovered his girlfriend rachel delaney, 18, was having an affair with his stepfather stan crowther, 43. stan had been married to ashley's mother mandy rourke for ten years. stan and rachel moved in together and shortly after rachel, from chorley, lancashire, gave birth to a baby girl. stan assumed the baby to be his but dna tests have since revealed ashley is the four-year-old girl's real father. ashley's mother mandy rourke has now forgiven her ex-husband. stan and rachel are no longer together. \\ 
\bottomrule
\end{tabularx}
\caption{The task format of Summarization task in MTG.}
\end{table}

\begin{table}[h!]
\centering

\label{tab:task_translation}
\begin{tabularx}{\linewidth}{r X}
\toprule
Context →          & Oil falls after Iran claims US offered to remove sanctions, Trump denies \newline
Translate this sentence from English to German. \newline \newline \\ \midrule
Target Completion →   & Öl fällt, nachdem Iran behauptet, die USA hätten Aufhebung der Sanktionen angeboten, Trump dementiert \\ 
\bottomrule
\end{tabularx}
\caption{The task format of Translation.}
\end{table}

\section{Details of Experimental Results}
\label{Appendix.results}

\subsection{Results of Pretrained Language Models}

\begin{table}[h]
\centering
\resizebox{\textwidth}{!}{
\begin{tabular}{@{}llllllllllll|llllll@{}}
\toprule
          & en   & zh   & ar   & es   & fr   & ru   & th   & tr   & vi   & de   & Average & bg   & el   & hi   & sw   & ur   & Average \\ \midrule
BLOOM-7.1B &
  53.9 &
  35.5 &
  33.8 &
  48.7 &
  49.8 &
  42.6 &
  34.9 &
  34.9 &
  \textbf{47.4} &
  39.6 &
  42.1 &
  39.3 &
  35.5 &
  46.7 &
  37.9 &
  41.9 &
  40.3 \\
LLaMA-13B & 35.5 & 34.6 & 34.1 & 33.4 & 33.6 & 33.6 & 34.6 & 34.0 & 34.1 & 35.2 & 34.3    & 33.9 & 34.5 & 35.7 & 33.2 & 34.1 & 34.3    \\
\textsc\{Poly\}LM-13B &
  \textbf{54.6} &
  \textbf{35.9} &
  33.6 &
  \textbf{50.0} &
  \textbf{52.1} &
  \textbf{49.0} &
  \textbf{44.6} &
  \textbf{44.5} &
  46.7 &
  \textbf{50.2} &
  \textbf{46.1} &
  36.3 &
  33.8 &
  34.9 &
  34.4 &
  33.5 &
  34.6 \\ \bottomrule
\end{tabular}}
\label{tab:main_results_xnli}
\caption{Accuracy on XNLI. }
\end{table}

\begin{table}[h]
\centering
\begin{tabular}{@{}lllllllll@{}}
\toprule
           & en            & zh   & es   & fr   & ja   & ko            & de   & Average \\ \midrule
BLOOM-7.1B & \textbf{61.3} & 47.3 & 59.4 & 50.9 & 45.5 & 45.1          & 52.9 & 51.8    \\
LLaMA-13B  & 53.7          & 45.2 & 52.1 & 54.5 & 45.0 & \textbf{47.1} & 53.0 & 50.1    \\
\textsc{Poly}LM-13B & 60.9 & \textbf{56.0} & \textbf{59.7} & \textbf{56.6} & \textbf{51.0} & 46.8 & \textbf{60.3} & \textbf{55.9} \\ \bottomrule
\end{tabular}
\caption{Accuracy on PAWS-X.}
\label{tab:main_results_pawsx}
\end{table}

\begin{table}[h]
\centering
\begin{tabular}{@{}llllllll@{}}
\toprule
           & en   & zh   & ja   & pt            & ru            & fr   & Average \\ \midrule
BLOOM-7.1B & 82.2 & 74.4 & 58.5 & \textbf{76.8} & 56.8          & 71.1 & 70.0    \\
LLaMA-13B  & 86.8 & 70.0 & 59.9 & 71.5          & \textbf{70.8} & 68.7 & 71.3    \\
\textsc{Poly}LM-13B & \textbf{84.6} & \textbf{76.6} & \textbf{65.7} & 74.9 & 65.1 & \textbf{73.5} & \textbf{73.4} \\ \bottomrule
\end{tabular}
\label{tab:main_results_xwinograd}
\caption{Accuracy on XWinograd. }
\end{table}

\begin{table}[h]
\centering
\resizebox{\textwidth}{!}{
\begin{tabular}{@{}llllllll|llllll@{}}
\toprule
           & id   & it            & th   & tr   & vi            & zh   & Avg. & et   & ht            & qu            & sw   & ta            & Avg.       \\ \midrule
BLOOM-7.1B & 69.8 & 52.8          & 55.4 & 51.2 & \textbf{70.8} & 65.2 & 60.9    & 48.2 & 50.8          & 50.8 & 51.6 & 59.2 & 52.1 \\
LLaMA-13B  & 57.8 & \textbf{67.2} & 54.6 & 53.0 & 53.8          & 58.4 & 57.5    & 48.2 & 52.8 & 50.2          & 51.2 & 54.4          & 51.4          \\
\textsc{Poly}LM-13B &
  \textbf{70.2} &
  66.0 &
  \textbf{58.6} &
  \textbf{57.8} &
  \textbf{70.8} &
  \textbf{67.0} &
  \textbf{65.1} &
  49.8 &
  50.4 &
  50.4 &
  51.8 &
  55.0 &
  51.5 \\ \bottomrule
\end{tabular}}
\caption{Accuracy on XCOPA. The left part presents the results of languages we mainly considered in the training phrase, while the right part shows the other languages in the testsets. `Avg.' means the average accuracy.}
\label{tab:main_results_xcopa}
\end{table}

\begin{table}[h]
\centering
\resizebox{\textwidth}{!}{
\begin{tabular}{@{}lllllll|lllll@{}}
\toprule
           & ar            & en   & id            & ko   & ru            & Average & fi   & bn   & sw   & te   & Average \\ \midrule
BLOOM-7.1B & 42.6          & 51.6 & \textbf{48.7} & 8.6  & 33.8          & 37.1    & 17.5 & 55.1 & 56.8 & 40.9 & 42.6    \\
LLaMA-13B  & \textbf{49.7} & 54.4 & 43.4          & 49.7 & \textbf{41.8} & 47.8    & 40.2 & 32.0 & 33.5 & 8.5  & 28.6    \\
\textsc{Poly}LM-13B & 44.9 & \textbf{58.0} & 48.6 & \textbf{53.8} & 39.5 & \textbf{49.0} & 20.9 & 2.9 & 22.0 & 3.7 & 12.4 \\ \bottomrule
\end{tabular}}
\caption{F1 scores on the TyDiQA-GoldP benchmark under one-shot conditions. The left part presents the results of languages we mainly considered in the training phrase, while the right part shows the other languages in the testsets.}
\label{table:main_results_tydiqa}
\end{table}

\begin{table}[h!]
\centering
\begin{tabular}{@{}clcccccc@{}}
\toprule
Task    &            & zh            & en            & es            & fr            & de            & Average       \\ \midrule
   & BLOOM-7.1B & \textbf{16.6} & \textbf{13.2} & \textbf{11.0} & \textbf{13.1} & \textbf{6.6}  & \textbf{12.1} \\
SG      & LLaMA-13B  & 16.5          & 12.5          & 2.0           & 8.1           & 0.7           & 8.0           \\
        & \textsc{Poly}LM-13B & 10.3          & 10.3          & 7.8           & 9.2           & 6.3           & 8.8           \\ \midrule
        & BLOOM-7.1B & 12.4          & 14.2          & 10.1          & 10.9          & 5.7           & 10.7          \\
TG & LLaMA-13B  & 11.5          & \textbf{19.6} & \textbf{15.9} & \textbf{16.8} & \textbf{10.6} & \textbf{14.9} \\
        & \textsc{Poly}LM-13B & \textbf{16.8} & 16.5          & 14.2          & 14.3          & 10.1          & 14.4          \\ \midrule
        & BLOOM-7.1B & 13.8          & 15.5          & 15.0          & 13.5          & 7.3           & 13.0          \\
QG      & LLaMA-13B  & \textbf{20.2} & \textbf{16.3} & 14.6          & 13.1          & 4.5           & 13.7          \\
        & \textsc{Poly}LM-13B & 20.0          & 15.9          & \textbf{17.1} & \textbf{13.9} & \textbf{7.5}  & \textbf{14.9} \\ \midrule
        & BLOOM-7.1B & 11.0          & 10.3          & 10.0          & 9.7           & 7.5           & 9.7           \\
Sum     & LLaMA-13B  & 14.6          & \textbf{18.1} & \textbf{8.1}  & 10.3          & 7.6           & 11.7          \\
        & \textsc{Poly}LM-13B & \textbf{16.6} & 17.0          & 16.9          & \textbf{15.1} & \textbf{11.6} & \textbf{15.4} \\ \midrule
        & BLOOM-7.1B & 13.5          & 13.5          & 11.5          & 11.8          & 6.8           & 11.4          \\
Average & LLaMA-13B  & 15.7          & \textbf{16.6} & 10.1          & 12.1          & 5.9           & 12.1          \\
   & \textsc{Poly}LM-13B & \textbf{17.5} & 14.9          & \textbf{14.0} & \textbf{13.1} & \textbf{8.9}  & \textbf{13.7} \\ \bottomrule
\end{tabular}
\caption{Rouge scores on the MTG benchmark under one-shot conditions. Results are presented in two dimensions: language directions and task types. The bottom row shows the mean values of diverse language directions across all tasks. Similarly, the rightmost column depicts the average values of varied tasks across all language directions.}
\label{table:main_results_mtg}
\end{table}

\begin{table}[h]
\resizebox{\textwidth}{!}{
\begin{tabular}{@{}lccccc|ccccc@{}}
\toprule
           & de$\to$en & ja$\to$en & ru$\to$en & zh$\to$en & Avg. & en$\to$de & en$\to$ja & en$\to$ru & en$\to$zh & Avg. \\ \midrule
BLOOM-7.1B & 23.9  & 6.5   & 17.7  & 16.3  & 16.1    & 3.7   & 2.1   & 1.4   & 8.3   & 3.9     \\
LLaMA-13B  & \textbf{36.9} & \textbf{13.6} & \textbf{32.7} & \textbf{22.6} & \textbf{26.5} & 23.6          & 9.6           & 16.8          & 7.4           & 14.3          \\
\textsc{Poly}LM-13B & 33.9          & 10.2          & 32.4          & 22.0          & 24.6          & \textbf{24.8} & \textbf{17.0} & \textbf{17.8} & \textbf{19.0} & \textbf{19.7} \\ \bottomrule
\end{tabular}}
\caption{Translation BLEU scores on the WMT20 machine translation task under one-shot conditions. `Avg.' means the average BLEU scores of translations to English or from English.}
\label{table:main_results_mt}
\end{table}

\subsection{Results of SFT Models}

\begin{table}[h]
\centering
\resizebox{\textwidth}{!}{
\begin{tabular}{@{}llllllllllll|llllll@{}}
\toprule
                 & en   & zh   & ar            & es   & fr   & ru   & th   & tr   & vi   & de   & Average & bg   & el   & hi   & sw   & ur   & Average \\ \midrule
BLOOMZ-MT-7.1B   & 44.9 & 33.2 & \textbf{36.0} & 36.1 & 47.5 & 38.1 & 33.5 & 33.4 & 37.9 & 42.5 & 38.3    & 35.5 & 35.1 & 39.6 & 33.2 & 37.9 & 36.3    \\
LLaMA-Alpaca-13B & 35.7 & 34.6 & 33.2          & 33.3 & 33.4 & 33.4 & 36.2 & 34.7 & 34.6 & 33.7 & 34.3    & 34.3 & 34.1 & 35.8 & 33.0 & 32.8 & 34.0    \\
\textsc{Poly}LM-\mySFTDatasetName-13B &
  \textbf{54.3} &
  \textbf{36.0} &
  33.3 &
  \textbf{47.0} &
  \textbf{49.6} &
  \textbf{46.7} &
  \textbf{38.4} &
  \textbf{44.1} &
  \textbf{44.6} &
  \textbf{48.0} &
  \textbf{44.2} &
  36.1 &
  32.7 &
  34.1 &
  33.1 &
  33.8 &
  34.0 \\ \bottomrule
\end{tabular}}
\caption{Accuracy of instruction-followed models on XNLI.}
\end{table}

\begin{table}[h]
\centering
\begin{tabular}{@{}lllllllll@{}}
\toprule
                 & en   & zh            & es   & fr            & ja            & ko   & de   & Average \\ \midrule
BLOOMZ-MT-7.1B   & 61.2 & \textbf{58.5} & 58.8 & \textbf{61.1} & \textbf{54.4} & 49.5 & 56.7 & 57.2    \\
LLaMA-Alpaca-13B & 53.3 & 45.3          & 53.5 & 54.1          & 46.0          & 48.3 & 54.3 & 50.7    \\
\textsc{Poly}LM-\mySFTDatasetName-13B & \textbf{65.3} & 55.9 & \textbf{63.0} & 59.8 & 52.3 & \textbf{53.7} & \textbf{64.9} & \textbf{59.3} \\ \bottomrule
\end{tabular}
\caption{Accuracy of instruction-followed models on PAWS-X.}
\end{table}

\begin{table}[h]
\centering
\begin{tabular}{@{}llllllll@{}}
\toprule
               & en   & zh   & ja   & pt   & ru   & fr   & Average \\ \midrule
BLOOMZ-MT-7.1B & 83.5 & 71.0 & 56.4 & 65.4 & 53.7 & 68.7 & 66.5    \\
LLaMA-Alpaca-13B                      & \textbf{88.6} & 67.3          & 61.4          & \textbf{73.0} & \textbf{72.1} & \textbf{78.3} & \textbf{73.5} \\
\textsc{Poly}LM-\mySFTDatasetName-13B & 83.9          & \textbf{73.6} & \textbf{65.2} & 72.2          & 67.9          & 71.1          & 72.3          \\ \bottomrule
\end{tabular}
\caption{Accuracy of instruction-followed models on XWinograd.}
\end{table}

\begin{table}[h]
\centering
\resizebox{\textwidth}{!}{
\begin{tabular}{@{}llllllll|llllll@{}}
\toprule
                 & id   & it            & th   & tr   & vi   & zh   & Avg. & et   & ht   & qu            & sw   & ta   & Avg. \\ \midrule
BLOOMZ-MT-7.1B &
  58.6 &
  51.8 &
  53.6 &
  53.2 &
  58.8 &
  62.2 &
  56.4 &
  49.6 &
  53.8 &
  49.4 &
  53.0 &
  58.2 &
  52.8 \\
LLaMA-Alpaca-13B & 55.4 & \textbf{70.8} & 54.6 & 53.0 & 53.0 & 60.0 & 57.8    & 47.2 & 53.0 & 51.8 & 51.0 & 56.0 & 51.8    \\
\textsc{Poly}LM-\mySFTDatasetName-13B &
  \textbf{71.6} &
  66.8 &
  \textbf{60.2} &
  \textbf{58.8} &
  \textbf{71.8} &
  \textbf{75.6} &
  \textbf{67.5} &
  48.4 &
  52.0 &
  50.4 &
  50.8 &
  55.4 &
  51.4 \\ \bottomrule
\end{tabular}}
\caption{Accuracy of instruction-followed models on XCOPA.}
\end{table}

\begin{table}[h]
\centering
\resizebox{\textwidth}{!}{
\begin{tabular}{@{}lllllll|lllll@{}}
\toprule
                 & en            & ar   & id   & ko   & ru   & Avg. & bn   & fi   & sw   & te   & Avg. \\ \midrule
BLOOMZ-MT-7.1B   & 22.4          & 36.6 & 26.9 & 5.8  & 9.1  & 20.2    & 26.7 & 2.4  & 14.4 & 26.5 & 17.5    \\
LLaMA-Alpaca-13B & \textbf{59.2} & 20.8 & 48.6 & 19.3 & 37.7 & 37.1    & 11.0 & 50.6 & 20.7 & 5.7  & 22.0    \\
\textsc{Poly}LM-\mySFTDatasetName-13B & 58.7 & \textbf{50.7} & \textbf{52.1} & \textbf{30.1} & \textbf{40.3} & \textbf{46.4} & 2.5 & 8.5 & 4.6 & 1.9 & 4.4 \\ \bottomrule
\end{tabular}
}
\caption{F1 scores of instruction-followed models on the TyDiQA-GoldP benchmark.}
\end{table}

\begin{table}[h]
\centering
\begin{tabular}{@{}llllllll@{}}
\toprule
Task    &                                       & zh   & en   & es   & fr   & de   & Average       \\ \midrule
        & BLOOMZ-MT-7.1B                        & 12.1 & 12.6 & 9.8  & 12.1 & 5.6  & 10.4          \\
SG      & LLaMA-Alpaca-13B                      & 3.8  & 10.0 & 4.8  & 5.7  & 5.6  & 6.0           \\
 & \textsc{Poly}LM-\mySFTDatasetName-13B & 13.9          & 9.5           & 9.6           & 11.5          & 8.0           & \textbf{10.5} \\ \midrule
        & BLOOMZ-MT-7.1B                        & 18.5 & 25.7 & 18.9 & 16.8 & 9.6  & \textbf{17.9} \\
TG      & LLaMA-Alpaca-13B                      & 9.3  & 25.1 & 15.5 & 15.5 & 10.1 & 15.1          \\
        & \textsc{Poly}LM-\mySFTDatasetName-13B & 15.3 & 23.7 & 16.6 & 15.9 & 11.7 & 16.6          \\ \midrule
        & BLOOMZ-MT-7.1B                        & 23.8 & 31.9 & 29.6 & 26.4 & 9.2  & \textbf{24.2} \\
QG      & LLaMA-Alpaca-13B                      & 4.7  & 27.5 & 27.3 & 19.2 & 18.1 & 19.4          \\
        & \textsc{Poly}LM-\mySFTDatasetName-13B & 21.8 & 25.7 & 27.2 & 19.3 & 16.3 & 22.1          \\ \midrule
        & BLOOMZ-MT-7.1B                        & 16.5 & 18.4 & 19.7 & 18.4 & 11.0 & \textbf{16.8} \\
Sum     & LLaMA-Alpaca-13B                      & 2.6  & 22.0 & 16.9 & 15.2 & 13.4 & 14.0          \\
        & \textsc{Poly}LM-\mySFTDatasetName-13B & 15.4 & 21.5 & 18.1 & 15.2 & 13.0 & 16.6          \\ \midrule
 & BLOOMZ-MT-7.1B                        & \textbf{17.7} & \textbf{22.2} & \textbf{19.5} & \textbf{18.4} & 8.9           & \textbf{17.3} \\
Average & LLaMA-Alpaca-13B                      & 5.1  & 21.2 & 16.1 & 13.9 & 11.8 & 13.6          \\
 & \textsc{Poly}LM-\mySFTDatasetName-13B & 16.6          & 20.1          & 17.9          & 15.5          & \textbf{12.3} & 16.5          \\ \bottomrule
\end{tabular}
\caption{Rouge scores of instruction-followed models on the MTG benchmark.}
\end{table}

\begin{table}[h]
\centering
\resizebox{\textwidth}{!}{
\begin{tabular}{@{}llllll|lllll@{}}
\toprule
           & de$\to$en & ja$\to$en & ru$\to$en & zh$\to$en & Avg. & en$\to$de & en$\to$ja & en$\to$ru & en$\to$zh & Avg. \\ \midrule
BLOOMZ-MT-7.1B   & 17.99 & 5.09  & 7.67  & 12.89 & 10.91   & 2.94  & 0.99  & 0.77  & 8.40  & 3.3     \\
LLaMA-Alpaca-13B & 30.0  & 9.9   & 20.3  & 16.4  & 19.14   & 16.8  & 3.6   & 10.6  & 3.0   & 8.5     \\
\textsc{Poly}LM-\mySFTDatasetName-13B &
  \textbf{34.2} &
  \textbf{14.3} &
  \textbf{31.8} &
  \textbf{25.0} &
  \textbf{26.32} &
  \textbf{25.9} &
  \textbf{16.2} &
  \textbf{17.9} &
  \textbf{21.9} &
  \textbf{20.5} \\ \bottomrule
\end{tabular}}
\caption{Translation BLEU scores of instruction-followed models on the WMT20 machine translation task.}
\end{table}

\section{Demonstrations for Qualitative Analysis}
\label{appendix:demo}

\begin{figure}[h]
    \centering
    \includegraphics[width=\textwidth]{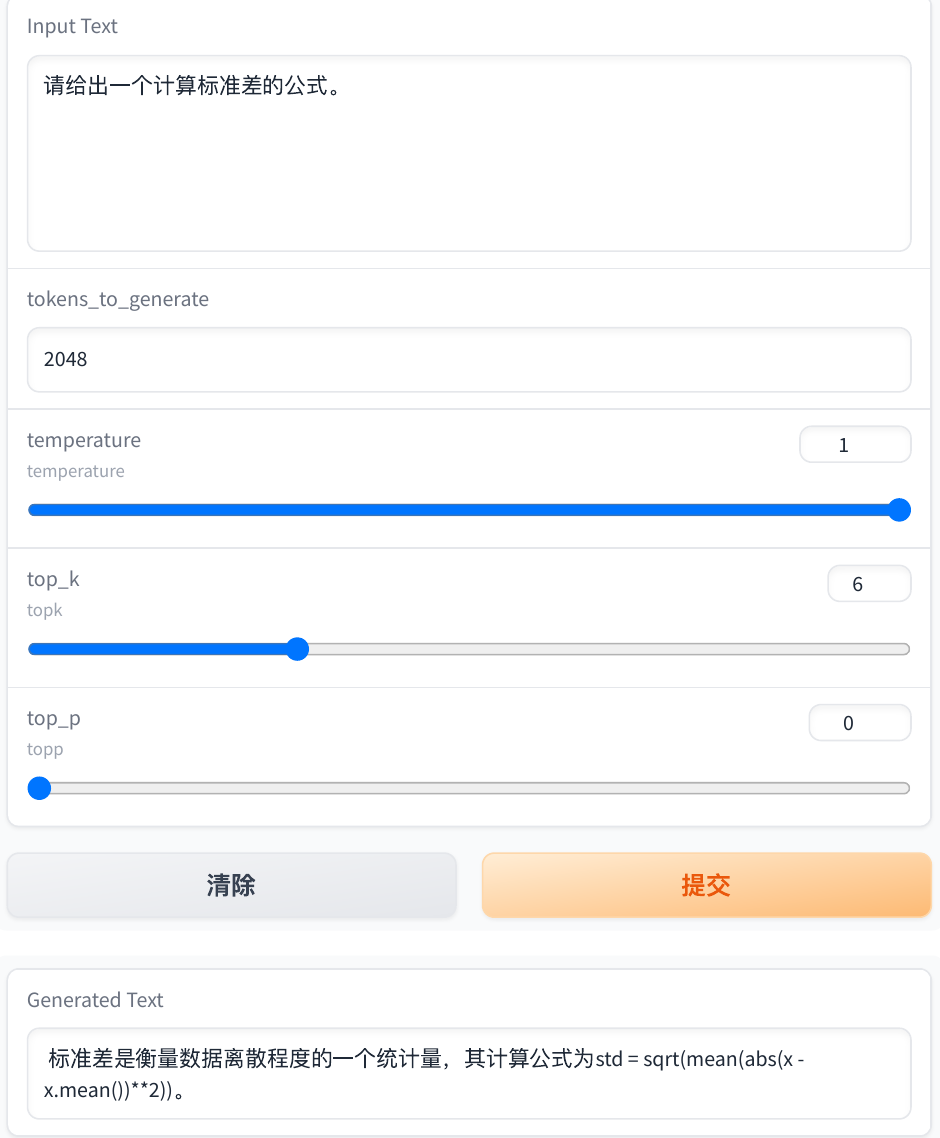}
\end{figure}

\begin{figure}[h]
    \centering
    \includegraphics[width=\textwidth]{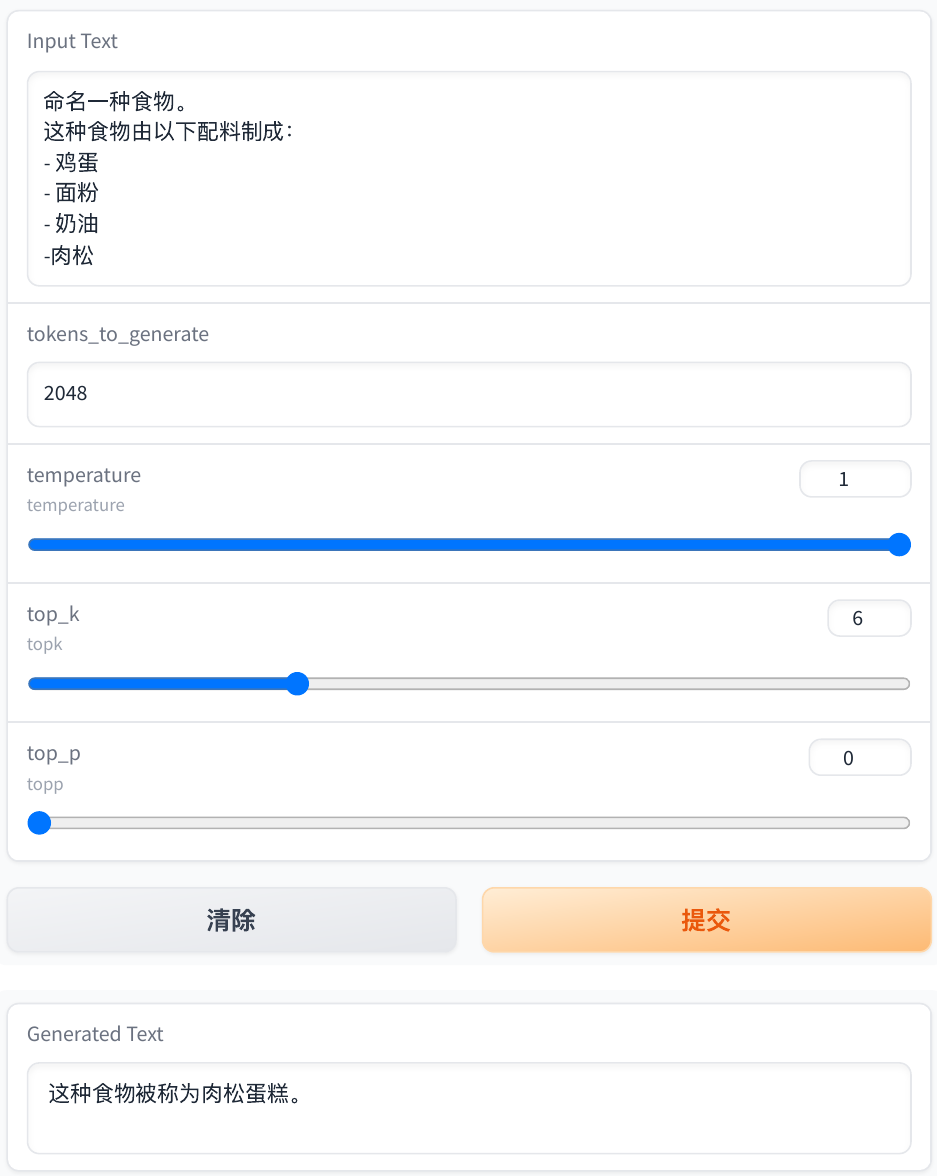}
\end{figure}

\begin{figure}[h]
    \centering
    \includegraphics[width=\textwidth]{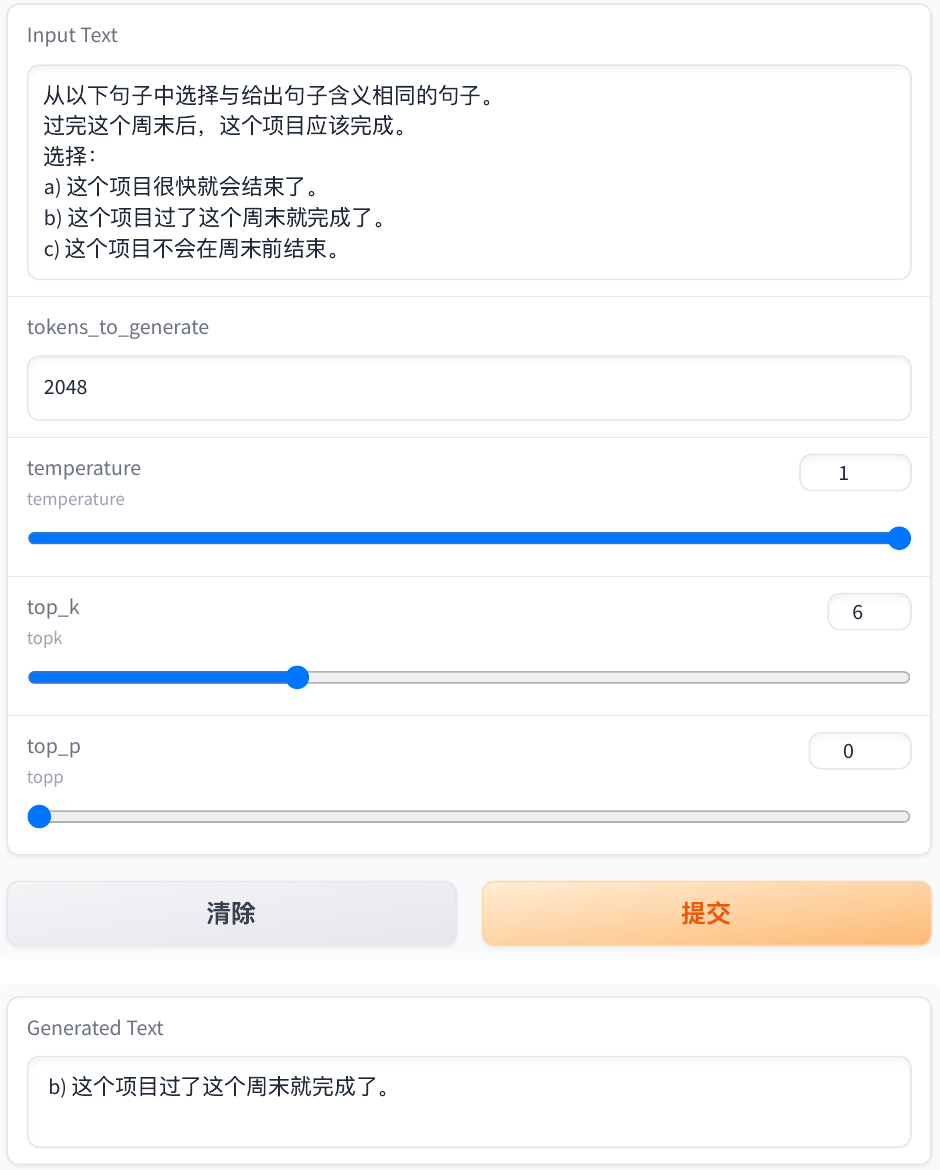}
\end{figure}

\begin{figure}[h]
    \centering
    \includegraphics[width=\textwidth]{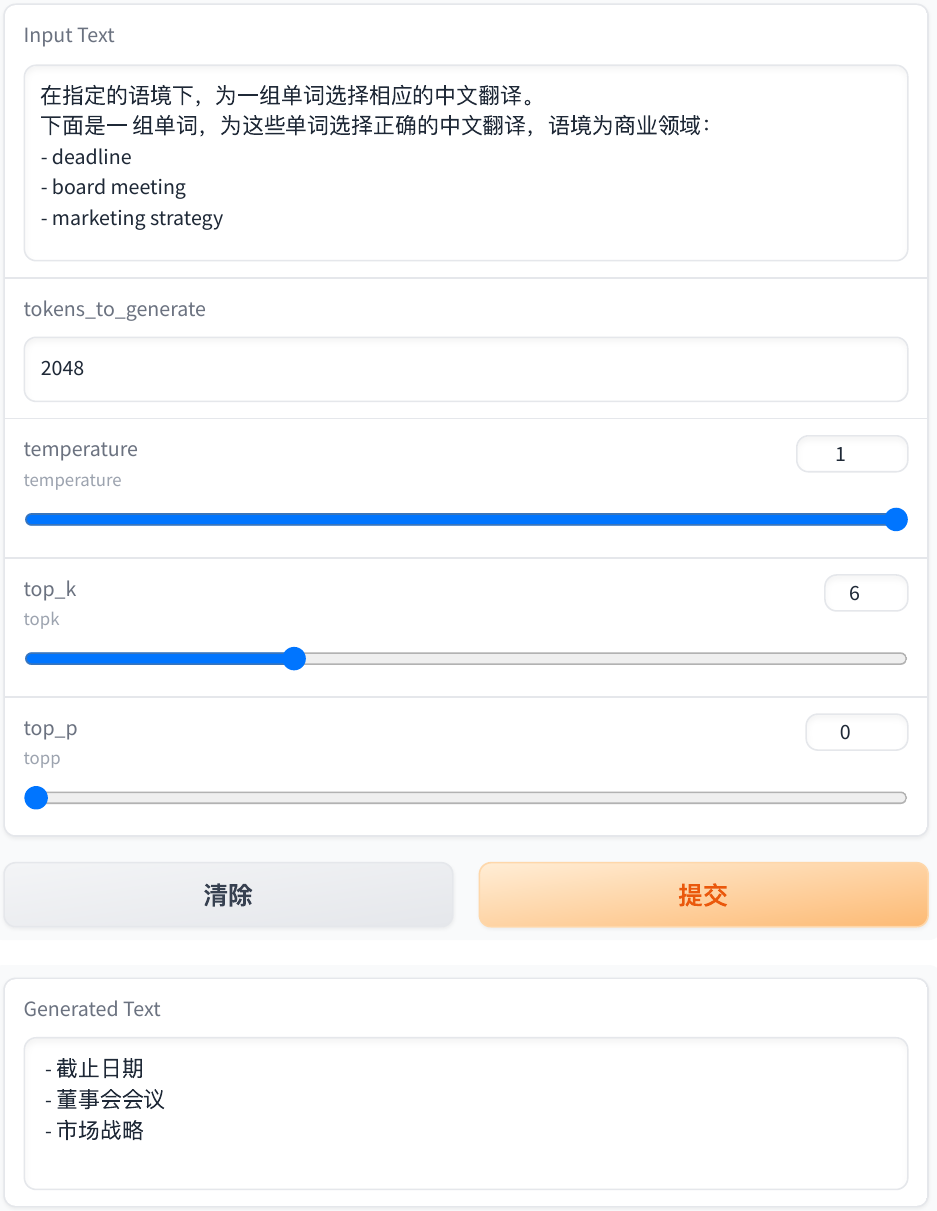}
\end{figure}

\begin{figure}[h]
    \centering
    \includegraphics[width=\textwidth]{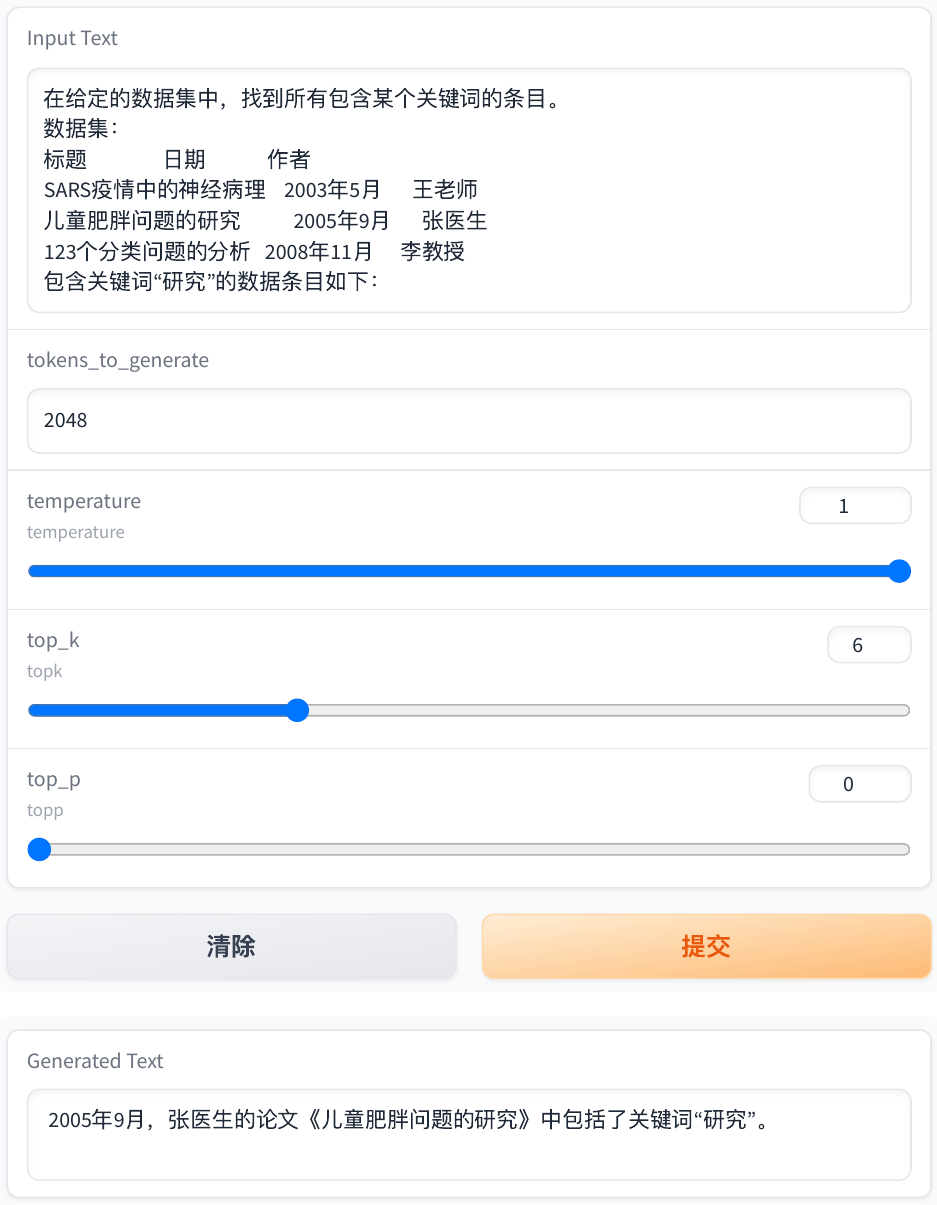}
\end{figure}

\begin{figure}[h]
    \centering
    \includegraphics[width=\textwidth]{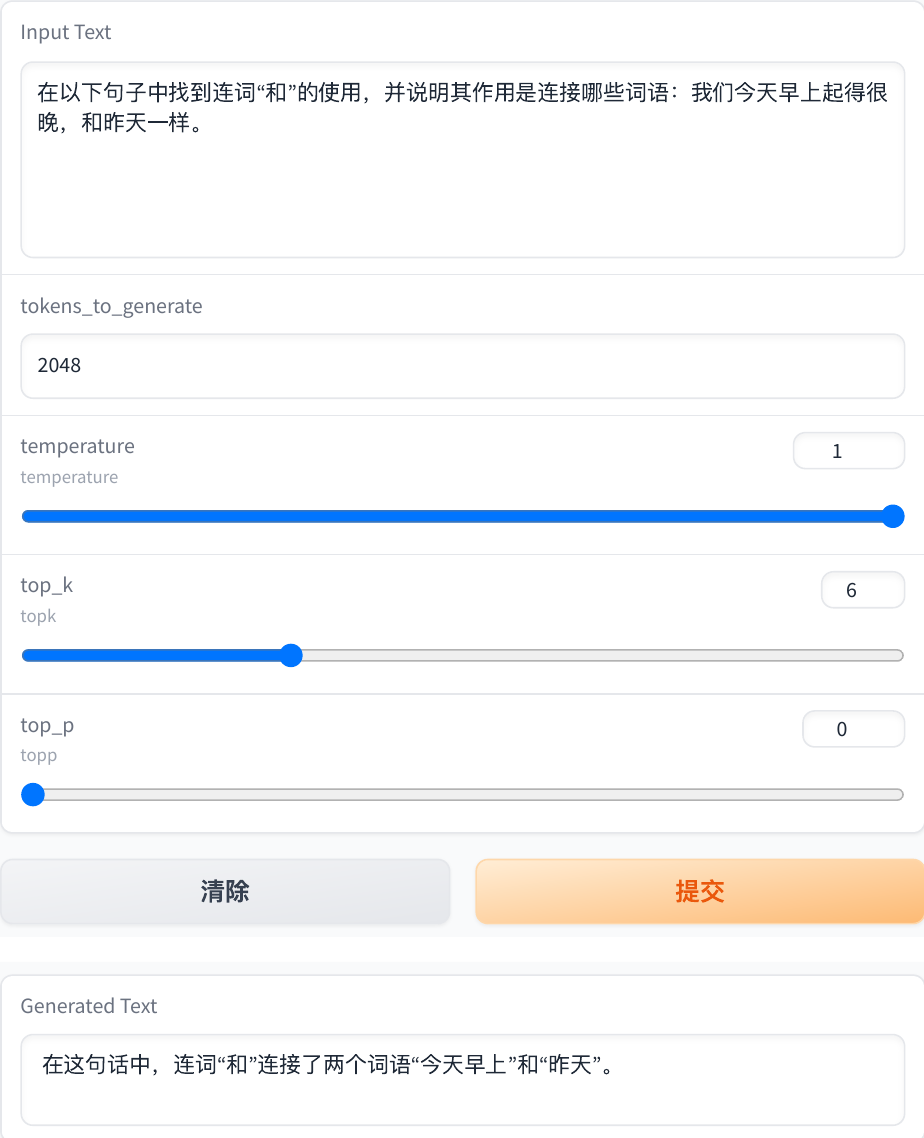}
\end{figure}

\begin{figure}[h]
    \centering
    \includegraphics[width=\textwidth]{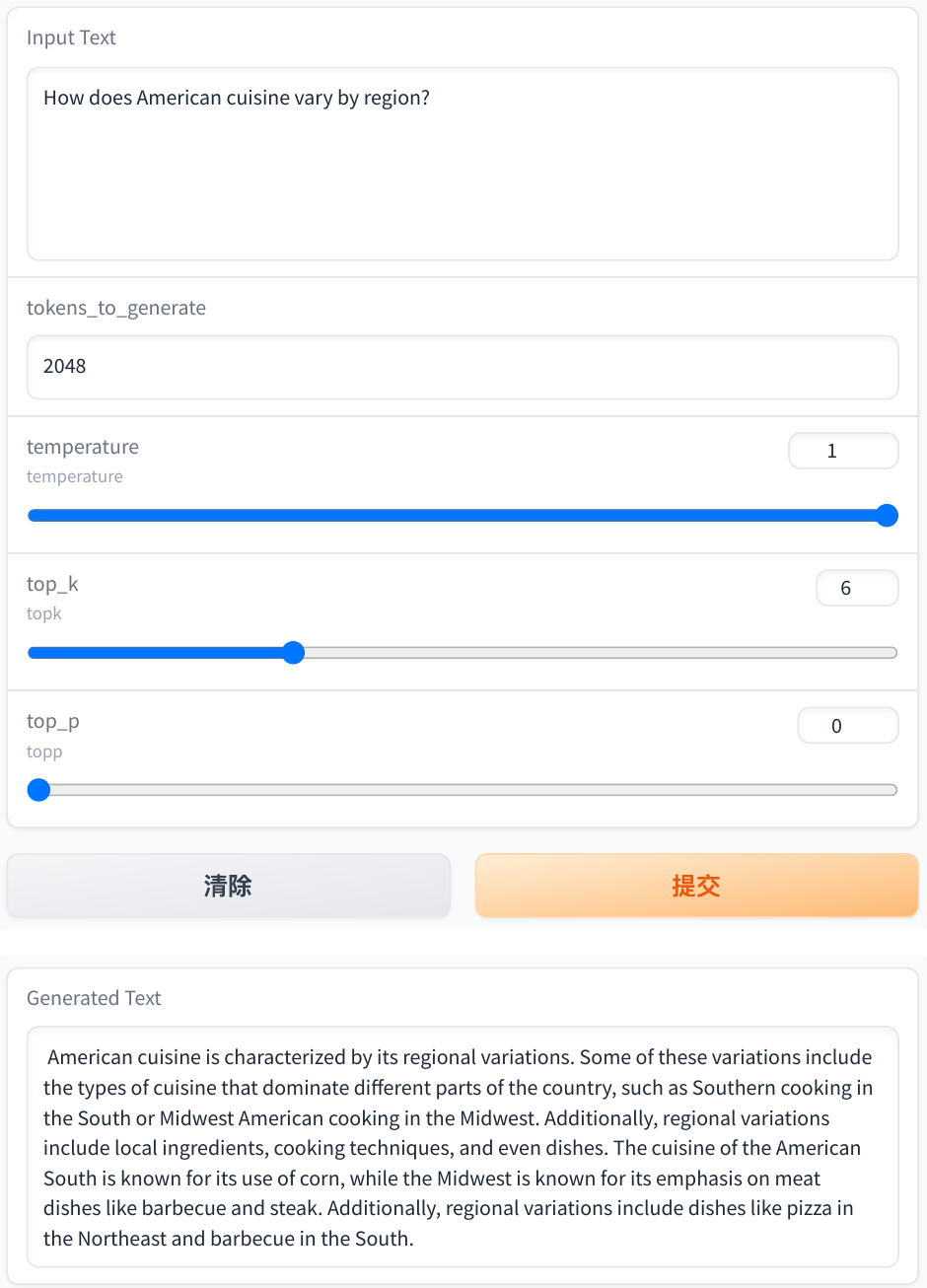}
\end{figure}
\begin{figure}[h]
    \centering
    \includegraphics[width=\textwidth]{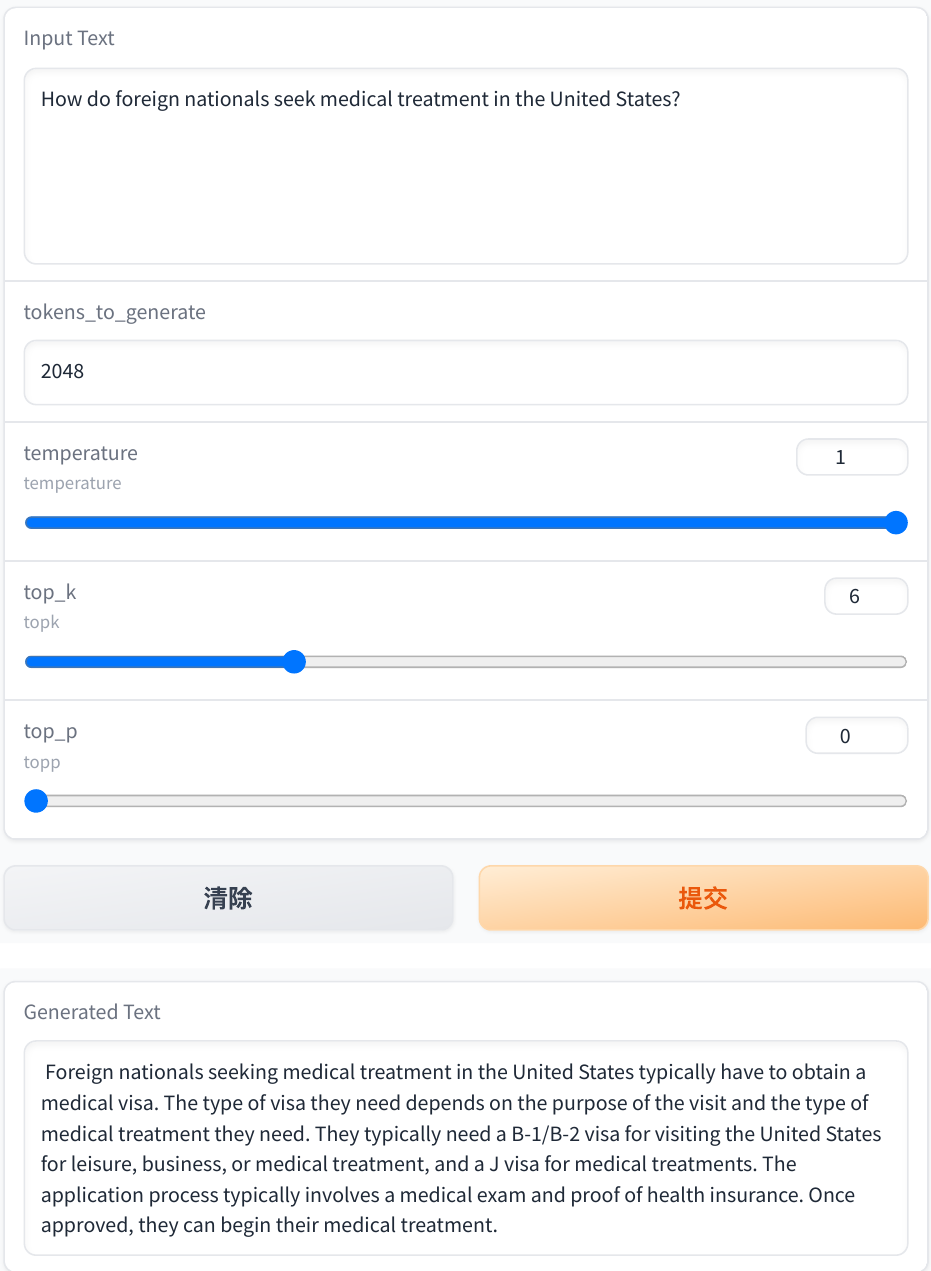}
\end{figure}

\begin{figure}[h]
    \centering
    \includegraphics[width=\textwidth]{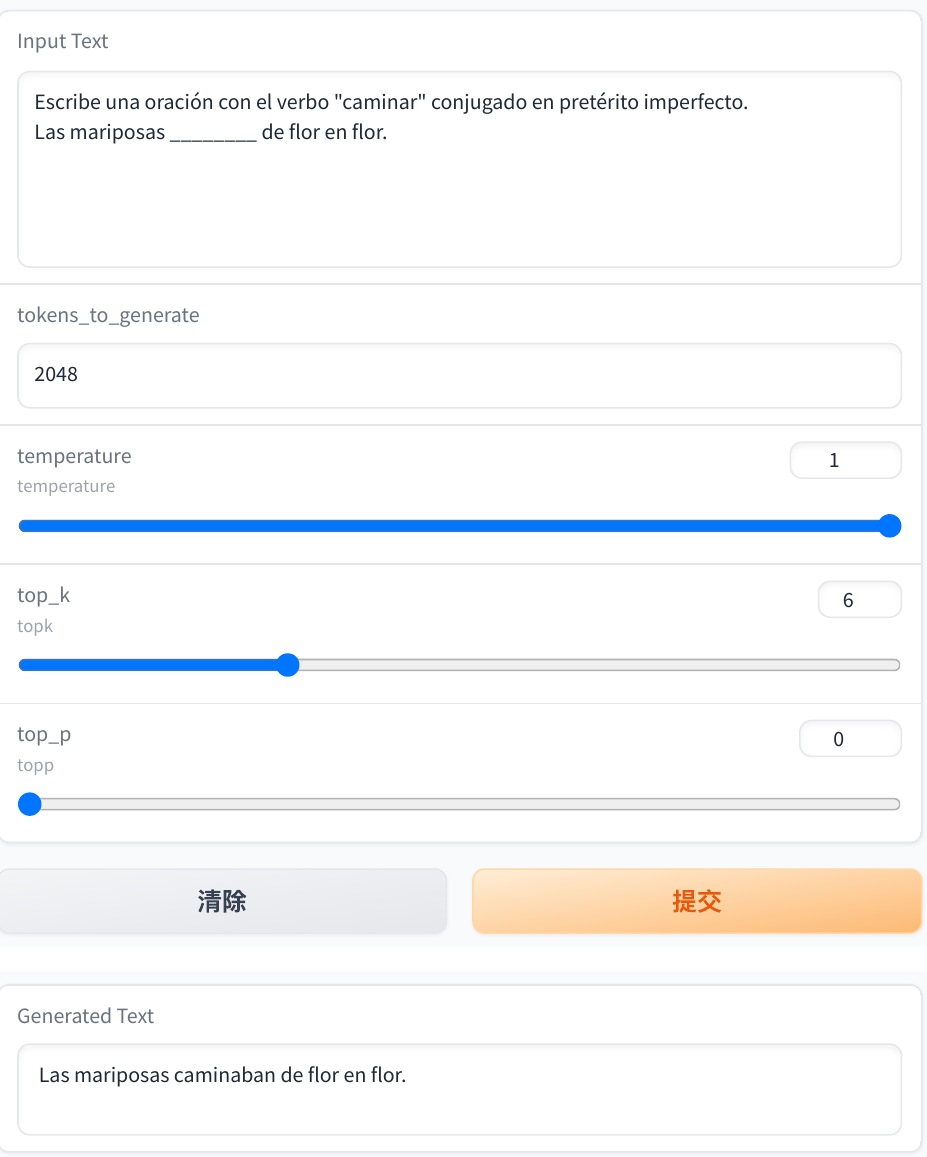}
\end{figure}

\begin{figure}[h]
    \centering
    \includegraphics[width=\textwidth]{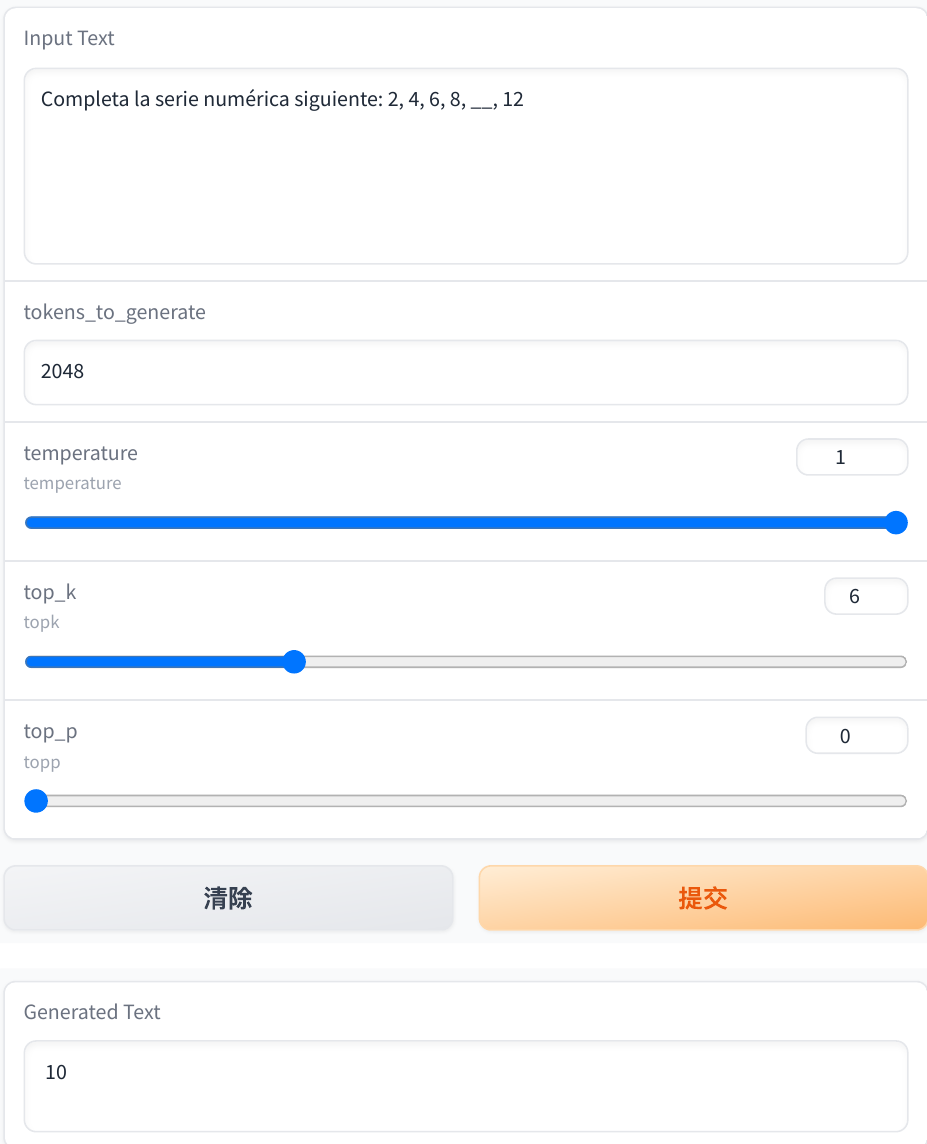}
\end{figure}

\begin{figure}[h]
    \centering
    \includegraphics[width=\textwidth]{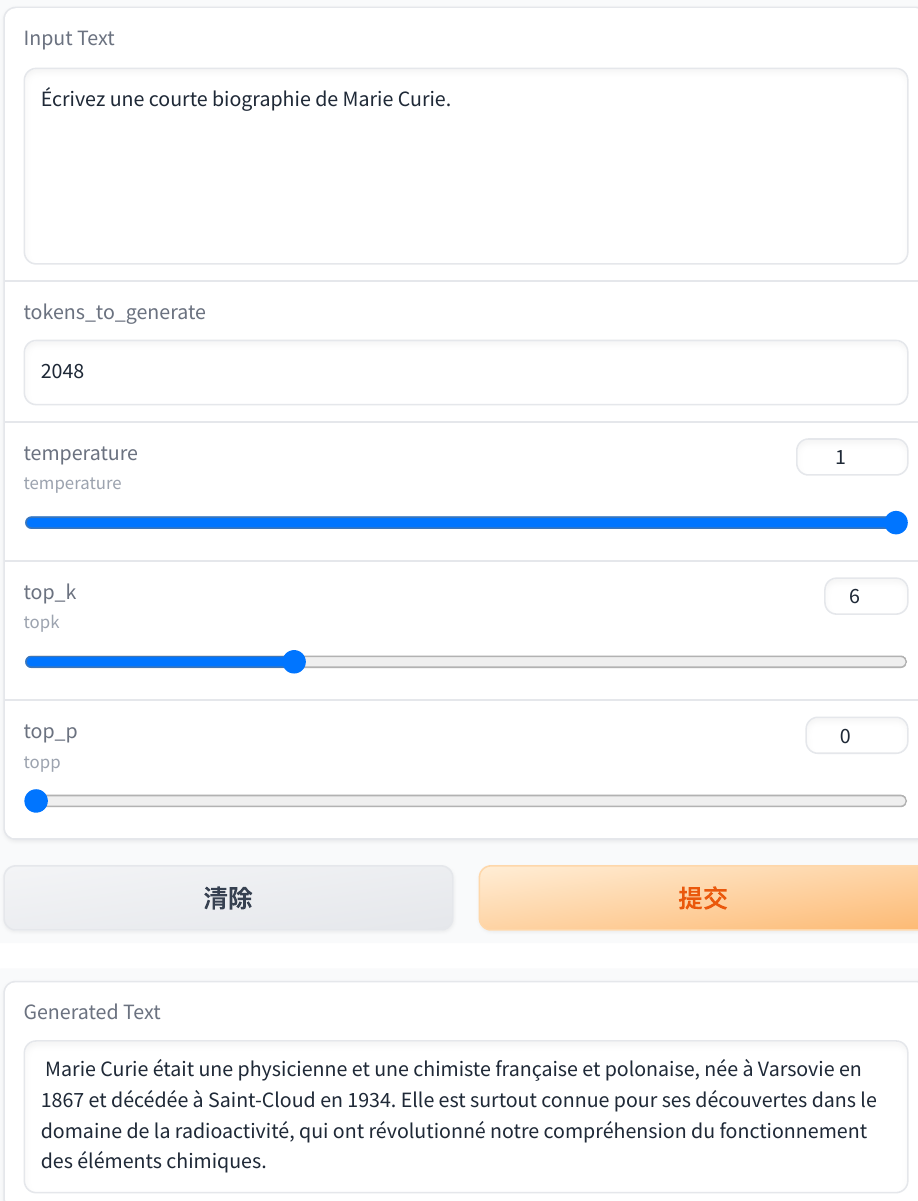}
\end{figure}

\begin{figure}[h]
    \centering
    \includegraphics[width=\textwidth]{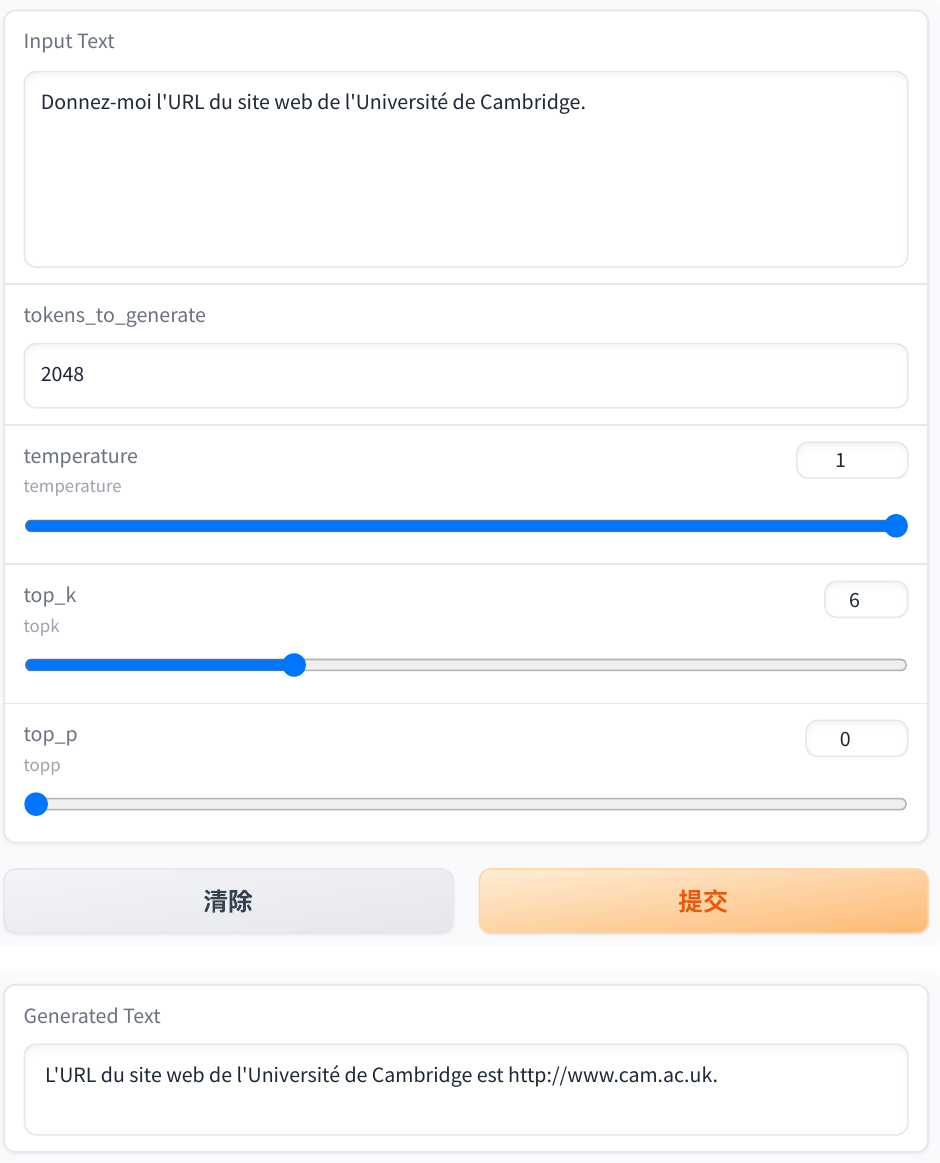}
\end{figure}

\begin{figure}[h]
    \centering
    \includegraphics[width=\textwidth]{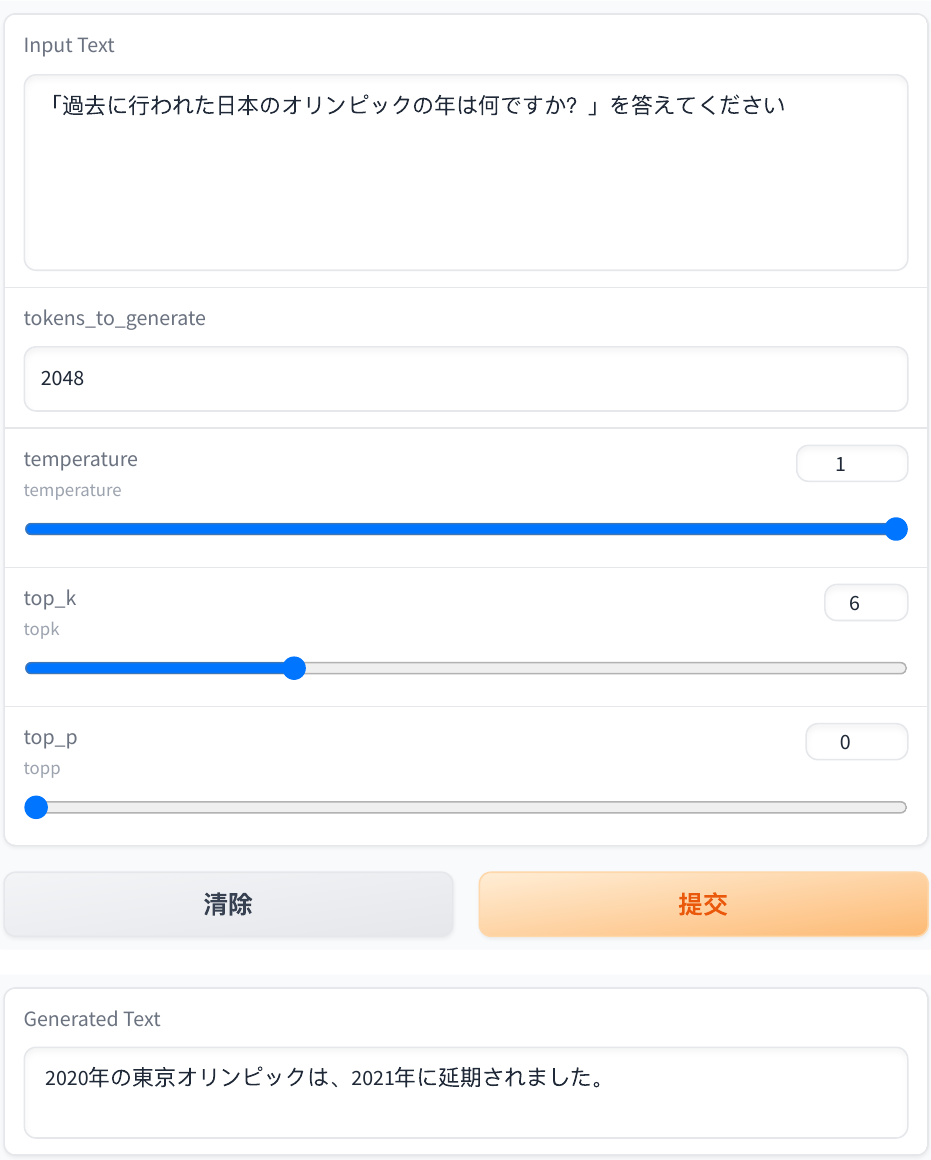}
\end{figure}

\begin{figure}[h]
    \centering
    \includegraphics[width=\textwidth]{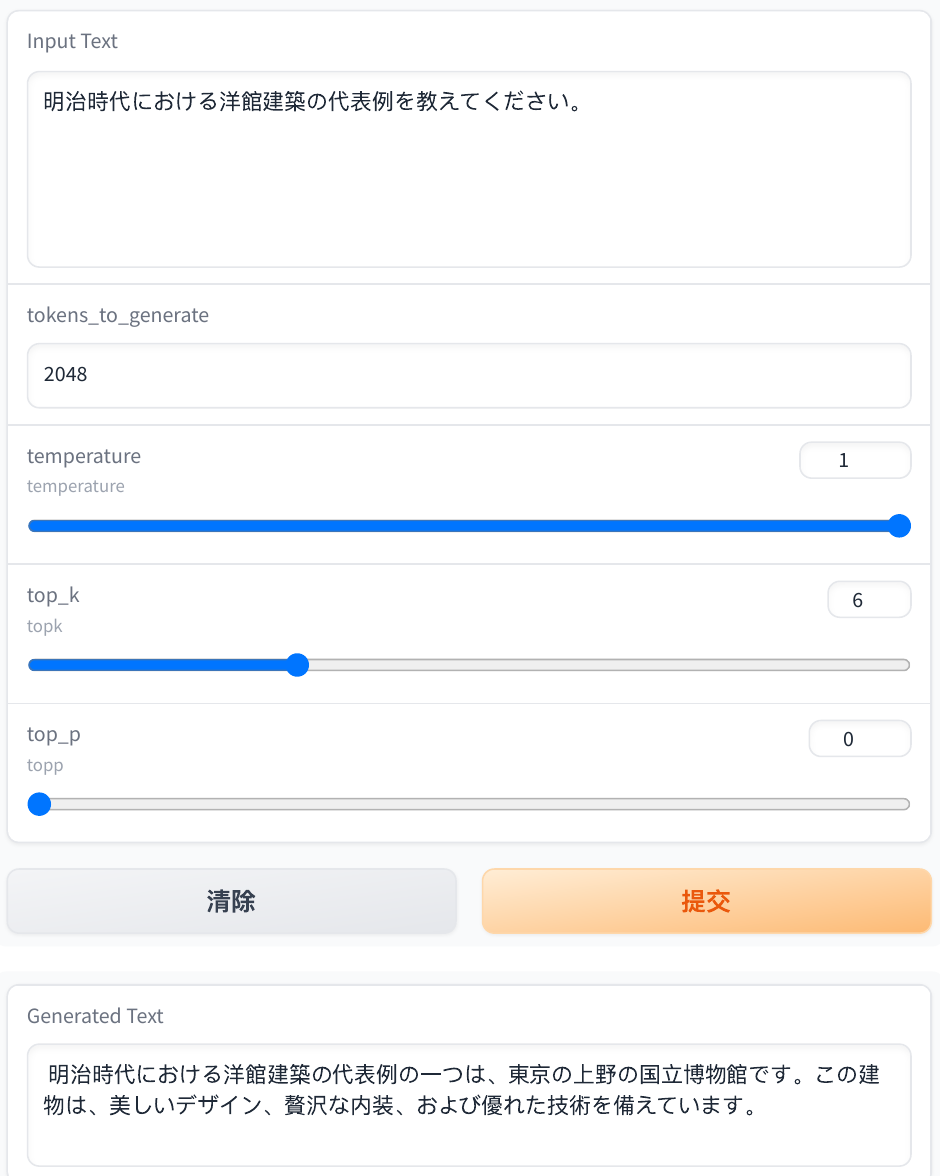}
\end{figure}

\begin{figure}[h]
    \centering
    \includegraphics[width=\textwidth]{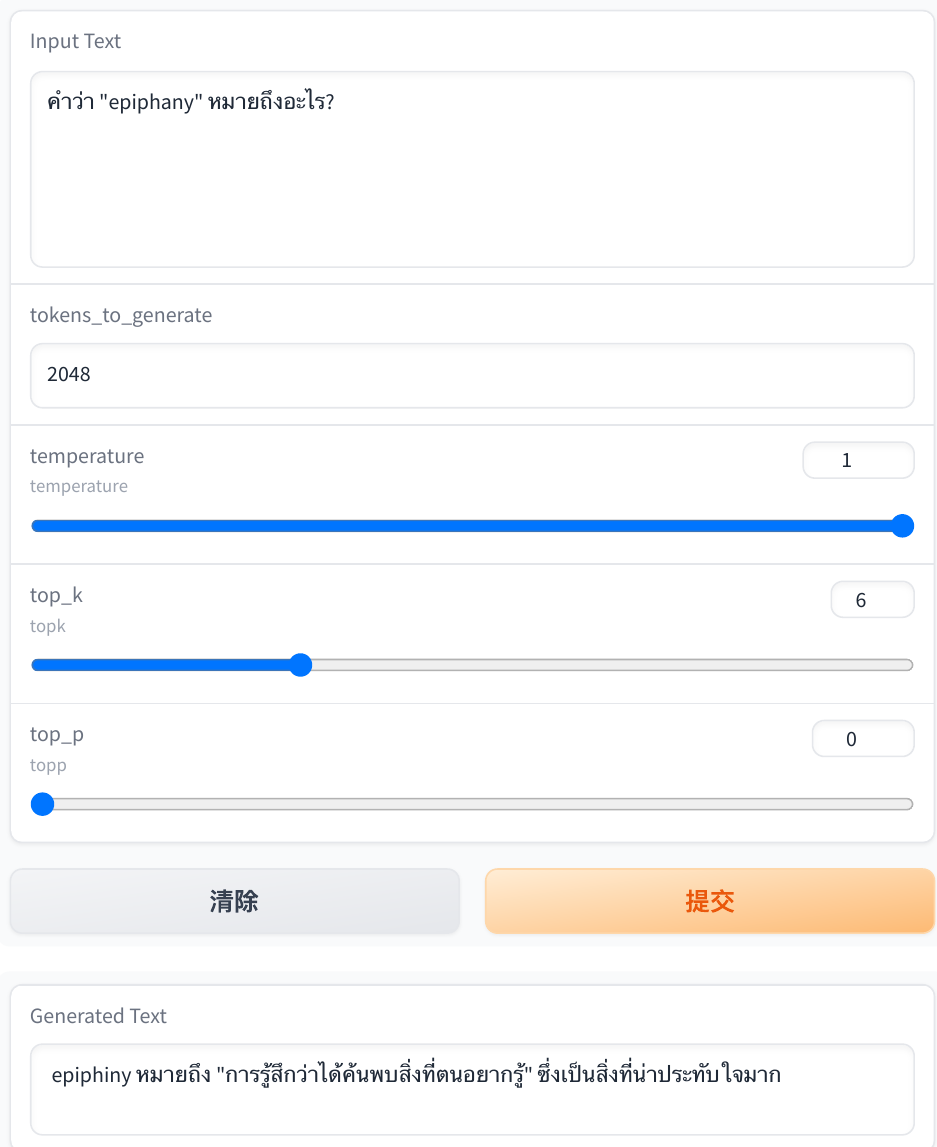}
\end{figure}

\begin{figure}[h]
    \centering
    \includegraphics[width=\textwidth]{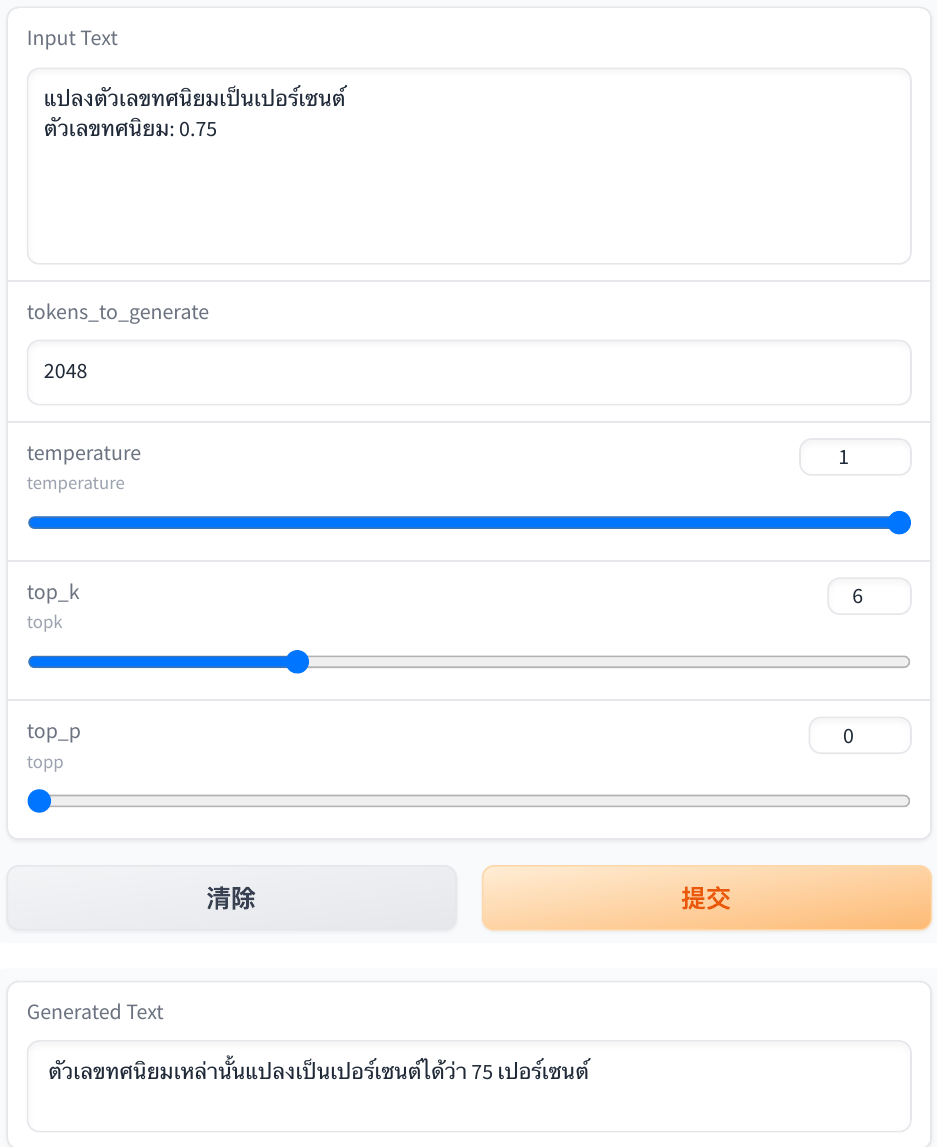}
\end{figure}

%% file: iclr2023_conference.bbl
\begin{thebibliography}{70}
\providecommand{\natexlab}[1]{#1}
\providecommand{\url}[1]{\texttt{#1}}
\expandafter\ifx\csname urlstyle\endcsname\relax
  \providecommand{\doi}[1]{doi: #1}\else
  \providecommand{\doi}{doi: \begingroup \urlstyle{rm}\Url}\fi

\bibitem[Ahuja et~al.(2023)Ahuja, Hada, Ochieng, Jain, Diddee, Maina, Ganu,
  Segal, Axmed, Bali, and Sitaram]{Ahuja2023MEGAME}
Kabir Ahuja, Rishav Hada, Millicent~A. Ochieng, Prachi Jain, Harshita Diddee,
  Samuel Maina, Tanuja Ganu, Sameer Segal, Maxamed Axmed, Kalika Bali, and
  Sunayana Sitaram.
\newblock Mega: Multilingual evaluation of generative ai.
\newblock \emph{ArXiv}, abs/2303.12528, 2023.

\bibitem[Anand et~al.(2023)Anand, Nussbaum, Duderstadt, Schmidt, and
  Mulyar]{gpt4all}
Yuvanesh Anand, Zach Nussbaum, Brandon Duderstadt, Benjamin Schmidt, and Andriy
  Mulyar.
\newblock Gpt4all: Training an assistant-style chatbot with large scale data
  distillation from gpt-3.5-turbo.
\newblock \url{https://github.com/nomic-ai/gpt4all}, 2023.

\bibitem[Anil et~al.(2023)Anil, Dai, Firat, Johnson, Lepikhin, Passos, Shakeri,
  Taropa, Bailey, Chen, et~al.]{anil2023palm}
Rohan Anil, Andrew~M Dai, Orhan Firat, Melvin Johnson, Dmitry Lepikhin,
  Alexandre Passos, Siamak Shakeri, Emanuel Taropa, Paige Bailey, Zhifeng Chen,
  et~al.
\newblock Palm 2 technical report.
\newblock \emph{arXiv preprint arXiv:2305.10403}, 2023.

\bibitem[Barrault et~al.(2020)Barrault, Biesialska, Bojar, Costa-juss{\`a},
  Federmann, Graham, Grundkiewicz, Haddow, Huck, Joanis, Kocmi, Koehn, Lo,
  Ljube{\v{s}}i{\'c}, Monz, Morishita, Nagata, Nakazawa, Pal, Post, and
  Zampieri]{barrault-etal-2020-findings}
Lo{\"\i}c Barrault, Magdalena Biesialska, Ond{\v{r}}ej Bojar, Marta~R.
  Costa-juss{\`a}, Christian Federmann, Yvette Graham, Roman Grundkiewicz,
  Barry Haddow, Matthias Huck, Eric Joanis, Tom Kocmi, Philipp Koehn, Chi-kiu
  Lo, Nikola Ljube{\v{s}}i{\'c}, Christof Monz, Makoto Morishita, Masaaki
  Nagata, Toshiaki Nakazawa, Santanu Pal, Matt Post, and Marcos Zampieri.
\newblock Findings of the 2020 conference on machine translation ({WMT}20).
\newblock In \emph{Proceedings of the Fifth Conference on Machine Translation},
  pp.\  1--55, Online, November 2020. Association for Computational
  Linguistics.
\newblock URL \url{https://aclanthology.org/2020.wmt-1.1}.

\bibitem[Bengio et~al.(2000)Bengio, Ducharme, and Vincent]{bengio2000neural}
Yoshua Bengio, R{\'e}jean Ducharme, and Pascal Vincent.
\newblock A neural probabilistic language model.
\newblock In \emph{Advances in neural information processing systems}, 2000.

\bibitem[Bengio et~al.(2009)Bengio, Louradour, Collobert, and
  Weston]{conf/icml/BengioLCW09}
Yoshua Bengio, J{\'{e}}r{\^{o}}me Louradour, Ronan Collobert, and Jason Weston.
\newblock Curriculum learning.
\newblock In \emph{Proceedings of the 26th Annual International Conference on
  Machine Learning, {ICML} 2009}, volume 382, pp.\  41--48. {ACM}, 2009.
\newblock URL \url{https://doi.org/10.1145/1553374.1553380}.

\bibitem[Biderman et~al.(2023)Biderman, Schoelkopf, Anthony, Bradley, O'Brien,
  Hallahan, Khan, Purohit, Prashanth, Raff, et~al.]{biderman2023pythia}
Stella Biderman, Hailey Schoelkopf, Quentin Anthony, Herbie Bradley, Kyle
  O'Brien, Eric Hallahan, Mohammad~Aflah Khan, Shivanshu Purohit, USVSN~Sai
  Prashanth, Edward Raff, et~al.
\newblock Pythia: A suite for analyzing large language models across training
  and scaling.
\newblock \emph{arXiv preprint arXiv:2304.01373}, 2023.

\bibitem[Brown et~al.(2020)Brown, Mann, Ryder, Subbiah, Kaplan, Dhariwal,
  Neelakantan, Shyam, Sastry, Askell, et~al.]{gpt3}
Tom Brown, Benjamin Mann, Nick Ryder, Melanie Subbiah, Jared~D Kaplan, Prafulla
  Dhariwal, Arvind Neelakantan, Pranav Shyam, Girish Sastry, Amanda Askell,
  et~al.
\newblock Language models are few-shot learners.
\newblock \emph{Advances in neural information processing systems},
  33:\penalty0 1877--1901, 2020.

\bibitem[Chen et~al.(2023)Chen, Wong, Chen, and Tian]{chen2023extending}
Shouyuan Chen, Sherman Wong, Liangjian Chen, and Yuandong Tian.
\newblock Extending context window of large language models via positional
  interpolation, 2023.

\bibitem[Chen et~al.(2021)Chen, Song, Wu, Wang, Xu, Chen, Zhou, and
  Li]{Chen2021MTGAB}
Yiran Chen, Zhenqiao Song, Xianze Wu, Danqing Wang, Jingjing Xu, Jiaze Chen,
  Hao Zhou, and Lei Li.
\newblock Mtg: A benchmarking suite for multilingual text generation.
\newblock In \emph{NAACL-HLT}, 2021.

\bibitem[Chowdhery et~al.(2022)Chowdhery, Narang, Devlin, Bosma, Mishra,
  Roberts, Barham, Chung, Sutton, Gehrmann, Schuh, Shi, Tsvyashchenko, Maynez,
  Rao, Barnes, Tay, Shazeer, Prabhakaran, Reif, Du, Hutchinson, Pope, Bradbury,
  Austin, Isard, Gur-Ari, Yin, Duke, Levskaya, Ghemawat, Dev, Michalewski,
  Garc{\'i}a, Misra, Robinson, Fedus, Zhou, Ippolito, Luan, Lim, Zoph,
  Spiridonov, Sepassi, Dohan, Agrawal, Omernick, Dai, Pillai, Pellat,
  Lewkowycz, Moreira, Child, Polozov, Lee, Zhou, Wang, Saeta, D{\'i}az, Firat,
  Catasta, Wei, Meier-Hellstern, Eck, Dean, Petrov, and
  Fiedel]{Chowdhery2022PaLMSL}
Aakanksha Chowdhery, Sharan Narang, Jacob Devlin, Maarten Bosma, Gaurav Mishra,
  Adam Roberts, Paul Barham, Hyung~Won Chung, Charles Sutton, Sebastian
  Gehrmann, Parker Schuh, Kensen Shi, Sasha Tsvyashchenko, Joshua Maynez,
  Abhishek Rao, Parker Barnes, Yi~Tay, Noam~M. Shazeer, Vinodkumar Prabhakaran,
  Emily Reif, Nan Du, Benton~C. Hutchinson, Reiner Pope, James Bradbury, Jacob
  Austin, Michael Isard, Guy Gur-Ari, Pengcheng Yin, Toju Duke, Anselm
  Levskaya, Sanjay Ghemawat, Sunipa Dev, Henryk Michalewski, Xavier Garc{\'i}a,
  Vedant Misra, Kevin Robinson, Liam Fedus, Denny Zhou, Daphne Ippolito, David
  Luan, Hyeontaek Lim, Barret Zoph, Alexander Spiridonov, Ryan Sepassi, David
  Dohan, Shivani Agrawal, Mark Omernick, Andrew~M. Dai,
  Thanumalayan~Sankaranarayana Pillai, Marie Pellat, Aitor Lewkowycz, Erica
  Moreira, Rewon Child, Oleksandr Polozov, Katherine Lee, Zongwei Zhou, Xuezhi
  Wang, Brennan Saeta, Mark D{\'i}az, Orhan Firat, Michele Catasta, Jason Wei,
  Kathleen~S. Meier-Hellstern, Douglas Eck, Jeff Dean, Slav Petrov, and Noah
  Fiedel.
\newblock Palm: Scaling language modeling with pathways.
\newblock \emph{ArXiv}, abs/2204.02311, 2022.

\bibitem[Clark et~al.(2020)Clark, Choi, Collins, Garrette, Kwiatkowski,
  Nikolaev, and Palomaki]{tydiqa}
Jonathan~H. Clark, Eunsol Choi, Michael Collins, Dan Garrette, Tom Kwiatkowski,
  Vitaly Nikolaev, and Jennimaria Palomaki.
\newblock Tydi qa: A benchmark for information-seeking question answering in
  typologically diverse languages.
\newblock \emph{Transactions of the Association for Computational Linguistics},
  2020.

\bibitem[Conneau et~al.(2019)Conneau, Khandelwal, Goyal, Chaudhary, Wenzek,
  Guzm{\'a}n, Grave, Ott, Zettlemoyer, and Stoyanov]{Conneau2019UnsupervisedCR}
Alexis Conneau, Kartikay Khandelwal, Naman Goyal, Vishrav Chaudhary, Guillaume
  Wenzek, Francisco Guzm{\'a}n, Edouard Grave, Myle Ott, Luke Zettlemoyer, and
  Veselin Stoyanov.
\newblock Unsupervised cross-lingual representation learning at scale.
\newblock In \emph{ACL}, 2019.

\bibitem[Cui et~al.(2023)Cui, Yang, and Yao]{cui2023efficient}
Yiming Cui, Ziqing Yang, and Xin Yao.
\newblock Efficient and effective text encoding for chinese llama and alpaca.
\newblock \emph{arXiv preprint arXiv:2304.08177}, 2023.

\bibitem[Devlin et~al.(2019)Devlin, Chang, Lee, and
  Toutanova]{devlin-etal-2019-bert}
Jacob Devlin, Ming-Wei Chang, Kenton Lee, and Kristina Toutanova.
\newblock {BERT}: Pre-training of deep bidirectional transformers for language
  understanding.
\newblock In \emph{Proceedings of the 2019 Conference of the North {A}merican
  Chapter of the Association for Computational Linguistics: Human Language
  Technologies, Volume 1 (Long and Short Papers)}, pp.\  4171--4186,
  Minneapolis, Minnesota, June 2019. Association for Computational Linguistics.
\newblock \doi{10.18653/v1/N19-1423}.
\newblock URL \url{https://aclanthology.org/N19-1423}.

\bibitem[Gao et~al.(2020)Gao, Biderman, Black, Golding, Hoppe, Foster, Phang,
  He, Thite, Nabeshima, et~al.]{gao2020pile}
Leo Gao, Stella Biderman, Sid Black, Laurence Golding, Travis Hoppe, Charles
  Foster, Jason Phang, Horace He, Anish Thite, Noa Nabeshima, et~al.
\newblock The pile: An 800gb dataset of diverse text for language modeling.
\newblock \emph{arXiv preprint arXiv:2101.00027}, 2020.

\bibitem[Gao et~al.(2021)Gao, Tow, Biderman, Black, DiPofi, Foster, Golding,
  Hsu, McDonell, Muennighoff, Phang, Reynolds, Tang, Thite, Wang, Wang, and
  Zou]{eval-harness}
Leo Gao, Jonathan Tow, Stella Biderman, Sid Black, Anthony DiPofi, Charles
  Foster, Laurence Golding, Jeffrey Hsu, Kyle McDonell, Niklas Muennighoff,
  Jason Phang, Laria Reynolds, Eric Tang, Anish Thite, Ben Wang, Kevin Wang,
  and Andy Zou.
\newblock A framework for few-shot language model evaluation, Sep 2021.
\newblock URL \url{https://doi.org/10.5281/zenodo.5371628}.

\bibitem[Glorot \& Bengio(2010)Glorot and Bengio]{Glorot2010UnderstandingTD}
Xavier Glorot and Yoshua Bengio.
\newblock Understanding the difficulty of training deep feedforward neural
  networks.
\newblock In \emph{International Conference on Artificial Intelligence and
  Statistics}, 2010.

\bibitem[Gordon et~al.(2011)Gordon, Kozareva, and
  Roemmele]{Gordon2011SemEval2012T7}
Andrew~S. Gordon, Zornitsa Kozareva, and Melissa Roemmele.
\newblock Semeval-2012 task 7: Choice of plausible alternatives: An evaluation
  of commonsense causal reasoning.
\newblock In \emph{International Workshop on Semantic Evaluation}, 2011.

\bibitem[Heafield(2011)]{heafield2011kenlm}
Kenneth Heafield.
\newblock Kenlm: Faster and smaller language model queries.
\newblock In \emph{Proceedings of the sixth workshop on statistical machine
  translation}, pp.\  187--197, 2011.

\bibitem[Hendrycks \& Gimpel(2016)Hendrycks and
  Gimpel]{Hendrycks2016GaussianEL}
Dan Hendrycks and Kevin Gimpel.
\newblock Gaussian error linear units (gelus).
\newblock \emph{arXiv: Learning}, 2016.

\bibitem[Hoffmann et~al.(2022)Hoffmann, Borgeaud, Mensch, Buchatskaya, Cai,
  Rutherford, Casas, Hendricks, Welbl, Clark, et~al.]{hoffmann2022chinchilla}
Jordan Hoffmann, Sebastian Borgeaud, Arthur Mensch, Elena Buchatskaya, Trevor
  Cai, Eliza Rutherford, Diego de~Las Casas, Lisa~Anne Hendricks, Johannes
  Welbl, Aidan Clark, et~al.
\newblock Training compute-optimal large language models.
\newblock \emph{arXiv preprint arXiv:2203.15556}, 2022.

\bibitem[Jaegle et~al.(2021)Jaegle, Gimeno, Brock, Vinyals, Zisserman, and
  Carreira]{conf/icml/JaegleGBVZC21}
Andrew Jaegle, Felix Gimeno, Andy Brock, Oriol Vinyals, Andrew Zisserman, and
  Jo{\~{a}}o Carreira.
\newblock Perceiver: General perception with iterative attention.
\newblock In \emph{Proceedings of the 38th International Conference on Machine
  Learning, {ICML} 2021}, volume 139 of \emph{Proceedings of Machine Learning
  Research}, pp.\  4651--4664. {PMLR}, 2021.
\newblock URL \url{http://proceedings.mlr.press/v139/jaegle21a.html}.

\bibitem[Joulin et~al.(2016)Joulin, Grave, Bojanowski, Douze, J{\'e}gou, and
  Mikolov]{joulin2016fasttext}
Armand Joulin, Edouard Grave, Piotr Bojanowski, Matthijs Douze, H{\'e}rve
  J{\'e}gou, and Tomas Mikolov.
\newblock Fasttext. zip: Compressing text classification models.
\newblock \emph{arXiv preprint arXiv:1612.03651}, 2016.

\bibitem[Kaplan et~al.(2020)Kaplan, McCandlish, Henighan, Brown, Chess, Child,
  Gray, Radford, Wu, and Amodei]{kaplan2020scaling}
Jared Kaplan, Sam McCandlish, Tom Henighan, Tom~B Brown, Benjamin Chess, Rewon
  Child, Scott Gray, Alec Radford, Jeffrey Wu, and Dario Amodei.
\newblock Scaling laws for neural language models.
\newblock \emph{arXiv preprint arXiv:2001.08361}, 2020.

\bibitem[Kudo \& Richardson(2018)Kudo and Richardson]{kudo2018sentencepiece}
Taku Kudo and John Richardson.
\newblock Sentencepiece: A simple and language independent subword tokenizer
  and detokenizer for neural text processing.
\newblock \emph{arXiv preprint arXiv:1808.06226}, 2018.

\bibitem[Kumar et~al.(2010)Kumar, Packer, and Koller]{conf/nips/KumarPK10}
M.~Pawan Kumar, Benjamin Packer, and Daphne Koller.
\newblock Self-paced learning for latent variable models.
\newblock In \emph{Advances in Neural Information Processing Systems 23: 24th
  Annual Conference on Neural Information Processing Systems 2010}, pp.\
  1189--1197. Curran Associates, Inc., 2010.
\newblock URL
  \url{https://proceedings.neurips.cc/paper/2010/hash/e57c6b956a6521b28495f2886ca0977a-Abstract.html}.

\bibitem[Lewis et~al.(2020)Lewis, Oguz, Rinott, Riedel, and Schwenk]{MLQA}
Patrick Lewis, Barlas Oguz, Ruty Rinott, Sebastian Riedel, and Holger Schwenk.
\newblock {MLQA}: Evaluating cross-lingual extractive question answering.
\newblock In \emph{Proceedings of the 58th Annual Meeting of the Association
  for Computational Linguistics}, 2020.

\bibitem[Lin et~al.(2022)Lin, Mihaylov, Artetxe, Wang, Chen, and
  Lopez]{lin-etal-2022-xglm}
Xi~Victoria Lin, Todor Mihaylov, Mikel Artetxe, Tianlu Wang, Shuohui Chen, and
  Adam Lopez.
\newblock Few-shot learning with multilingual generative language models.
\newblock In \emph{Proceedings of the 2022 Conference on Empirical Methods in
  Natural Language Processing}, pp.\  9019--9052, 2022.

\bibitem[Longpre et~al.(2020)Longpre, Lu, and Daiber]{Longpre2020MKQAAL}
S.~Longpre, Yi~Lu, and Joachim Daiber.
\newblock Mkqa: A linguistically diverse benchmark for multilingual open domain
  question answering.
\newblock \emph{Transactions of the Association for Computational Linguistics},
  9:\penalty0 1389--1406, 2020.

\bibitem[Longpre et~al.(2023)Longpre, Hou, Vu, Webson, Chung, Tay, Zhou, Le,
  Zoph, Wei, et~al.]{longpre2023flan}
Shayne Longpre, Le~Hou, Tu~Vu, Albert Webson, Hyung~Won Chung, Yi~Tay, Denny
  Zhou, Quoc~V Le, Barret Zoph, Jason Wei, et~al.
\newblock The flan collection: Designing data and methods for effective
  instruction tuning.
\newblock \emph{arXiv preprint arXiv:2301.13688}, 2023.

\bibitem[Mikolov et~al.(2010)Mikolov, Karafiát, Burget, Cernocký, and
  Khudanpur]{conf/interspeech/MikolovKBCK10}
Tomas Mikolov, Martin Karafiát, Lukás Burget, Jan Cernocký, and Sanjeev
  Khudanpur.
\newblock Recurrent neural network based language model.
\newblock In Takao Kobayashi, Keikichi Hirose, and Satoshi Nakamura (eds.),
  \emph{INTERSPEECH}, pp.\  1045--1048. ISCA, 2010.
\newblock URL
  \url{http://dblp.uni-trier.de/db/conf/interspeech/interspeech2010.html#MikolovKBCK10}.

\bibitem[Muennighoff et~al.(2022)Muennighoff, Wang, Sutawika, Roberts,
  Biderman, Scao, Bari, Shen, Yong, Schoelkopf,
  et~al.]{muennighoff2022crosslingual}
Niklas Muennighoff, Thomas Wang, Lintang Sutawika, Adam Roberts, Stella
  Biderman, Teven~Le Scao, M~Saiful Bari, Sheng Shen, Zheng-Xin Yong, Hailey
  Schoelkopf, et~al.
\newblock Crosslingual generalization through multitask finetuning.
\newblock \emph{arXiv preprint arXiv:2211.01786}, 2022.

\bibitem[OpenAI(2023)]{openai2023gpt4}
OpenAI.
\newblock Gpt-4 technical report.
\newblock \emph{arXiv preprint arXiv:2303.08774}, 2023.

\bibitem[Ouyang et~al.(2022)Ouyang, Wu, Jiang, Almeida, Wainwright, Mishkin,
  Zhang, Agarwal, Slama, Gray, Schulman, Hilton, Kelton, Miller, Simens,
  Askell, Welinder, Christiano, Leike, and Lowe]{ouyang2022training}
Long Ouyang, Jeffrey Wu, Xu~Jiang, Diogo Almeida, Carroll Wainwright, Pamela
  Mishkin, Chong Zhang, Sandhini Agarwal, Katarina Slama, Alex Gray, John
  Schulman, Jacob Hilton, Fraser Kelton, Luke Miller, Maddie Simens, Amanda
  Askell, Peter Welinder, Paul Christiano, Jan Leike, and Ryan Lowe.
\newblock Training language models to follow instructions with human feedback.
\newblock In Alice~H. Oh, Alekh Agarwal, Danielle Belgrave, and Kyunghyun Cho
  (eds.), \emph{Advances in Neural Information Processing Systems}, 2022.
\newblock URL \url{https://openreview.net/forum?id=TG8KACxEON}.

\bibitem[Papineni et~al.(2002)Papineni, Roukos, Ward, and
  Zhu]{Papineni2002BleuAM}
Kishore Papineni, Salim Roukos, Todd Ward, and Wei-Jing Zhu.
\newblock Bleu: a method for automatic evaluation of machine translation.
\newblock In \emph{Annual Meeting of the Association for Computational
  Linguistics}, 2002.

\bibitem[Penedo et~al.(2023)Penedo, Malartic, Hesslow, Cojocaru, Cappelli,
  Alobeidli, Pannier, Almazrouei, and Launay]{penedo2023refinedweb}
Guilherme Penedo, Quentin Malartic, Daniel Hesslow, Ruxandra Cojocaru,
  Alessandro Cappelli, Hamza Alobeidli, Baptiste Pannier, Ebtesam Almazrouei,
  and Julien Launay.
\newblock The refinedweb dataset for falcon llm: Outperforming curated corpora
  with web data, and web data only.
\newblock \emph{arXiv preprint arXiv:2306.01116}, 2023.

\bibitem[Peng et~al.(2023)Peng, Li, He, Galley, and Gao]{peng2023instruction}
Baolin Peng, Chunyuan Li, Pengcheng He, Michel Galley, and Jianfeng Gao.
\newblock Instruction tuning with gpt-4.
\newblock \emph{arXiv preprint arXiv:2304.03277}, 2023.

\bibitem[Peters et~al.(2018)Peters, Neumann, Iyyer, Gardner, Clark, Lee, and
  Zettlemoyer]{peters-etal-2018-deep}
Matthew~E. Peters, Mark Neumann, Mohit Iyyer, Matt Gardner, Christopher Clark,
  Kenton Lee, and Luke Zettlemoyer.
\newblock Deep contextualized word representations.
\newblock In \emph{Proceedings of the 2018 Conference of the North {A}merican
  Chapter of the Association for Computational Linguistics: Human Language
  Technologies, Volume 1 (Long Papers)}, pp.\  2227--2237, New Orleans,
  Louisiana, June 2018. Association for Computational Linguistics.
\newblock \doi{10.18653/v1/N18-1202}.
\newblock URL \url{https://aclanthology.org/N18-1202}.

\bibitem[Ponti et~al.(2020)Ponti, {s}, Majewska, Liu, Vuli'{c}, and
  Korhonen]{ponti2020xcopa}
Edoardo~M. Ponti, Goran~Glava {s}, Olga Majewska, Qianchu Liu, Ivan Vuli'{c},
  and Anna Korhonen.
\newblock {XCOPA: A} multilingual dataset for causal commonsense reasoning.
\newblock \emph{arXiv preprint}, 2020.
\newblock URL \url{https://ducdauge.github.io/files/xcopa.pdf}.

\bibitem[Post(2018)]{post-2018-call}
Matt Post.
\newblock A call for clarity in reporting {BLEU} scores.
\newblock In \emph{Proceedings of the Third Conference on Machine Translation:
  Research Papers}, pp.\  186--191, Belgium, Brussels, October 2018.
  Association for Computational Linguistics.
\newblock URL \url{https://www.aclweb.org/anthology/W18-6319}.

\bibitem[Press et~al.(2022)Press, Smith, and Lewis]{press2022alibi}
Ofir Press, Noah~A. Smith, and Mike Lewis.
\newblock Train short, test long: Attention with linear biases enables input
  length extrapolation, 2022.

\bibitem[Radford et~al.(2018)Radford, Narasimhan, Salimans, and
  Sutskever]{radford2018improving}
Alec Radford, Karthik Narasimhan, Tim Salimans, and Ilya Sutskever.
\newblock Improving language understanding by generative pre-training.
\newblock \emph{URL https://s3-us-west-2. amazonaws.
  com/openai-assets/researchcovers/languageunsupervised/language understanding
  paper. pdf}, 2018.

\bibitem[Radford et~al.(2019)Radford, Wu, Child, Luan, Amodei, Sutskever,
  et~al.]{radford2019language}
Alec Radford, Jeffrey Wu, Rewon Child, David Luan, Dario Amodei, Ilya
  Sutskever, et~al.
\newblock Language models are unsupervised multitask learners.
\newblock \emph{OpenAI blog}, 1\penalty0 (8):\penalty0 9, 2019.

\bibitem[Rae et~al.(2021)Rae, Borgeaud, Cai, Millican, Hoffmann, Song,
  Aslanides, Henderson, Ring, Young, et~al.]{rae2021scaling}
Jack~W Rae, Sebastian Borgeaud, Trevor Cai, Katie Millican, Jordan Hoffmann,
  Francis Song, John Aslanides, Sarah Henderson, Roman Ring, Susannah Young,
  et~al.
\newblock Scaling language models: Methods, analysis \& insights from training
  gopher.
\newblock \emph{arXiv preprint arXiv:2112.11446}, 2021.

\bibitem[Raffel et~al.(2020)Raffel, Shazeer, Roberts, Lee, Narang, Matena,
  Zhou, Li, and Liu]{raffel2020exploring}
Colin Raffel, Noam Shazeer, Adam Roberts, Katherine Lee, Sharan Narang, Michael
  Matena, Yanqi Zhou, Wei Li, and Peter~J Liu.
\newblock Exploring the limits of transfer learning with a unified text-to-text
  transformer.
\newblock \emph{The Journal of Machine Learning Research}, 21\penalty0
  (1):\penalty0 5485--5551, 2020.

\bibitem[Scao et~al.(2022)Scao, Fan, Akiki, Pavlick, Ili{\'c}, Hesslow,
  Castagn{\'e}, Luccioni, Yvon, Gall{\'e}, et~al.]{scao2022bloom}
Teven~Le Scao, Angela Fan, Christopher Akiki, Ellie Pavlick, Suzana Ili{\'c},
  Daniel Hesslow, Roman Castagn{\'e}, Alexandra~Sasha Luccioni, Fran{\c{c}}ois
  Yvon, Matthias Gall{\'e}, et~al.
\newblock Bloom: A 176b-parameter open-access multilingual language model.
\newblock \emph{arXiv preprint arXiv:2211.05100}, 2022.

\bibitem[Sennrich et~al.(2015)Sennrich, Haddow, and Birch]{sennrich2015neural}
Rico Sennrich, Barry Haddow, and Alexandra Birch.
\newblock Neural machine translation of rare words with subword units.
\newblock \emph{arXiv preprint arXiv:1508.07909}, 2015.

\bibitem[Shliazhko et~al.(2022)Shliazhko, Fenogenova, Tikhonova, Mikhailov,
  Kozlova, and Shavrina]{shliazhko2022mgpt}
Oleh Shliazhko, Alena Fenogenova, Maria Tikhonova, Vladislav Mikhailov,
  Anastasia Kozlova, and Tatiana Shavrina.
\newblock mgpt: Few-shot learners go multilingual, 2022.

\bibitem[Shoeybi et~al.(2020)Shoeybi, Patwary, Puri, LeGresley, Casper, and
  Catanzaro]{shoeybi2020megatronlm}
Mohammad Shoeybi, Mostofa Patwary, Raul Puri, Patrick LeGresley, Jared Casper,
  and Bryan Catanzaro.
\newblock Megatron-lm: Training multi-billion parameter language models using
  model parallelism, 2020.

\bibitem[Smith et~al.(2022)Smith, Patwary, Norick, LeGresley, Rajbhandari,
  Casper, Liu, Prabhumoye, Zerveas, Korthikanti, et~al.]{smith2022using}
Shaden Smith, Mostofa Patwary, Brandon Norick, Patrick LeGresley, Samyam
  Rajbhandari, Jared Casper, Zhun Liu, Shrimai Prabhumoye, George Zerveas,
  Vijay Korthikanti, et~al.
\newblock Using deepspeed and megatron to train megatron-turing nlg 530b, a
  large-scale generative language model.
\newblock \emph{arXiv preprint arXiv:2201.11990}, 2022.

\bibitem[Su et~al.(2021)Su, Lu, Pan, Murtadha, Wen, and Liu]{su2021roformer}
Jianlin Su, Yu~Lu, Shengfeng Pan, Ahmed Murtadha, Bo~Wen, and Yunfeng Liu.
\newblock Roformer: Enhanced transformer with rotary position embedding.
\newblock \emph{CoRR}, abs/2104.09864, 2021.
\newblock URL \url{https://arxiv.org/abs/2104.09864}.

\bibitem[Taori et~al.(2023)Taori, Gulrajani, Zhang, Dubois, Li, Guestrin,
  Liang, and Hashimoto]{taori2023alpaca}
Rohan Taori, Ishaan Gulrajani, Tianyi Zhang, Yann Dubois, Xuechen Li, Carlos
  Guestrin, Percy Liang, and Tatsunori~B. Hashimoto.
\newblock Stanford alpaca: An instruction-following llama model.
\newblock \url{https://github.com/tatsu-lab/stanford_alpaca}, 2023.

\bibitem[Tikhonov \& Ryabinin(2021)Tikhonov and Ryabinin]{tikhonov2021heads}
Alexey Tikhonov and Max Ryabinin.
\newblock It's all in the heads: Using attention heads as a baseline for
  cross-lingual transfer in commonsense reasoning, 2021.

\bibitem[Touvron et~al.(2023)Touvron, Lavril, Izacard, Martinet, Lachaux,
  Lacroix, Rozi{\`e}re, Goyal, Hambro, Azhar, et~al.]{touvron2023llama}
Hugo Touvron, Thibaut Lavril, Gautier Izacard, Xavier Martinet, Marie-Anne
  Lachaux, Timoth{\'e}e Lacroix, Baptiste Rozi{\`e}re, Naman Goyal, Eric
  Hambro, Faisal Azhar, et~al.
\newblock Llama: Open and efficient foundation language models.
\newblock \emph{arXiv preprint arXiv:2302.13971}, 2023.

\bibitem[Vaswani et~al.(2017)Vaswani, Shazeer, Parmar, Uszkoreit, Jones, Gomez,
  Kaiser, and Polosukhin]{Vaswani2017Attention}
Ashish Vaswani, Noam Shazeer, Niki Parmar, Jakob Uszkoreit, Llion Jones,
  Aidan~N Gomez, Lukasz Kaiser, and Illia Polosukhin.
\newblock Attention is all you need.
\newblock In \emph{Advances in Neural Information Processing Systems 30, {NIPS}
  2017 4-9 December 2017, Long Beach, CA, {USA}}, pp.\  5998--6008, 2017.
\newblock URL \url{http://papers.nips.cc/paper/7181-attention-is-all-you-need}.

\bibitem[Wang et~al.(2018)Wang, Singh, Michael, Hill, Levy, and
  Bowman]{Wang2018GLUEAM}
Alex Wang, Amanpreet Singh, Julian Michael, Felix Hill, Omer Levy, and
  Samuel~R. Bowman.
\newblock Glue: A multi-task benchmark and analysis platform for natural
  language understanding.
\newblock \emph{ArXiv}, abs/1804.07461, 2018.

\bibitem[Wang et~al.(2019)Wang, Pruksachatkun, Nangia, Singh, Michael, Hill,
  Levy, and Bowman]{Wang2019SuperGLUEAS}
Alex Wang, Yada Pruksachatkun, Nikita Nangia, Amanpreet Singh, Julian Michael,
  Felix Hill, Omer Levy, and Samuel~R. Bowman.
\newblock Superglue: A stickier benchmark for general-purpose language
  understanding systems.
\newblock In \emph{Neural Information Processing Systems}, 2019.

\bibitem[Wang et~al.(2022)Wang, Kordi, Mishra, Liu, Smith, Khashabi, and
  Hajishirzi]{wang2022self}
Yizhong Wang, Yeganeh Kordi, Swaroop Mishra, Alisa Liu, Noah~A Smith, Daniel
  Khashabi, and Hannaneh Hajishirzi.
\newblock Self-instruct: Aligning language model with self generated
  instructions.
\newblock \emph{arXiv preprint arXiv:2212.10560}, 2022.

\bibitem[Wang et~al.(2023)Wang, Ivison, Dasigi, Hessel, Khot, Chandu, Wadden,
  MacMillan, Smith, Beltagy, and Hajishirzi]{wang2023far}
Yizhong Wang, Hamish Ivison, Pradeep Dasigi, Jack Hessel, Tushar Khot,
  Khyathi~Raghavi Chandu, David Wadden, Kelsey MacMillan, Noah~A. Smith,
  Iz~Beltagy, and Hannaneh Hajishirzi.
\newblock How far can camels go? exploring the state of instruction tuning on
  open resources, 2023.

\bibitem[Wei et~al.(2022)Wei, Bosma, Zhao, Guu, Yu, Lester, Du, Dai, and
  Le]{wei2022finetuned}
Jason Wei, Maarten Bosma, Vincent Zhao, Kelvin Guu, Adams~Wei Yu, Brian Lester,
  Nan Du, Andrew~M. Dai, and Quoc~V Le.
\newblock Finetuned language models are zero-shot learners.
\newblock In \emph{International Conference on Learning Representations}, 2022.
\newblock URL \url{https://openreview.net/forum?id=gEZrGCozdqR}.

\bibitem[Xiong et~al.(2020)Xiong, Yang, He, Zheng, Zheng, Xing, Zhang, Lan,
  Wang, and Liu]{xiong2020layer}
Ruibin Xiong, Yunchang Yang, Di~He, Kai Zheng, Shuxin Zheng, Chen Xing,
  Huishuai Zhang, Yanyan Lan, Liwei Wang, and Tieyan Liu.
\newblock On layer normalization in the transformer architecture.
\newblock In \emph{International Conference on Machine Learning}, pp.\
  10524--10533. PMLR, 2020.

\bibitem[Xu et~al.(2023)Xu, Sun, Zheng, Geng, Zhao, Feng, Tao, and
  Jiang]{xu2023wizardlm}
Can Xu, Qingfeng Sun, Kai Zheng, Xiubo Geng, Pu~Zhao, Jiazhan Feng, Chongyang
  Tao, and Daxin Jiang.
\newblock Wizardlm: Empowering large language models to follow complex
  instructions, 2023.

\bibitem[Xue et~al.(2020)Xue, Constant, Roberts, Kale, Al-Rfou, Siddhant,
  Barua, and Raffel]{xue2020mt5}
Linting Xue, Noah Constant, Adam Roberts, Mihir Kale, Rami Al-Rfou, Aditya
  Siddhant, Aditya Barua, and Colin Raffel.
\newblock mt5: A massively multilingual pre-trained text-to-text transformer.
\newblock \emph{arXiv preprint arXiv:2010.11934}, 2020.

\bibitem[Yang et~al.(2019)Yang, Zhang, Tar, and Baldridge]{Yang2019PAWSXAC}
Yinfei Yang, Yuan Zhang, Chris Tar, and Jason Baldridge.
\newblock Paws-x: A cross-lingual adversarial dataset for paraphrase
  identification.
\newblock In \emph{EMNLP}, 2019.

\bibitem[Ye et~al.(2023)Ye, Hwang, Yang, Yun, Kim, and Seo]{ye2023context}
Seonghyeon Ye, Hyeonbin Hwang, Sohee Yang, Hyeongu Yun, Yireun Kim, and Minjoon
  Seo.
\newblock In-context instruction learning.
\newblock \emph{arXiv preprint arXiv:2302.14691}, 2023.

\bibitem[Zeng et~al.(2022)Zeng, Liu, Du, Wang, Lai, Ding, Yang, Xu, Zheng, Xia,
  et~al.]{zeng2022glm}
Aohan Zeng, Xiao Liu, Zhengxiao Du, Zihan Wang, Hanyu Lai, Ming Ding, Zhuoyi
  Yang, Yifan Xu, Wendi Zheng, Xiao Xia, et~al.
\newblock Glm-130b: An open bilingual pre-trained model.
\newblock \emph{arXiv preprint arXiv:2210.02414}, 2022.

\bibitem[Zhang et~al.(2022)Zhang, Roller, Goyal, Artetxe, Chen, Chen, Dewan,
  Diab, Li, Lin, et~al.]{zhang2022opt}
Susan Zhang, Stephen Roller, Naman Goyal, Mikel Artetxe, Moya Chen, Shuohui
  Chen, Christopher Dewan, Mona Diab, Xian Li, Xi~Victoria Lin, et~al.
\newblock Opt: Open pre-trained transformer language models.
\newblock \emph{arXiv preprint arXiv:2205.01068}, 2022.

\bibitem[Zhang et~al.(2019)Zhang, Baldridge, and He]{paws2019naacl}
Yuan Zhang, Jason Baldridge, and Luheng He.
\newblock {PAWS: Paraphrase Adversaries from Word Scrambling}.
\newblock In \emph{NAACL}, 2019.

\bibitem[Zhou et~al.(2023)Zhou, Liu, Xu, Iyer, Sun, Mao, Ma, Efrat, Yu, Yu,
  Zhang, Ghosh, Lewis, Zettlemoyer, and Levy]{zhou2023lima}
Chunting Zhou, Pengfei Liu, Puxin Xu, Srini Iyer, Jiao Sun, Yuning Mao, Xuezhe
  Ma, Avia Efrat, Ping Yu, Lili Yu, Susan Zhang, Gargi Ghosh, Mike Lewis, Luke
  Zettlemoyer, and Omer Levy.
\newblock Lima: Less is more for alignment, 2023.

\end{thebibliography}
